%% file: uni-parser.tex
\documentclass{article}

\usepackage{techreport}
\usepackage{geometry}
\usepackage{multirow}
\usepackage{graphicx}
\usepackage{multicol}
\usepackage{hyperref}
\usepackage{adjustbox}
\usepackage{caption}
\usepackage{subcaption}

\usepackage[utf8]{inputenc} 
\usepackage[T1]{fontenc}    
\usepackage{hyperref}       
\usepackage{url}            
\usepackage{booktabs}       
\usepackage{amsfonts}       
\usepackage{nicefrac}       
\usepackage{microtype}      
\usepackage{xcolor}         
\usepackage{threeparttable} 

\usepackage{tikz}           

\title{\raisebox{-0.2\height}{\includegraphics[height=2.0em]{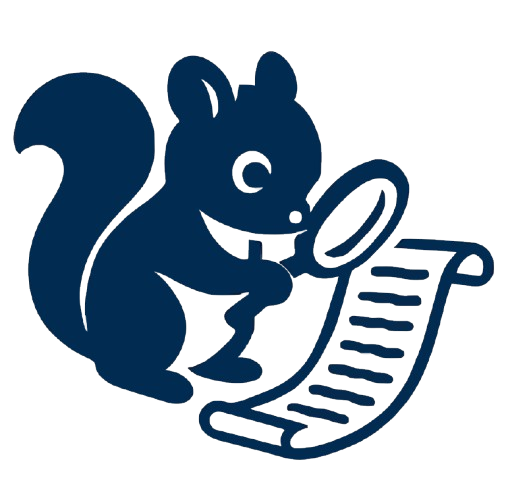}} ~Uni-Parser Technical Report}

\author{%
  DP Technology * \\\\
  December 16, 2025
}

\begin{document}

\maketitle

\vspace{-2.0em} 

\begin{center}
    \renewcommand{\arraystretch}{0.8}
    \begin{tabular}{c} 
        \href{https://uni-parser.github.io}{
            \raisebox{-0.6ex}{\includegraphics[height=1.2em]{images/uniparser_logo.png}}
            \;\;\texttt{https://uni-parser.github.io}
        } \\[0.2em]
        \href{https://huggingface.co/UniParser}{
            \raisebox{-0.5ex}{\includegraphics[height=1.0em]{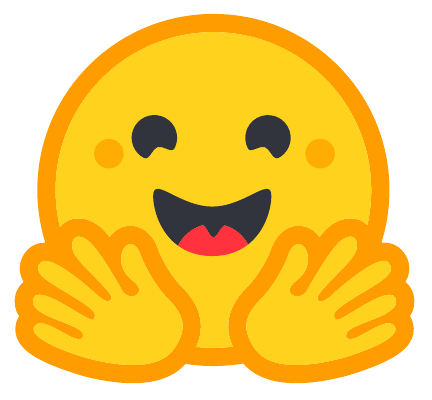}}
            \;\;\texttt{https://huggingface.co/UniParser}
        } 
    \end{tabular}
\end{center}


\input{sec/0_abstract}

\input{sec/1_intro}

\input{sec/3_framework}

\input{sec/4_infra}

\input{sec/5_data}

\input{sec/6_performance}

\input{sec/7_downstream}

\input{sec/8_experience}

\input{sec/9_summary}

{
\small
\bibliography{uni-parser}
}

\newpage
\appendix

\input{appendix/authors}

\input{appendix/examples}

\end{document}

%% file: sec/0_abstract.tex
\begin{abstract}
This technical report introduces \textit{Uni-Parser}, an industrial-grade document parsing engine tailored for scientific literature and patents, delivering high throughput, robust accuracy, and cost efficiency. Unlike pipeline-based document parsing methods, Uni-Parser employs a modular, loosely coupled multi-expert architecture that preserves fine-grained cross-modal alignments across text, equations, tables, figures, and chemical structures, while remaining easily extensible to emerging modalities. The system incorporates adaptive GPU load balancing, distributed inference, dynamic module orchestration, and configurable modes that support either holistic or modality-specific parsing. Optimized for large-scale cloud deployment, Uni-Parser achieves a processing rate of up to 20 PDF pages per second on 8 × NVIDIA RTX 4090D GPUs, enabling cost-efficient inference across billions of pages. This level of scalability facilitates a broad spectrum of downstream applications, ranging from literature retrieval and summarization to the extraction of chemical structures, reaction schemes, and bioactivity data, as well as the curation of large-scale corpora for training next-generation large language models and AI4Science models.
\end{abstract}

%% file: sec/1_intro.tex
\section{Introduction}

The rapid advancement of large language models (LLMs) has significantly expanded the scope of document-centric applications, ranging from intelligent assistants and domain-specific agents to automated knowledge base construction. Central to these developments is the ability to reliably parse and structure information from PDF documents, which remain the dominant medium for disseminating scientific knowledge. High-quality structured data extracted from scientific literature is particularly critical for downstream LLM applications, enabling accurate reasoning, retrieval-augmented generation, and decision-making across a wide spectrum of scientific and industrial tasks.  

Among scientific domains, chemical and biomedical literature holds exceptional importance and immense commercial value. Parsing such documents enables the creation of comprehensive molecular and reaction databases, bioactivity repositories, material structure archives, and experimental characterization datasets. These resources not only accelerate drug discovery and materials design but also serve as the foundation for emerging AI4Science research. However, despite decades of progress in OCR, table recognition, and layout analysis, the majority of the hundreds of millions of scientific and patent PDFs remain underexploited. This is largely because the PDF format, while optimized for human readability, poses extraordinary challenges for computational processing: extraction is prohibitively costly.

Current approaches, both pipeline-based \cite{livathinos2025docling, cui2025paddleocr, wang2024mineru, mathpix} and VLM-based \cite{blecher2023nougat, wei2024got, nassar2025smoldocling, li2025monkeyocr, ocrflux, rednote_hilab_dots_ocr_2025, mistral_ocr}, face three major challenges. First, they are computationally inefficient, making it prohibitively costly to parse tens of millions of documents at scale. Second, their performance on non-textual modalities such as formulas, tables, charts, and chemical structures remains limited, with low accuracy and poor robustness. Third, the complex and heterogeneous layouts of scientific and patent documents often lead to unreliable segmentation and structural analysis, further degrading downstream usability. Pipeline methods offer higher throughput but poor generalization, while VLM methods generalize better but suffer from low efficiency and limited extensibility. Collectively, these limitations hinder the construction of large-scale, high-quality knowledge bases required for both academic research and industrial innovation.

To address these challenges, we present \textit{Uni-Parser}, an industrial-grade, multi-modal PDF parsing engine purpose-built for scientific literature and patents. Uni-Parser follows the pipeline-based methods, and combines high throughput with state-of-the-art accuracy through a modular, loosely coupled architecture composed of specialized expert models for different modalities. The system introduces a set of key innovations:  

\begin{itemize}
\item \textbf{High-efficiency large-scale inference}: A distributed microservice design with dynamic GPU load balancing enables real-time parsing throughput, supporting fast and cost-effective inference over billions of document pages.  
\item \textbf{Accurate multi-modal parsing}: A suite of domain-specialized, lightweight expert models achieves state-of-the-art accuracy across text, equations, tables, figures, and chemical structures.  
\item \textbf{Robust layout recognition for scientific and patent documents}: A newly designed layout analysis and reading order algorithm tailored to complex publishing formats greatly improves reliability in handling dense, irregular, and domain-specific page structures.
\end{itemize}  

Together, these contributions establish Uni-Parser as a scalable and extensible foundation for structured document understanding. By transforming unstructured PDFs into clean, machine-actionable representations, Uni-Parser not only supports immediate applications such as literature retrieval, summarization, and knowledge extraction, but also enables the construction of domain-specific repositories in chemistry, materials science, and biomedicine—paving the way for data-driven AI4Science innovation.

%% file: sec/3_framework.tex
\section{Algorithm Framework}

\subsection{Overall Framework}

\begin{figure}[htbp]
  \centering
  \includegraphics[width=1.0\textwidth]{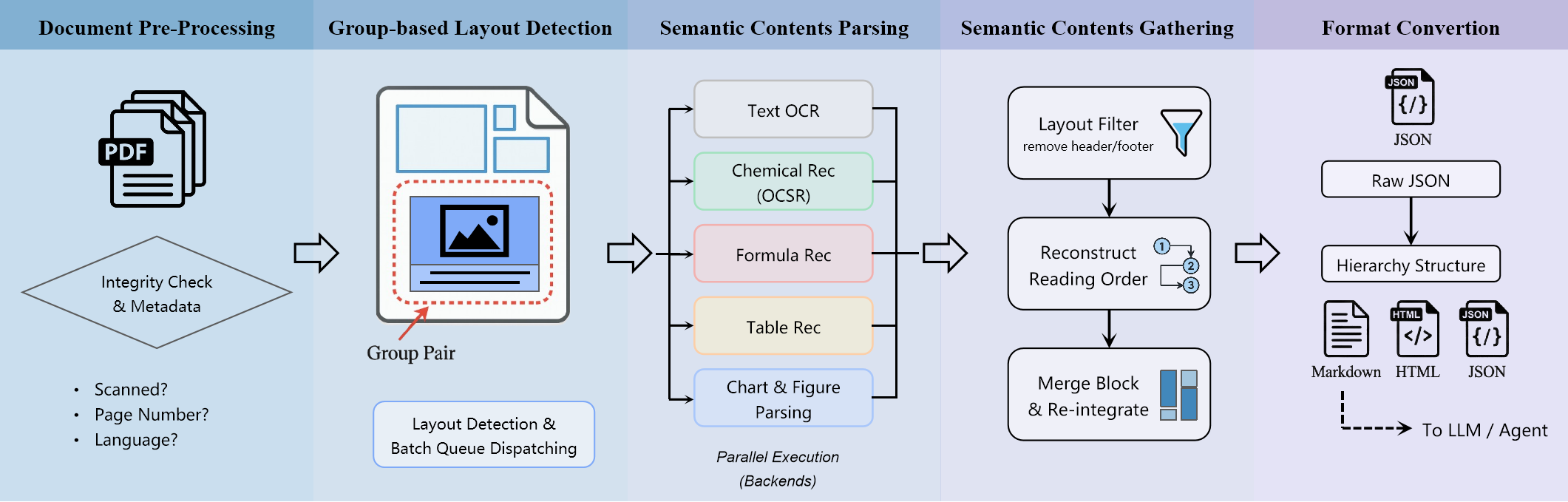}
  \caption{Sketch of the Uni-Parser pipeline. Uni-Parser converts unstructured PDFs into clean, hierarchical, multimodal outputs (text, formulas, tables, figures, and chemical structures). These enriched representations are designed to be readily consumed by large language models, enabling more accurate understanding, reasoning, and document-level operations.}
  \label{fig:pipeline}
\end{figure}

Uni-Parser adopts a modular design with a strong emphasis on extensibility. As shown in the figure, it first conducts validation and pre-processing on PDF file inputs. It then converts these unstructured documents into machine-readable formats (e.g., JSON) or LLM-compatible representations (e.g., Markdown and HTML) through a sequence of models and processing stages. These stages are organized into five principal components:

\textbf{Document Pre-Processing}:
During pre-processing, the system ingests PDFs from user uploads or URLs, verifies file integrity and encryption, and extracts metadata such as page count, dimensions, and text accessibility. Documents with corrupted or garbled text are marked as non-extractable. If embedded text layers exist, they are directly extracted; otherwise, a lightweight OCR method provides partial text for language identification. Uni-Parser supports over 80 languages for OCR mode. This stage typically takes around one hundred milliseconds and supports parallel processing of multiple PDFs.

\textbf{Group-based Layout Detection}:
The layout detection model locates each semantic block in the PDF pages and identifies its semantic category. Unlike conventional approaches \cite{cui2025paddleocr, wang2024mineru}, our group-based layout detection model recognizes naturally paired semantic components—such as image–caption, table–title, and molecule–identifier—preserving visual–semantic associations that are critical for structured information extraction and accurate reading-order construction. Finally, different semantic blocks are rendered at dynamic resolutions and forwarded to designated models for image-to-text processing. We employ a greedy batch-stacking strategy for rendering page images, performing layout recognition, and dispatching semantic groups to downstream microservices. As soon as a batch accumulates in the queue, it is immediately processed by subsequent modules, allowing the latency of this stage to be almost entirely masked by other processing steps.

\textbf{Semantic Contents Parsing}:
Semantic contents are parsed by routing each block to the appropriate specialized model. General text blocks are processed with OCR, tables with table recognition models, mathematical formulas with formula recognition models, molecular structures with optical chemical structure recognition (OCSR), chemical reactions with dedicated reaction extraction models, and charts with chart parsing models. In total, over ten sub-models and specialized procedures handle diverse content types, including text, tables, molecular structures, chemical reactions, formulas, and charts. All models operate in parallel to maximize efficiency. For image blocks, the system can retain either visual descriptions or raw image data, depending on downstream application requirements. This stage is the most time-consuming stage.

\textbf{Semantic Contents Gathering}:
Building on the parsed results from the previous stage, this phase filters out non-essential elements such as headers and footers while preserving key combinations like figure–caption or table–caption pairs and horizontal separators. Content blocks within each page are reordered to reflect the logical structure of the document. Multimodal elements embedded in text lines or table cells—such as inline equations, chemical structures, or charts—are reintegrated into their corresponding text or table. Cross-page and multi-column content, including tables, paragraphs, and reaction schemes, is merged to maintain coherence. The system also incorporates the original PDF’s section hierarchy to guide the final organization. The output of this stage provides a fully structured, semantically enriched representation of the document, suitable for downstream analysis and reading-order reconstruction.

\textbf{Output Formatting and Semantic Chunking}:
To support diverse downstream tasks, the fully parsed document can be exported in task-specific formats, or as a complete plain text, interleaved image–text, Markdown, or HTML. Thanks to group-based layout detection and the merging of semantic content across columns and pages, reconstructed paragraphs or semantic groups are output as properly chunked data, which improves semantic coherence. This approach streamlines chunking and facilitates more efficient downstream processing, such as retrieval-augmented generation (RAG). Furthermore, PDF section headings are integrated into the document structure when available. These metadata elements enrich the final representation with hierarchical navigation cues, improving both user-facing applications and LLM-driven analysis.

\subsection{Group-based Layout Analysis}
\label{subsec:layout}
Layout detection is a foundational prerequisite for all subsequent document analysis tasks and represents a pivotal component within document parsing systems. Its efficacy ultimately establishes the performance ceiling for the entire system. Acknowledging the profound diversity and complexity of scientific document layouts, along with the critical importance of grouping relationships among heterogeneous semantic elements, we introduce a layout understanding paradigm that diverges significantly from conventional approaches \cite{cui2025paddleocr, zhao2024doclayout, pfitzmann2022doclaynet, livathinos2025docling}: a group-based tree-structured layout representation.

\begin{figure}[htbp]
  \centering
  \includegraphics[width=1.0\textwidth]{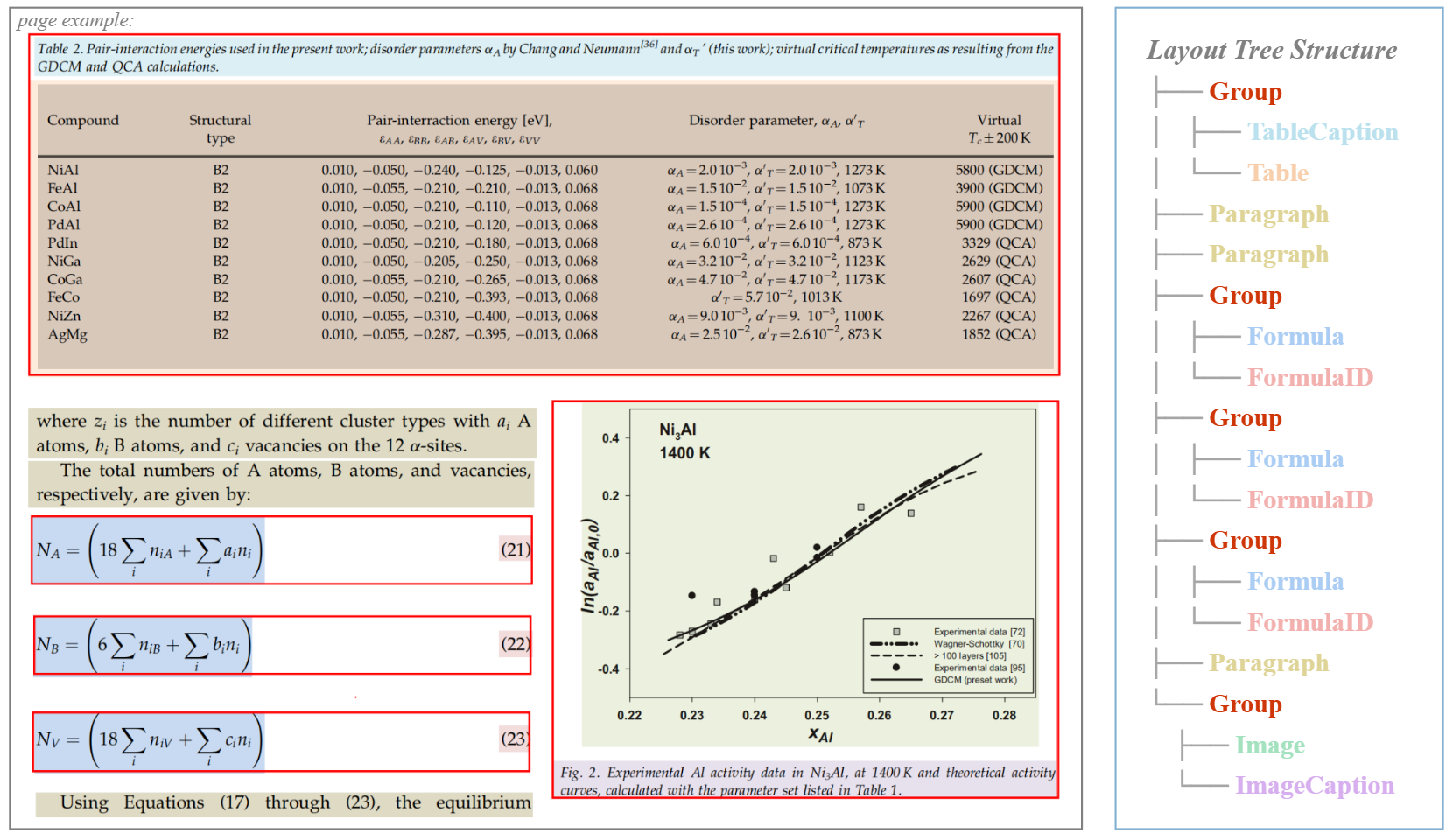}
  \caption{An inference example of group-based layout detection used Uni-Parser-LD. The final output is a hierarchical layout tree structure.}
  \label{fig:layout_example}
\end{figure}

Operationally, we conceptualize the layout of a document page as a hierarchical organization, enabling the aggregation of semantically related elements into coherent and logical groups. For instance, figures are paired with their captions, tables with their titles, equations with their reference numbers, and molecular structures with their identifiers, among other relational pairings. Figure~\ref{fig:layout_example} shows an example of the layout tree structure. Although document layouts inherently involve multiple levels of hierarchy, in our annotation we restrict the structure to only two layers: the bottom layer and the top layer. The bottom layer covers all fundamental semantic components, serving as the parent nodes of the layout structure tree, whereas the top layer comprises content nested within the bottom layer or other top level semantic elements, functioning as child nodes. The corresponding semantic categories are summarized in Table~\ref{tab:layout_categories}. Importantly, this design still allows downstream post-processing to recover the nested semantic content across multiple hierarchical levels. Based on these concepts, we constructed a group-based layout detection model named \textit{Uni-Parser-LD}. 

\begin{table}[h!]
\small
\centering
\begin{threeparttable}
\caption{Layout type used in Uni-Parser-LD}
\label{tab:layout_categories}
\begin{tabular}{lcccc}
\toprule
\textbf{Layout Type} & \textbf{Layout Layer} & \textbf{Role} & \textbf{Semantic Parsing Category} \\
\midrule
Document Title & \multirow{7}{*}{Bottom Layer} & Main Text \& Structure Role & OCR \\
Section Title  &                       & Main Text \& Structure Role & OCR \\
Paragraph      &                       & Main Text                   & OCR \\
References     &                       & Supplementary Text          & OCR \\
Table of Contents  &                       & Supplementary Text          & OCR \\
Key-value Item      &                       & Main Text                   & OCR \\
Code Block     &                       & Figure / Multi-modal Text          & OCR / Code Parsing \\
\midrule
Header         & \multirow{7}{*}{Bottom Layer} & Functional Role             & -- \\
Footer         &                       & Functional Role             & -- \\
Footnote       &                       & Supplementary Text          & OCR \\
Sidebar        &                       & Functional Role             & -- \\
Page Number    &                       & Functional Role             & OCR \\
Watermark      &                       & Functional Role             & -- \\
Divider Line   &                       & Functional Role             & -- \\
\midrule
Formula\tnote{1}        & \multirow{3}{*}{Bottom Layer} & Multi-modal Text            & Formula Recognition \\
Table\tnote{2}     &                       & Multi-modal Text            & Table Structure Recognition \\
Image\tnote{3}        &                       & Figure / Multi-modal Text   & - \\
\midrule
Formula (Inline)        & \multirow{5}{*}{Top Layer}    & Multi-modal Text            & Formula Recognition \\
Molecule\tnote{4}      &                       & Multi-modal Text            & OCSR \\
Chemical Reaction\tnote{5}      &                       & Figure / Multi-modal Text            & Reaction Parsing \\
Chart\tnote{5}       &                       & Figure / Multi-modal Text            & Chart to Table \\
Figure\tnote{5}  &                       & Figure / Multi-modal Text   & Image Caption \\
\bottomrule
\end{tabular}
\begin{tablenotes}
\item[1] Grouped with formula ID.
\item[2] Grouped with table caption and table footnote.
\item[3] Grouped with image caption.
\item[4] Grouped with molecule identifier and Markush description.
\item[5] Grouped with figure legend and figure caption.
\end{tablenotes}
\end{threeparttable}
\end{table}

Considering the wide variety of authoring and rendering styles across layouts and modalities, we employ a large-scale dataset of PDF pages for training. We built an in-house layout detection dataset containing 500k pages. Of these, 220k pages are carefully human annotated with group-based layout labels from a diverse corpus. The database primarily consists of scientific journal data and patent data from various patent offices. It also includes preprints, books, and other types of data across different fields, spanning a total of 85 languages. For more details, refer to Section~\ref{sec:data}. The remaining pages are synthetic data used for pretraining. Our experiments show that when high-quality, large-scale real data is available, pretraining with existing synthetic datasets, such as DocSynth300K~\cite{zhao2024doclayout}, can be counterproductive. This is due to their low-fidelity rendering, stylistic deviations from authentic documents, and limited diversity imposed by manually defined generation rules, which fail to capture the complexity of real-world layouts. To address this, we generate synthetic samples from real layouts using controlled modifications—including element merging, spatial perturbations, and semantic content substitutions—allowing the model to effectively leverage synthetic data for pretraining.

To meet real-time processing requirements, our selection of a backbone architecture for layout detection prioritizes high efficiency. A survey of contemporary object detection methods reveals that models such as RT-DETRv2~\cite{rtdetr}, YOLOv12~\cite{tian2025yolov12}, and D-FINE~\cite{peng2024dfine} can exhibit training instability in our layout detection task. Meanwhile, the extensive modifications made to YOLOv8~\cite{jocher2023ultralytics} in DocLayout-YOLO~\cite{zhao2024doclayout} yield only marginal performance improvements in our setting. Consequently, after balancing performance with inference speed, we select a modified DETR-based architecture as the backbone for our group-based layout detection model.

The choice of input resolution presents another critical trade-off between processing speed and detection accuracy. While prior research~\cite{steiner2024paligemma} suggests that a resolution of $896 \times 896$ suffices for most documents, our empirical evaluations demonstrate its insufficiency for achieving consistently high performance. Therefore, to accommodate diverse document aspect ratios and strike a superior balance between speed and accuracy, we adopt an input resolution of $1024 \times 768$.

\subsection{OCR}

Uni-Parser operates in two complementary OCR modes: extraction and recognition. When textual content can be directly extracted from the PDF in the first stage, the system bypasses OCR and utilizes the extracted text directly. Otherwise, or when explicitly requested by the user, Uni-Parser performs recognition using OCR model, such as the PP-OCRv5~\cite{cui2025paddleocr} series. The default OCR backbone supports a broad character set encompassing Simplified and Traditional Chinese, English, Japanese, and a wide range of special symbols. For other languages, the system automatically invokes the corresponding language-specific PP-OCRv5 model. For more complex cases—such as multilingual mixtures or documents containing numerous out-of-vocabulary OCR symbols—we additionally support switching to PaddleOCR-VL~\cite{cui2025paddleocrvlboostingmultilingualdocument} in a high-quality mode, trading inference speed for improved robustness in challenging scenarios, with support for more than 109 languages (ISO 639). In practice, additional preprocessing is applied in certain scenarios. For example, when text is embedded within tables, Uni-Parser employs a text orientation detector to ensure accurate recognition.

\begin{figure}[htbp]
  \centering
  \includegraphics[width=1.0\textwidth]{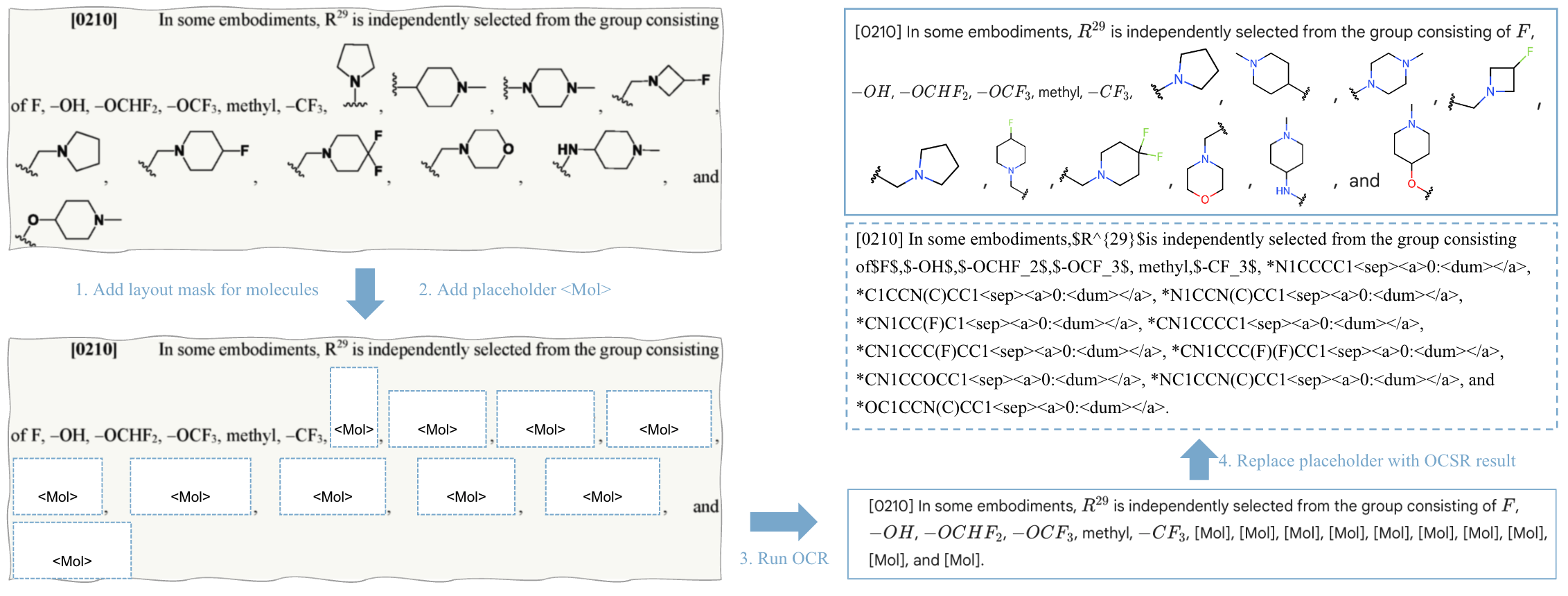}
  \caption{An example of the OCR model inference workflow in Uni-Parser. When a top-layer layout element overlaps a bottom-layer layout block, the system substitutes it with a placeholder before performing OCR. The placeholder is then resolved during post-processing, enabling fast and accurate multimodal parsing.}
  \label{fig:ocr_example}
\end{figure}

After OCR, the system reconstructs the document by aligning recognized text with the spatial layout returned by the layout detection module. Inline non-textual elements—such as chemical structures and mathematical expressions embedded within text lines—are explicitly identified and preserved. Shown in Figure~\ref{fig:ocr_example}, during OCR, these elements are temporarily replaced with modality-specific placeholders, which are subsequently parsed by dedicated modules (e.g., OCSR or formula recognition) and reintegrated into the final semantic representation of the document. 

\subsection{Table Structure Recognition}
Tables represent compact yet structurally complex layouts, as each cell may contain heterogeneous semantic blocks, including plain text, mathematical formulas, molecular structures, reaction schemes, images, or even nested sub-tables. The presence of such multi-level nesting makes it challenging for a single end-to-end model to perform generalized OCR on tables. To address this, we adopt a modular strategy that decouples table structure recognition from multimodal content parsing. This approach not only improves interpretability but also enhances robustness and overall performance.

Before structure recognition, Uni-Parser recovers the orientation of tables, since rotated layouts frequently occur in patents and supplementary materials of scientific articles, particularly when tables are extended across pages. Instead of employing a dedicated orientation model, we leverage layout predictions and auxiliary metadata to perform lightweight orientation recovery, achieving accuracy comparable to that of specialized models while significantly improving efficiency.

We adopt an efficient table structure recognition model, SLANet~\cite{li2022ppslanet}, trained on a corpus of one million tables that integrates a cleaned version of PubTabNet~\cite{zhong2019image}, SynthTabNet~\cite{nassar2022tableformer}, and our in-house synthetic dataset of line-based tables enriched with multimodal content. The synthetic dataset covers both bordered and borderless formats and incorporates diverse elements, including molecule structures, charts, images, formulas, and text. This comprehensive training strategy allows SLANet to achieve strong generalization and superior performance on complex real-world table layouts.

\begin{figure}[htbp]
  \centering
  \includegraphics[width=1.0\textwidth]{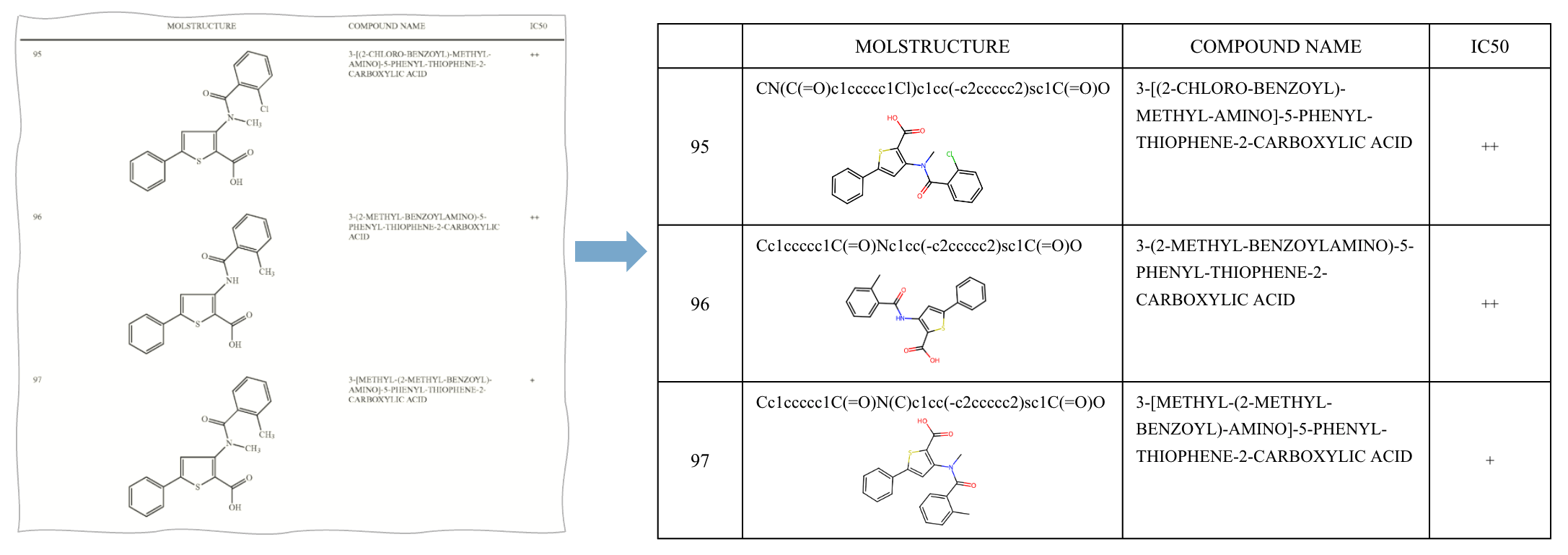}
  \caption{An example of table structure recognition results produced by Uni-Parser. By decoupling table structure recognition from table content recognition, the system achieves improved robustness, supports multimodal nesting within tables.}
  \label{fig:table_example}
\end{figure}

\begin{figure}[htbp]
  \centering
  \includegraphics[width=1.0\textwidth]{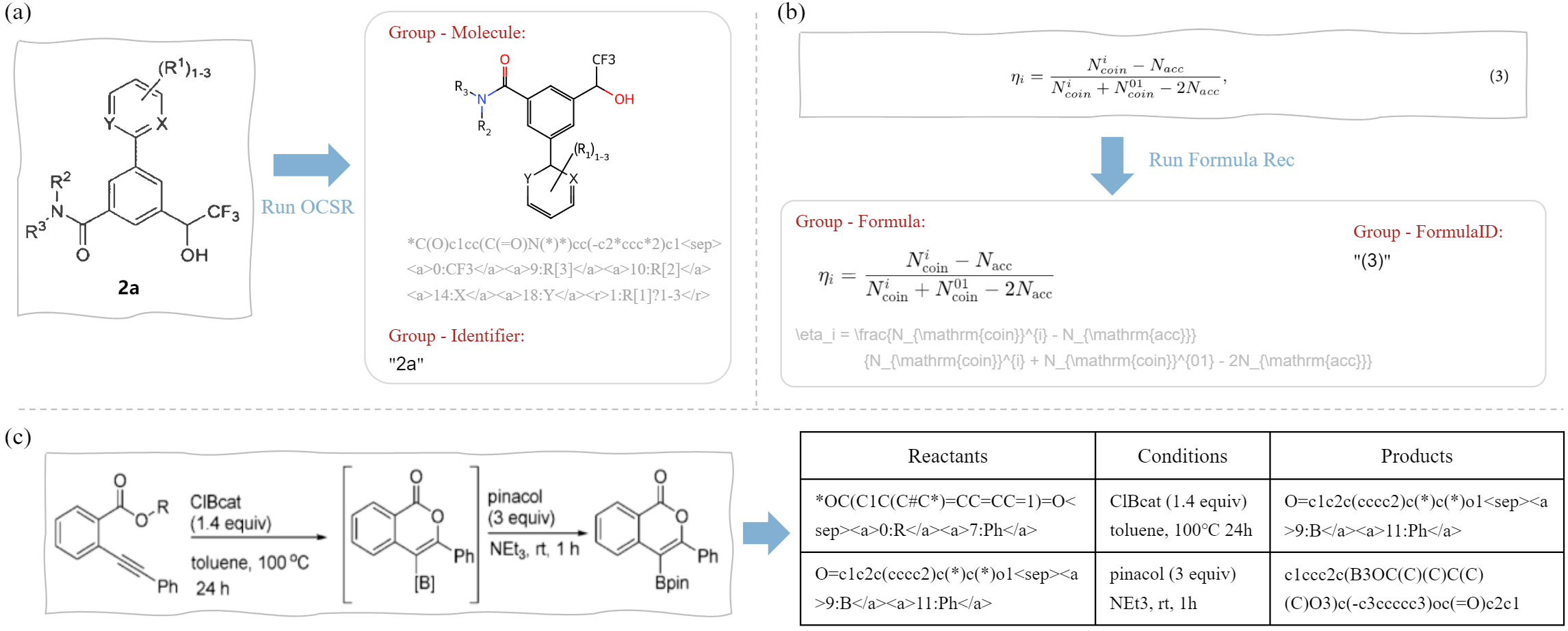}
  \caption{Examples of multi-modal recognition results produced by Uni-Parser. (a) Molecular structures are correctly associated with their corresponding identifiers. (b) Mathematical formulas are accurately linked to their formula IDs. (c) An organic chemical reaction is parsed into a structured reactant--condition--product triplet.}
  \label{fig:formula_mol_example}
\end{figure}


\subsection{Formula Recognition}

The system integrates a mathematical formula recognition module fine-tuned from PP-Formula~\cite{cui2025paddleocr}. The module converts mathematical expressions and chemical equations into LaTeX sequences in an end-to-end manner. It handles both standalone formulas and inline formulas or equations within paragraph text. For standalone formulas, the layout stage groups the expression with its reference number to form a structured representation.

\subsection{Chemical Expression Recognition}

Chemical structures are a fundamental modality in scientific literature and patents, particularly in chemistry, pharmaceuticals, biology, and materials. Uni-Parser adopts the end-to-end OCSR architecture MolParser~\cite{fang2024molparser} for molecular recognition, which translates molecular images into an Extended SMILES (E-SMILES) representation. We further introduce \textit{MolParser 1.5}, which extends MolParser with an expanded pretraining corpus. In addition to the synthetic MolParser-7M dataset, we add a real-world dataset of 10 million images pseudo-labeled via cross-validation using multiple MolParser models and a fine-tuned MolScribe~\cite{qian2023molscribe}. With a larger proportion of in-the-wild chiral molecules and Markush structures, MolParser 1.5 yields a more balanced and comprehensive pretraining dataset, and after fine-tuning on the MolParser-SFT dataset, it produces a more robust end-to-end OCSR model.

For chemical reaction image recognition, we adopt a pipeline approach. First, we detect the texts and molecular structures within the reaction in layout stage. Then, we identify the associations among blocks to construct a reactant–condition–product graph. Finally, we aggregate the results to obtain the parsed reaction equation.

\subsection{Chart and Scientific Figure Understanding}
Uni-Parser incorporates an optional module for chart and scientific figure understanding. When activated, the system converts charts into their underlying data tables. For figures that are not amenable to accurate tabular representation—such as spectra or complex scientific diagrams—the system generates detailed image captions that convey essential information, including key numerical values.

For chart understanding, we fine-tuned a Qwen-2.5-VL-3B~\cite{bai2025qwen2} model on a dataset consisting of 500k generated samples, 300k open-source samples, and 170k real-world charts. This training significantly enhances the model’s ability to convert charts into tables, with a particular focus on multi-subplot charts, which frequently appear in scientific literature.

To support accurate scientific figure captioning, we curated a large-scale, high-quality dataset of 4 million samples. The core of this dataset consists of 3 million scientific images meticulously collected from high-impact publications. To enhance caption quality, we leveraged multiple multi-modal large language models (MLLMs), including but not limited to GPT-5~\cite{openai2025gpt5} and Gemini 2.5~\cite{deepmind2025gemini25pro}, to rewrite the original captions by integrating visual content with raw captions and relevant contextual information from the associated papers. The dataset is further augmented with additional sources, including ChemPile~\cite{mirza2025chempile}, molecular descriptions from PubChem, Electron Microscopy image captions~\cite{wang2025uniem3muniversalelectronmicrograph}, and other experimental characterization datasets. Leveraging this comprehensive dataset, we fine-tuned a Qwen-2.5-VL-3B model, which named as \textit{SciParser}, enabling it to generate precise, context-aware captions for a diverse range of scientific figures.

\begin{figure}[htbp]
  \centering
  \includegraphics[width=\linewidth]{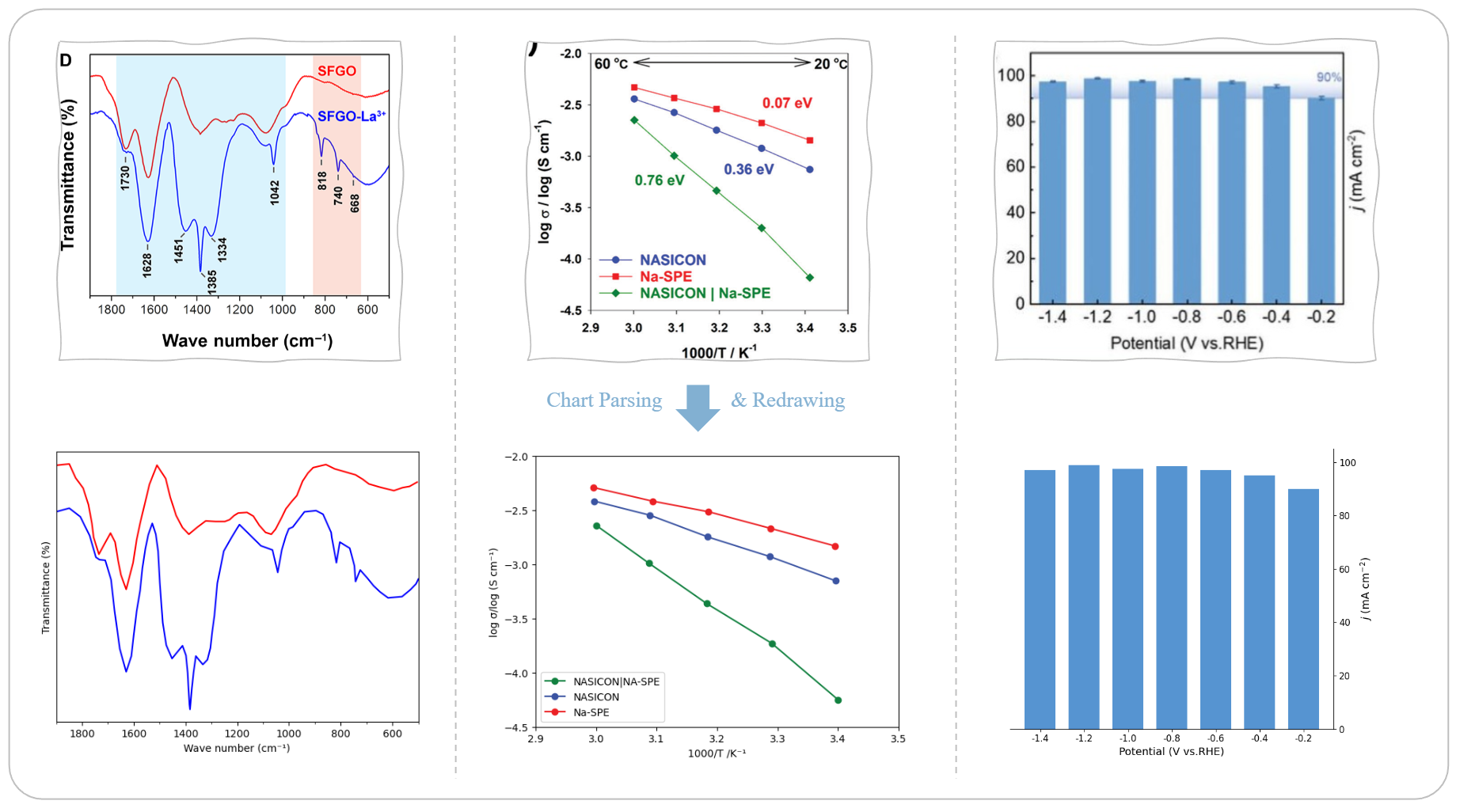}
  \caption{Examples of chart recognition results produced by Uni-Parser. Chart images are first parsed into underlying data tables, which are then re-rendered in a style consistent with the original figures for visualization and presentation.}
  \label{fig:chartparsing_demo}
\end{figure}

\begin{figure}[htbp]
  \centering
  \includegraphics[width=\linewidth]{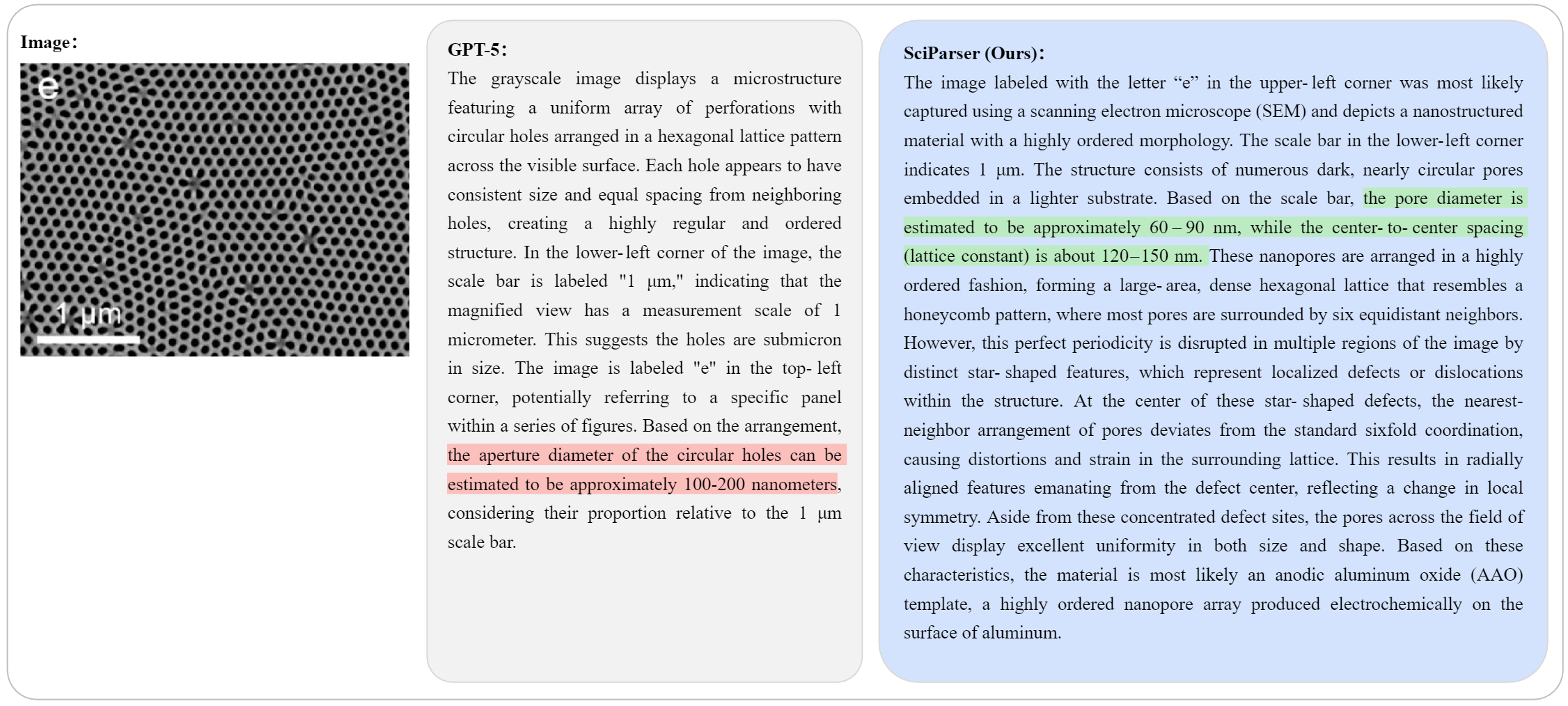}
  \caption{Examples of scientific figure captioning results produced by Uni-Parser. SciParser, a dedicated scientific image captioning submodule within Uni-Parser, converts scientific figures into high-information-density textual descriptions that capture key semantic and structural attributes, facilitating downstream understanding, retrieval, and database construction.}
  \label{fig:sciparser_demo}
\end{figure}

\subsection{Reading Order Recovery}

To accurately reconstruct the reading order of content blocks on each page, our system integrates a set of spatial and semantic heuristics tailored for complex, real-world documents.

\textbf{XY-cut.} This method recursively partitions the page along dominant whitespace regions, yielding a hierarchical binary reading tree that captures coarse layout structure.

\textbf{Gap-tree analysis.} It leverages inter-block whitespace, geometric proximity, and alignment cues to infer plausible reading flows, particularly in dense or irregular layouts.

\textbf{Group-based strategies.} These strategies cluster semantically related elements---even when spatially distant---such as linking figures with captions or molecules with identifiers, thereby preserving logical semantics prior to global ordering.

By combining these complementary techniques, the system produces a natural and coherent reading sequence across multi-column, multi-lingual, and multi-modal layouts.

\subsection{Cross-column and Cross-page Consolidation}

After establishing the per-page reading order, the system further consolidates items that are visually separated but semantically continuous:

\begin{itemize}
    \item \textbf{Cross-column merging:} reconnecting paragraph fragments split by column boundaries using text-flow continuity and linguistic coherence.
    \item \textbf{Cross-page merging:} linking entities that span pages---such as long tables, multi-step reaction schemes, or running paragraphs---based on layout carry-over cues and semantic consistency.
    \item \textbf{Multimodal linkage:} Associating diagrams with their corresponding identifiers or descriptions, even when they appear on adjacent pages. Representative examples are provided in Appendix~\ref{appendix:cross_example}.
\end{itemize}

This second-stage consolidation restores logically unified content into coherent units, improving the fidelity of the reconstructed document structure.

%% file: sec/4_infra.tex
\section{Infrastructure}

\subsection{Distributed Multi-Expert Architecture}

\textbf{Microservice architecture:} 
Uni-Parser adopts a microservice-based multi-expert architecture that enables large-scale distributed inference. Each modality-specific expert (e.g., text, molecules, formulas, tables, reactions, and charts) is deployed as an independent microservice, with multiple nodes running in parallel to process inference requests. Layout analysis first partitions documents into batches, which are distributed across nodes for concurrent processing. Detected regions are then enqueued into modality-specific task queues, where batched requests are asynchronously dispatched to the corresponding expert services. Finally, outputs from all modules are aggregated into a unified generalized OCR representation, followed by post-processing and structured result generation.

\textbf{Dynamic load balancing:}
A fine-grained scheduling layer dynamically allocates computational resources both within and across modules. This design supports adaptive scaling under varying workloads, prevents bottlenecks in individual experts, and ensures stable throughput in heterogeneous multi-modal parsing.

\begin{figure}[htbp]
  \centering
  \includegraphics[width=1.0\textwidth]{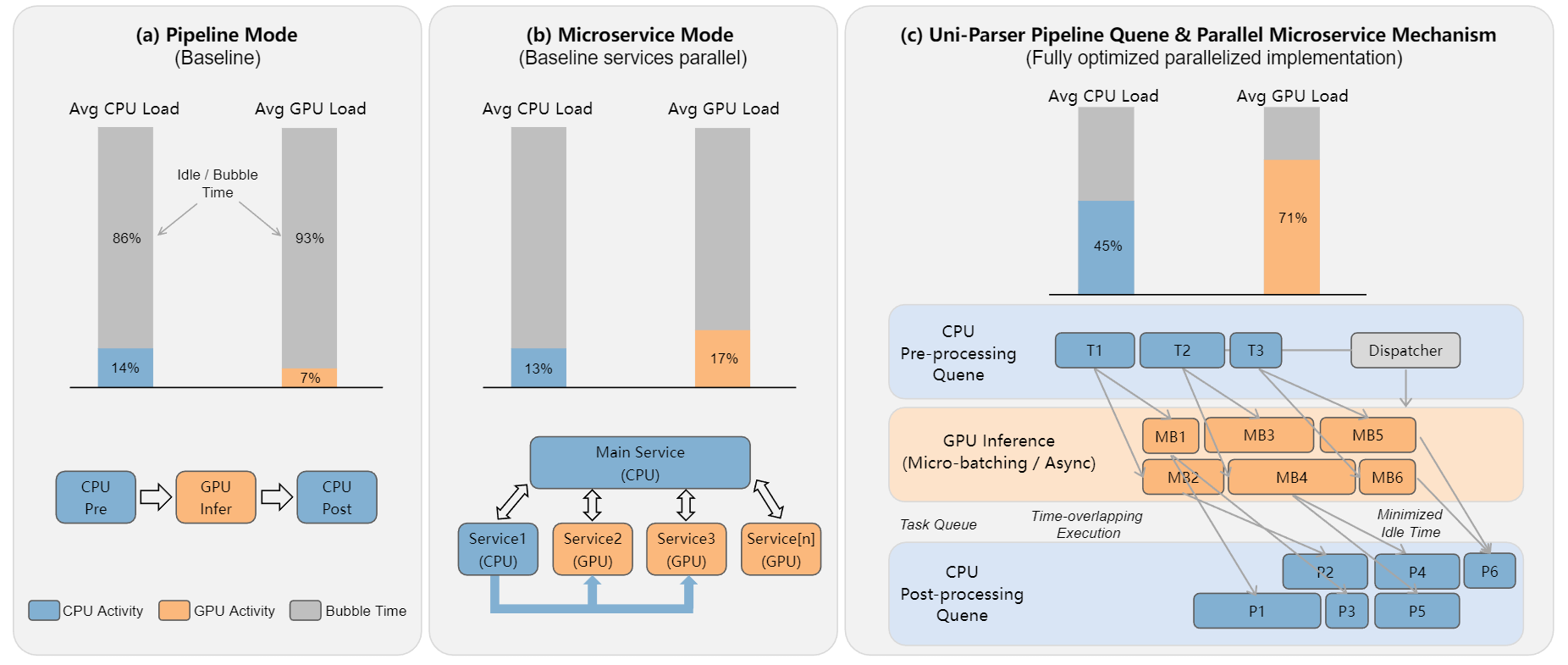}
  \caption{Comparison of document parsing infrastructures. (a) Sequential pipeline: Most existing document parsing frameworks adopt a strictly serial workflow, in which layout analysis is completed first, followed by generalized OCR tasks and subsequent post-processing. (b) Microservice-based parallelism: The tasks are dispatched to multiple microservices that operate in parallel, and the final results are aggregated through a gather stage. (c) Our Uni-Parser pipeline-parallel design: By enabling more frequent lightweight communication and non-blocking execution across modules, Uni-Parser fully exploits parallelism across heterogeneous microservices. Computation and communication are effectively overlapped across GPU and CPU resources, substantially improving throughput and reducing end-to-end latency.}
  \label{fig:bubble-time}
\end{figure}

\textbf{Pipeline Parallel:}
The inference runtime is optimized for efficient GPU parallelism and scheduling, employing multi-process server execution, micro-batching, and asynchronous task management to minimize idle time and sustain high GPU occupancy, while simultaneously ensuring effective utilization of CPU and memory resources across heterogeneous workloads. In particular, CPU pre-processing and post-processing, GPU model inference, and inter-service data transfer are orchestrated in a pipelined manner, enabling time-overlapping execution across stages and thereby further reducing latency in each instance. An analysis of bubble time is presented in Figure~\ref{fig:bubble-time}.

\textbf{Decoupled module updates:} Independent component upgrades that support rapid iteration and performance tuning without full system redeployment, with zero service interruption during updates. This enables a self-reinforcing data flywheel loop, allowing models to be updated on an hourly basis and continuously enhanced with newly acquired active learning samples.

\subsection{Deployment Scaling}

Uni-Parser is designed for elastic scaling in distributed inference environments. System throughput can be increased almost linearly by expanding the number of nodes assigned to each load-balanced microservice. Benchmark experiments demonstrate that the average parsing speed per PDF page scales almost linearly with the number of backend GPUs. 

The Uni-Parser system is deployed on a cluster of 240 NVIDIA L40 GPUs (48 GB each), with 22 CPU cores and 90 GB of host memory per GPU. Under this setting, Uni-Parser parses over 16 million documents within 6 days in the fast mode. Moreover, the system scales to a hybrid cluster of up to 1,000 GPUs (A800 + L20), exhibiting near-linear throughput scaling with respect to available computational resources. These results indicate that Uni-Parser provides a computationally efficient, scalable, and stable infrastructure for large-scale document parsing.

%% file: sec/5_data.tex
\section{Data Engineering}
\label{sec:data}

\subsection{Uni-Parser Data Engine}

The \textit{Uni-Parser Data Engine} is designed to efficiently generate and curate high-quality training data for PDF parsing models, combining synthetic generation, active learning on real documents, and self-training on unlabeled data to minimize human annotation effort while maintaining high accuracy.

\textbf{Step 1: Synthetic Data for Model Bootstrapping.}
Since all PDFs are human-generated rather than naturally structured, we employ large-scale synthetic data to bootstrap various models, including layout detection and several generalized OCR models. Diverse synthetic documents are generated and augmented extensively to pretrain models, providing a strong initialization that accelerates downstream learning and improves generalization.

\textbf{Step 2: Active Learning on Real Documents.}
After endowing the model with fundamental capabilities using large-scale synthetic data, we further improve its performance by iteratively incorporating real-world documents through an active learning framework. Our in-house training corpus is curated from multiple high-quality sources, including:
\begin{itemize}
\item approximately 170 million pages of scientific literature spanning a broad range of disciplines;
\item over 140 million pages of patent documents collected from patent offices worldwide;
\item around 20 million pages of books, technical reports, and miscellaneous documents obtained from Fine-PDFs~\cite{kydlicek2025finepdfs}.
\end{itemize}

An active learning data flywheel, similar to the one used in MolParser \cite{fang2024molparser}, is employed to select the most informative samples for annotation. Specifically, cross-model inference consistency is used as the primary scoring criterion to estimate sample uncertainty and guide data selection during each active learning cycle. In addition, this process is integrated with a curriculum learning strategy, whereby the model is progressively trained on examples of increasing structural and semantic complexity, enabling stable optimization and improved generalization in challenging real-world scenarios.

\textbf{Step 3: Self-Training on Unlabeled Data.}
To further scale training without additional manual labeling, we apply self-training on large collections of unlabeled documents. Predictions from multiple ensembled models are aggregated to estimate confidence scores, and only high-confidence predictions are treated as pseudo-labels. This approach effectively expands the training set at minimal cost while maintaining label quality.

By combining synthetic pretraining, active learning, and self-training, we construct a data flywheel that substantially reduces the amount of human annotations required, accelerates model convergence, and enables high-throughput, low-cost data preparation. Using this three-step approach, we reduce the total annotation volume by {95\%} and cut per-page or per-block annotation time by {90\%}, with {90\%} of annotations for our various models completed in just two months, ensuring rapid, scalable, and efficient model development for large-scale PDF parsing tasks.

\subsection{Uni-Miner Annotation Platform}

To better establish a data flywheel for Uni-Parser with an effective human-in-the-loop pipeline, we designed the \textit{Uni-Miner Annotation Platform} to ensure that samples selected through active learning can be corrected and refined by human annotators with high quality and low cost. The platform integrates a molecular drawing interface and supports annotation across multiple modalities, including molecular structures, text, and layout elements. In addition, we incorporate a recommendation module that assigns the most suitable samples to domain experts based on their specialization and past performance. Each annotator and reviewer is dynamically scored according to accuracy metrics, enabling the system to route the most critical or ambiguous cases to the most experienced and reliable individuals. This design significantly reduces annotation cost, improves throughput, and ensures consistent data quality through mechanisms such as inter-annotator agreement monitoring. In practice, each sample is reviewed by at least two annotators, providing further quality assurance for the downstream learning process.

%% file: sec/6_performance.tex
\section{Performance}

\begin{figure}[t]
  \centering
  \includegraphics[width=1.0\textwidth]{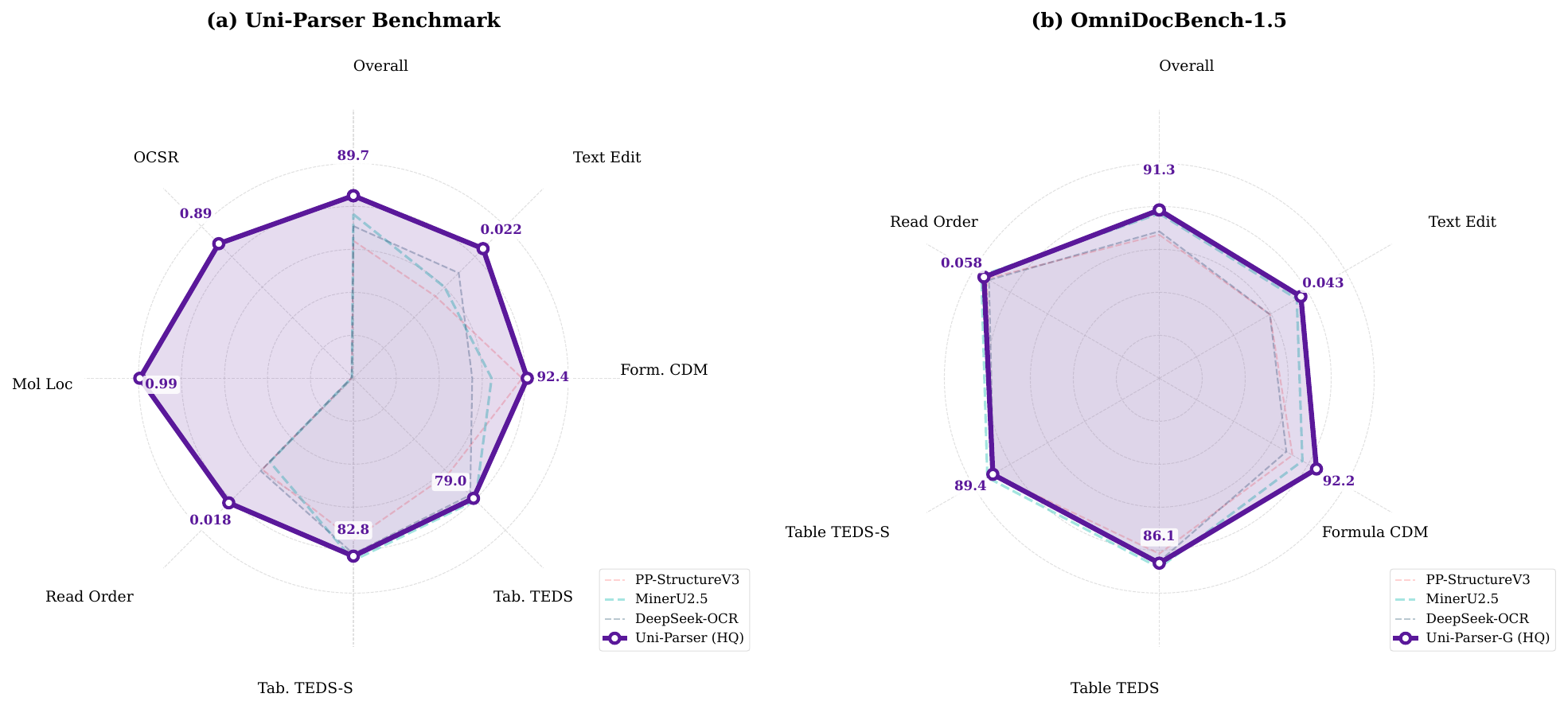}
  \caption{Performance comparison of Uni-Parser with representative document parsing systems across two benchmarks. 
  (a) Uni-Parser Benchmark, a benchmark proposed in this work to evaluate multimodal parsing capabilities on scientific documents and patent literature. 
  (b) OmniDocBench-1.5~\cite{ouyang2025omnidocbench}, a general-purpose benchmark for document parsing. 
  Uni-Parser demonstrates strong and well-balanced performance across most evaluation dimensions, indicating favorable overall capability and robustness compared to existing methods.}
  \label{fig:overall_performace}
\end{figure}

\subsection{Construction of the Uni-Parser Benchmark}
The Uni-Parser benchmark is designed to evaluate layout detection and semantic content recognition in scientific documents. It comprises 150 PDF files collected from international patent documents and research articles, with an emphasis on diversity in document structure, subject matter, domain, and language. The distribution of the benchmark is summarized in Table~\ref{tab:benchmark_distribution}.  
To ensure high data quality, the benchmark is annotated by more than ten domain experts, followed by two rounds of cross-review and an additional round of random quality inspection.

\begin{table}[t]
\small
\centering
\caption{Distribution of document sources in Uni-Parser benchmark.}
\renewcommand{\arraystretch}{1.2}
\begin{tabular}{llcc}
\toprule
\textbf{Source Type} & \textbf{Sub-Source} & \textbf{\#PDFs} & \textbf{Total Pages} \\
\midrule
\multirow{1}{*}{Patent Documents} 
    & From 20 Patent Offices (5 langs) & 50 & 1455 \\
\midrule
\multirow{5}{*}{Scientific Articles} 
    & xRxiv (bioRxiv, medRxiv, ChemRxiv, etc.) & 35 & 696 \\
    & arxiv & 20 & 231 \\
    & ChinaXiv & 15 & 168 \\
    & Nature Communications & 15 & 164 \\
    & Scientific Reports & 15 & 173 \\
\midrule
\multicolumn{2}{l}{\textbf{Total (All Documents)}} & \textbf{150} & \textbf{2887} \\
\bottomrule
\end{tabular}
\label{tab:benchmark_distribution}
\end{table}

\subsection{Overall Results}

We evaluate scientific document parsing performance using an end-to-end evaluation protocol following OmniDocBench~\cite{ouyang2025omnidocbench}. The \textit{Overall} metric excludes molecular-related scores, as existing baselines generally lack molecular recognition and localization capabilities and therefore consistently obtain zero scores on these components. Table~\ref{tab:performance_comparison} reports the comparative results on the proposed Uni-Parser benchmark.

\begin{table*}[htpb]
\centering
\caption{Performance comparison of scientific document parsing methods on Uni-Parser Benchmark}
\label{tab:performance_comparison}
\scriptsize
\setlength{\tabcolsep}{4pt} 
\begin{tabular}{l|l| cccccccc}
\toprule
\textbf{Type} & \textbf{Methods} & 
\begin{tabular}[t]{@{}c@{}}\textbf{Overall}\\\textit{excl.} Mol.\end{tabular}$\uparrow$ & 
\begin{tabular}[t]{@{}c@{}}\textbf{Text}\\\textit{Edit}\end{tabular}$\downarrow$ & 
\begin{tabular}[t]{@{}c@{}}\textbf{Form.}\\\textit{CDM}\end{tabular}$\uparrow$ & 
\begin{tabular}[t]{@{}c@{}}\textbf{Tab.}\\\textit{TEDS}\end{tabular}$\uparrow$ & 
\begin{tabular}[t]{@{}c@{}}\textbf{Tab.}\\\textit{TEDS-S}\end{tabular}$\uparrow$ & 
\begin{tabular}[t]{@{}c@{}}\textbf{Read Order}\\\textit{Edit}\end{tabular}$\downarrow$ & 
\begin{tabular}[t]{@{}c@{}}\textbf{Mol Loc}\\\textit{mAP@50}\end{tabular}$\uparrow$ & 
\begin{tabular}[t]{@{}c@{}}\textbf{OCSR}\\\textit{Acc}\end{tabular}$\uparrow$ \\
\midrule

\multirow{3}{*}{\textbf{Pipeline Tools}} 
& MinerU2-pipeline 2.1.1~\cite{wang2024mineru} & 83.24 & 0.063 & 86.14 & 69.88 & 78.25 & 0.065 & 0.000 & 0.000 \\
& PP-StructureV3~\cite{cui2025paddleocr} & 82.41 & 0.069 & 91.42 & 62.71 & 75.21 & 0.040 & 0.000 & 0.000 \\
& \textbf{Uni-Parser~(Fast)} & 83.06 & 0.061 & 91.74 & 63.54 & 74.97 & 0.047 & 0.994 & 0.886 \\
& \textbf{Uni-Parser~(HQ)} & 89.74 & 0.022 & 92.37 & 79.04 & 82.79 & 0.018 & 0.994 & 0.886 \\
\midrule
\multirow{4}{*}{\textbf{Specialized VLMs}}
& MinerU2-VLM~\cite{wang2024mineru} & 84.93 & 0.049 & 85.09 & 74.60 & 79.43 & 0.041 & 0.000 & 0.000 \\
& MinerU2.5~\cite{wang2024mineru} & 86.72 & 0.060 & 85.73 & 80.41 & 84.82 & 0.045 & 0.000 & 0.000 \\
& DeepSeek-OCR~\cite{wei2025deepseek} & 84.80 & 0.046 & 82.13 & 76.88 & 81.72 & 0.039 & 0.000 & 0.000 \\
& PaddleOCR-VL~\cite{cui2025paddleocrvlboostingmultilingualdocument} & 89.80 & 0.034 & 90.59 & 82.21 & 84.91 & 0.030 & 0.000 & 0.000 \\
\bottomrule
\end{tabular}
\end{table*}

We further report results on OmniDocBench-1.5~\cite{ouyang2025omnidocbench}, although this benchmark is not fully aligned with the core strengths of Uni-Parser. It mainly targets broad non-academic document domains (e.g., presentations and newspapers) and exhibits noticeable annotation noise. Nevertheless, for completeness, we include this evaluation. Leveraging the modular design of Uni-Parser, we replace the layout analysis module with PP-DocLayoutV2~\cite{cui2025paddleocrvlboostingmultilingualdocument} to accommodate this general-document OCR setting. In Table~\ref{tab:omnidocbench}, this variant is denoted as \textit{Uni-Parser-G}.

\begin{table*}[htbp]
\centering
\scriptsize
\setlength{\tabcolsep}{4pt}
\caption{Performance comparison of general document parsing methods on OmniDocBench-1.5~\cite{ouyang2025omnidocbench}.}
\begin{tabular}{l|l|c c c c c c}
\toprule
\textbf{Model Type} & \textbf{Methods} & \textbf{Overall}$\uparrow$ & \textbf{Text\textsuperscript{Edit}}$\downarrow$ & 
\textbf{Formula\textsuperscript{CDM}}$\uparrow$ &
\textbf{Table\textsuperscript{TEDS}}$\uparrow$ &
\textbf{Table\textsuperscript{TEDS-S}}$\uparrow$ &
\textbf{Read Order\textsuperscript{Edit}}$\downarrow$ \\
\midrule

\multirow{4}{*}{\textbf{Pipeline Tools}} 
  & Marker-1.8.2~\cite{marker2023} & 71.30 & 0.206 & 76.66 & 57.88 & 71.17 & 0.250 \\
  & MinerU2-pipeline~\cite{wang2024mineru} & 75.51 & 0.209 & 76.55 & 70.90 & 79.11 & 0.225 \\
  & PP-StructureV3~\cite{cui2025paddleocr} & 86.73 & 0.073 & 85.79 & 81.68 & 89.48 & 0.073 \\
  & \textbf{Uni-Parser-G~(Fast)} & 83.00 & 0.087 & 82.16 & 75.55 & 83.78 & 0.131 \\
  & \textbf{Uni-Parser-G~(HQ)} & 91.33 & 0.043 & 92.23 & 86.06 & 89.42 & 0.058 \\
\midrule

\multirow{5}{*}{\textbf{General VLMs}} 
  & GPT-4o & 75.02 & 0.217 & 79.70 & 67.07 & 76.09 & 0.148 \\
  & InternVL3-76B~\cite{zhu2025internvl3} & 80.33 & 0.131 & 83.42 & 70.64 & 77.74 & 0.113 \\
  & InternVL3.5-241B~\cite{zhu2025internvl3} & 82.67 & 0.142 & 87.23 & 75.00 & 81.28 & 0.125 \\
  & Qwen2.5-VL-72B~\cite{bai2025qwen2} & 87.02 & 0.094 & 88.27 & 82.15 & 86.22 & 0.102 \\
  & Gemini-2.5 Pro & 88.03 & 0.075 & 85.82 & 85.71 & 90.29 & 0.097 \\
\midrule

\multirow{14}{*}{\textbf{Specialized VLMs}}
  & Dolphin~\cite{feng2025dolphin} & 74.67 & 0.125 & 67.85 & 68.70 & 77.77 & 0.124 \\
  & OCRFlux-3B~\cite{ocrflux3b2024} & 74.82 & 0.193 & 68.03 & 75.75 & 80.23 & 0.202 \\
  & Mistral OCR~\cite{mistral_ocr} & 78.83 & 0.164 & 82.84 & 70.03 & 78.04 & 0.144 \\
  & POINTS-Reader~\cite{points-reader} & 80.98 & 0.134 & 79.20 & 77.13 & 81.66 & 0.145 \\
  & olmOCR-7B~\cite{poznanski2025olmocr} & 81.79 & 0.096 & 86.04 & 68.92 & 74.77 & 0.121 \\
  & MinerU2-VLM~\cite{wang2024mineru} & 85.56 & 0.078 & 80.95 & 83.54 & 87.66 & 0.086 \\
  & Nanonets-OCR-s~\cite{Nanonets-OCR-S} & 85.59 & 0.093 & 85.90 & 80.14 & 85.57 & 0.108 \\
  & MonkeyOCR-pro-1.2B~\cite{li2025monkeyocr} & 86.96 & 0.084 & 85.02 & 84.24 & 89.02 & 0.130 \\
  & DeepSeek-OCR (9-crops)~\cite{wei2025deepseek} & 87.36 & 0.073 & 84.14 & 85.25 & 89.01 & 0.085 \\
  & MonkeyOCR-3B~\cite{li2025monkeyocr} & 87.13 & 0.075 & 87.45 & 81.39 & 85.92 & 0.129 \\
  & dots.ocr~\cite{rednote_hilab_dots_ocr_2025} & 88.41 & 0.048 & 83.22 & 86.78 & 90.62 & 0.053 \\
  & MonkeyOCR-pro-3B~\cite{li2025monkeyocr} & 88.85 & 0.075 & 87.25 & 86.78 & 90.63 & 0.128 \\
  & MinerU2.5~\cite{wang2024mineru} & 90.67 & 0.047 & 88.46 & 88.22 & 92.38 & 0.044 \\
  & PaddleOCR-VL~\cite{cui2025paddleocrvlboostingmultilingualdocument} & 92.86 & 0.035 & 91.22 & 90.89 & 94.76 & 0.043 \\

\bottomrule
\end{tabular}
\label{tab:omnidocbench}
\end{table*}

Overall, Uni-Parser demonstrates a clear performance advantage in parsing scientific and patent documents. Its modular design enables adaptation to diverse document types while maintaining competitive accuracy, together with industrial-grade inference speed and robustness, making it well suited for large-scale deployment.

Additional qualitative results are provided in the appendix. Appendix~\ref{appendix:ldcase} highlights Uni-Parser-LD’s more fine-grained layout categorization, stronger group and sub-figure localization, and support for molecular structure localization. Appendix~\ref{appendix:rocase} demonstrates robust reading-order determination under complex layouts via a rule-based strategy, while Appendix~\ref{appendix:case} presents overall comparisons with other PDF parsing frameworks.

\subsection{Chemical Structure Parsing}

Chemical structures play a pivotal role in conveying molecular information, making their accurate recognition crucial for applications ranging from drug discovery and materials design to knowledge extraction from patents and scientific literature. However, most existing document parsing systems provide limited or no support for chemical structure recognition. To address this gap, Uni-Parser integrates two specialized submodels: \textit{Uni-Parser-LD}, which tackles molecular localization within layout detection, and \textit{MolParser 1.5}, which handles chemical structure recognition.

We evaluate Uni-Parser on both the Uni-Parser Benchmark and the recent third-party BioVista benchmark~\cite{yan2025biominer}. As shown in Table~\ref{tab:moldet_eval}, our group-based layout detection model, Uni-Parser-LD, outperforms our previous state-of-the-art molecular detection model, MolDet~\cite{fang2024molparser}, while also establishing correspondences between molecules and their identifiers—a capability absent in MolDet. Table~\ref{tab:ocsr_eval} shows that MolParser 1.5 consistently surpasses its predecessor across all benchmark subsets, delivering notable improvements in both accuracy and inference speed over mainstream open-source OCSR methods. Moreover, our approach remains highly competitive against costly commercial models~\cite{MolVision_2025,xiong2023alphaextractor}. Representative case studies are provided in Appendix~\ref{appendix:molparser}.

\begin{table*}[htbp]
\centering
\renewcommand{\arraystretch}{1.2}
\setlength{\tabcolsep}{2.8pt}
\caption{Comparison of chemical structure detection and recognition performance.}
\footnotesize
\begin{subtable}[t]{0.48\textwidth}
\centering
\caption{Molecule location performance}
\begin{tabular}{lcc}
\toprule
\textbf{Model} & \textbf{mAP@50} & \textbf{mAP@50-95} \\
\midrule
\multicolumn{3}{c}{\textit{Uni-Parser Benchmark}} \\
\midrule
\textbf{Uni-Parser-LD}      & \textbf{0.994} & \textbf{0.968} \\
MolDet-Doc-L\cite{fang2024molparser}      & 0.983 & 0.919 \\
MolDet-General-L\cite{fang2024molparser}  & 0.974 & 0.815 \\
\midrule
\multicolumn{3}{c}{\textit{BioVista Benchmark}} \\
\midrule
\textbf{Uni-Parser-LD}         & \textbf{0.981} & 0.844 \\
MolDet-Doc-L\cite{fang2024molparser}      & 0.961 & \textbf{0.871} \\
MolDet-General-L\cite{fang2024molparser}  & 0.945 & 0.815 \\
BioMiner\cite{yan2025biominer}           & 0.929 & - \\
MolMiner\cite{molminer}           & 0.899 & - \\
\bottomrule
\end{tabular}
\label{tab:moldet_eval}
\end{subtable}%
\hfill
\begin{subtable}[t]{0.5\textwidth}
\centering
\caption{OCSR accuracy for different types of molecules}
\begin{tabular}{lcccc}
\toprule
\textbf{Model} & \textbf{Full} & \textbf{Chiral} & \textbf{Markush} & \textbf{All} \\
\midrule
\multicolumn{5}{c}{\textit{Uni-Parser Benchmark}} \\
\midrule
\textbf{MolParser~1.5} & \textbf{0.979} & \textbf{0.809} & \textbf{0.805} & \textbf{0.886} \\
MolParser~1.0\cite{fang2024molparser} & 0.953 & 0.676 & 0.664 & 0.800  \\
MolScribe\cite{qian2023molscribe}  & 0.617 & 0.274 & 0.168 & 0.417 \\
\midrule
\multicolumn{5}{c}{\textit{BioVista Benchmark}} \\
\midrule
\textbf{MolParser~1.5} & \textbf{0.800} & \textbf{0.622} & \textbf{0.769} & \textbf{0.786} \\
MolParser~1.0\cite{fang2024molparser} & 0.669 & 0.352 & 0.733 & 0.703 \\
MolMiner\cite{molminer}      & 0.774 & 0.497 & 0.185 & 0.507 \\
MolScribe\cite{qian2023molscribe}     & 0.703 & 0.481 & 0.156 & 0.455 \\
MolNexTR\cite{chen2024molnextr}      & 0.695 & 0.419 & 0.045 & 0.401 \\
DECIMER\cite{rajan2023decimer}       & 0.545 & 0.326 & 0.000 & 0.298 \\
\bottomrule
\end{tabular}
\label{tab:ocsr_eval}
\end{subtable}
\label{tab:chem_eval}
\end{table*}

To further evaluate the capability of existing PDF parsing frameworks in handling chemical structures, we conduct a controlled comparison on a small test set of 141 simple molecules (excluding Markush structures). Since batch access to competing systems is not feasible, we restrict the evaluation to this limited set. We compare four key metrics: (i) molecule localization recall, (ii) the proportion of successfully parsed molecules (OCSR success rate), (iii) the final recognition accuracy, and (iv) molecule-identifier matching rate. As shown in Table~\ref{tab:pdf_eval_ocsr}, Uni-Parser achieves consistently strong performance across all metrics, whereas existing PDF parsing frameworks that claim to support chemical structure recognition perform significantly worse on this task.

\begin{table}[htbp]
\small
\centering
\caption{Comparison of PDF parsing systems with OCSR support on a tiny test set, provided purely as a qualitative analysis rather than a quantitative evaluation.}
\label{tab:pdf_eval_ocsr}
\renewcommand{\arraystretch}{1.1}
\setlength{\tabcolsep}{6pt}
\begin{tabular}{lcccccc}
\toprule
\textbf{Method} & \textbf{Test Date} & \textbf{Recall $\uparrow$} & \textbf{OCSR Success $\uparrow$} & \textbf{OCSR Acc $\uparrow$} & \textbf{Id Match $\uparrow$} & \textbf{Time $\downarrow$} \\
\midrule
\textbf{Uni-Parser}   & 2025-08-27 & 100\% & 100\% & 96.5\% & 100\% & 1.8 s \\
MathPix~\cite{mathpix}      & 2025-08-27 & 100\% & 75.9\% & 59.6\% & - & 66.1 s \\
MinerU.Chem~\cite{MinerU.Chem_2025} & 2025-09-17 & 66.7\% & 63.1\%  & 22.7\% & - & $\sim$7 min \\
\bottomrule
\end{tabular}
\end{table}

%% file: sec/7_downstream.tex
\section{Applications}

Uni-Parser opens up a wide range of downstream applications, spanning literature understanding, structured knowledge extraction, patent analysis, and large-scale data generation. Together, these applications demonstrate the framework’s versatility in advancing both scientific research and industrial practice.  

\textbf{Document Understanding.}  
Uni-Parser significantly enhances scientific literature workflows by supporting automatic document summarization, document-level question answering~\cite{cai2024uni}, paper-to-poster~\cite{pang2025paper2poster} and paper-to-PPT~\cite{dataflow-agent2025} generation, as well as intelligent retrieval and deep research~\cite{openai2025deepresearch}. Collectively, these capabilities streamline both knowledge consumption and dissemination for researchers across diverse disciplines.

\textbf{Structured Data Extraction.}  
By converting unstructured documents into structured representations, Uni-Parser supports the large-scale construction of domain-specific databases, such as paper citation database, scholar database, molecular libraries, reaction repositories, bioactivity database \cite{yan2025biominer}, experimental characterization database \cite{wang2025uniem3muniversalelectronmicrograph, xu2025toward, wang2025nmrexp}, and comprehensive entity knowledge bases. Such resources are critical for accelerating data-driven scientific discovery. 

\textbf{Patent Retrieval and Protection.}  
The framework further facilitates patent retrieval and prior-art verification, providing robust support for innovation discovery and intellectual property protection \cite{shi2024intelligent, zhuang2025doc2sar}. This enables more efficient navigation of complex patent landscapes in scientific and industrial settings.  

\textbf{Foundation Model Training.}  
Finally, Uni-Parser can serve as a powerful engine for large-scale training data generation, supplying high-quality structured inputs that reduce the cost and effort of manual curation. This capability is particularly valuable for advancing foundation models in scientific domains, where large, reliable datasets are indispensable \cite{li2024scilitllm, liao2025innovator, dataflow2025}.

%% file: sec/8_experience.tex
\section{Failed Approaches}

We summarize things that didn't work during the development of Uni-Parser.

In most scenarios of document intelligence, large-scale pretraining on synthetic data is a simple and effective strategy, because natural documents are largely computer-rendered and therefore exhibit strong domain consistency. However, as discussed in Section~\ref{subsec:layout}, this intuition breaks down for layout recognition in scientific literature. The layouts of scientific articles are shaped extensively by human editorial practices and creative design choices, making them difficult for heuristic layout engines to reproduce. As a result, synthetic data not only fails to cover the diversity of real-world layouts but can even distort the true data distribution.

For the OCSR task, atom–bond (graph-based) methods are intuitively appealing and have a long history of successful applications; they were also the first direction we explored. While these methods offer clear advantages for handling chirality, they struggle with the wide variety of challenging cases present in real scientific literature. Their strong reliance on rigid, hand-crafted rules fundamentally limits scalability—simply increasing training data provides little benefit. In addition, these methods require substantially more manual annotation effort, typically over 20× that of end-to-end approaches, further constraining their practicality. As a result, compared with end-to-end models, graph-based methods suffer from lower performance ceilings, slower inference, and prohibitively high annotation costs.

%% file: sec/9_summary.tex
\section{Future Work}

\textbf{Enhancing Core Components.}
Uni-Parser's distributed and modular pipeline architecture allows individual components to be easily upgraded or replaced, facilitating continuous improvement across different document types.  We plan to iteratively update the core components to further improve extraction quality across diverse document types:
\begin{itemize}
    \item \textbf{Layout detection:} Our current models are primarily tailored to scientific and patent documents. However, the diversity of document types and layouts is virtually limitless. We will continue to enhance our layout recognition models to support an increasingly broader range of scenarios, including newspapers and magazines, PPT slides, various book formats, and financial statements.
    \item \textbf{OCSR model:} Although our MolParser 1.5 already outperforms previous state-of-the-art methods, the recognition of chiral molecules still presents significant challenges. We will focus on exploring how to address the challenge of chirality recognition within end-to-end OCSR models.
    \item \textbf{Chemical reaction understanding:} Parsing chemical reactions in real-world literature remains highly challenging, with substantial room for improving generalization performance.
    \item \textbf{Chart understanding:} Currently, all existing Chart2Table models and general-purpose MLLMs fall far short of meeting industrial-level requirements for parsing charts in scientific literature, which exhibit a wide variety of types and styles. Chart parsing therefore still holds substantial room for further exploration.
    \item \textbf{Reading order:} We plan to incorporate machine learning–based reading order predictors, to enhance the generalization ability of reading order prediction under complex layouts.
    \item \textbf{Deployment optimization:} Techniques such as quantization (PTQ and QAT), distillation, pruning, and other inference acceleration methods will be explored, along with support for diverse hardware platforms, including Ascend NPUs.
\end{itemize}

\textbf{Uni-Parser-Tools for Easy Access.}
We will release an open-source toolkit, \textit{Uni-Parser-Tools}, providing remote access to Uni-Parser without requiring local compute. The toolkit includes example pipelines for downstream tasks, enabling rapid construction of structured scientific databases and supporting generative AI applications for scientific discovery.

\textbf{Benchmark Construction.}
Existing benchmarks (e.g., OmniDocBench and the Uni-Parser benchmark) are limited by inconsistent layout annotations and heterogeneous OCR outputs, and fail to adequately capture complex layouts and cross-page content. We plan to explore more robust, task-driven benchmarks tailored to downstream applications such as document understanding and structured data extraction, enabling fairer and more informative evaluations.

%% file: appendix/authors.tex
\section{Authorship and Acknowledgments}

Please cite this work as “DP Technology (2025)”. 

Correspondence regarding this technical report can be sent to fangxi@dp.tech \\

\begin{multicols}{2}
\raggedcolumns

\textbf{Principal Contributor} \\
Haoyi Tao \\

\textbf{Core Contributors}\footnotemark[1] \\
Chaozheng Huang \\
Han Lyu \\
Haocheng Lu \\
Junjie Wang \\
Shuwen Yang \\
Suyang Zhong \\
Xi Fang \\

\textbf{Contributors \& Acknowledgments}\footnotemark[1] \\
Changhong Chen \\
Chenkai Wu \\
Fanjie Xu \\
Hanzheng Li \\
Hengxin Cai \\
Jialu Shen \\
Jian Zhou \\
Jiankun Wang \\
Jiaxi Zhuang \\
Jinbo Hu \\
Jindi Guo \\
Jingwen Deng \\
Lin Yao \\
Mingjun Xu \\
Nan Wang \\
Ning Wang \\
Qingguo Zhou \\
Qiushi Huang \\
Shang Xiang \\
Shangqian Chen \\
Shaojie Chen \\
Shengyu Li \\
Xiaochen Cai \\
Xiaohong Ji \\
Xinyu Xiong \\
Xuan Xie \\
Yanhui Hong \\
Yaorui Shi \\
Yaqi Li \\
Yixuan Li \\
Yuan Gao \\
Zhenhao Wong \\
Zhifeng Gao \\
Zhiyuan Chang \\
Zhiyue Wang \\

\columnbreak


\textbf{Project Lead} \\
Xi Fang \\

\textbf{Team Management} \\
Guolin Ke \\
Linfeng Zhang \\
Xinyu Li \\

\end{multicols}

\footnotetext[1]{Contributors listed in alphabetized order.}

%% file: appendix/examples.tex
\section{Examples and Comparisons}
\subsection{MolParser-1.5 Case Study}
\label{appendix:molparser}
\footnotetext[1]{The evaluation was conducted on August 27, 2025.}

\begin{figure}[!htpb]
  \centering
  \includegraphics[width=\linewidth]{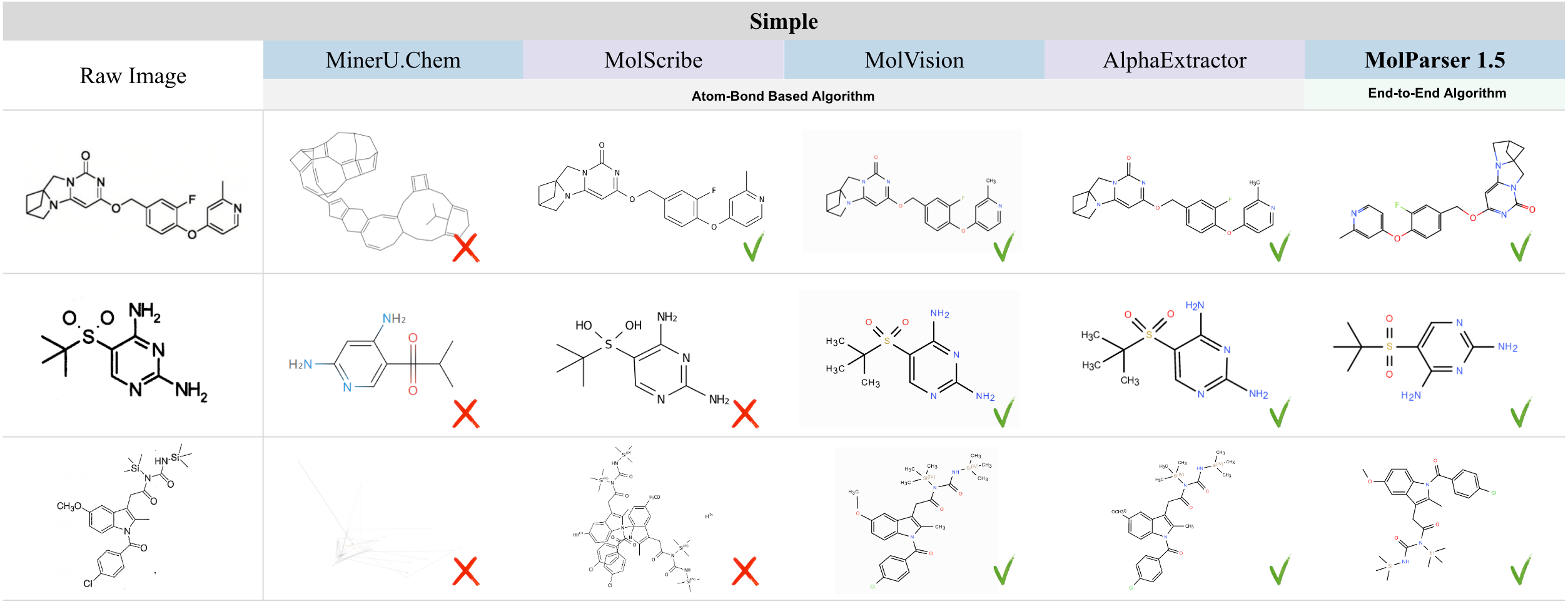}\vspace{6pt}
  \includegraphics[width=\linewidth]{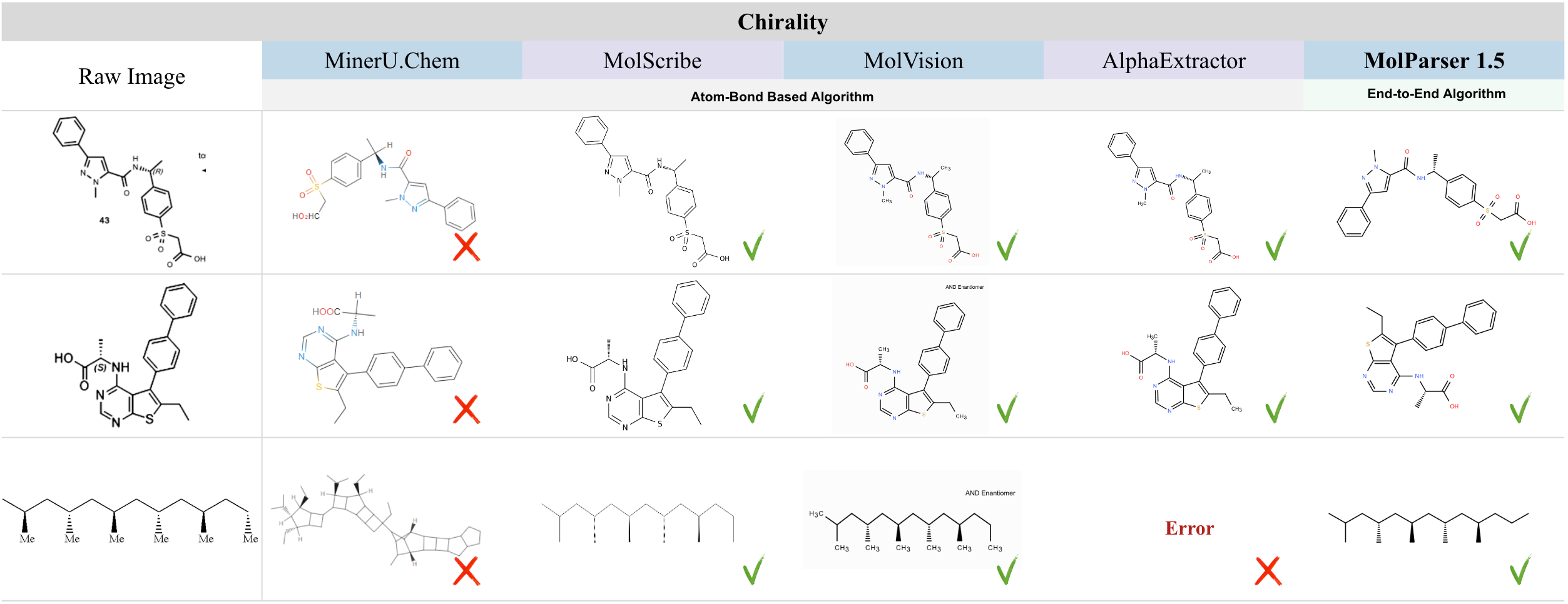}\vspace{6pt}
  \includegraphics[width=\linewidth]{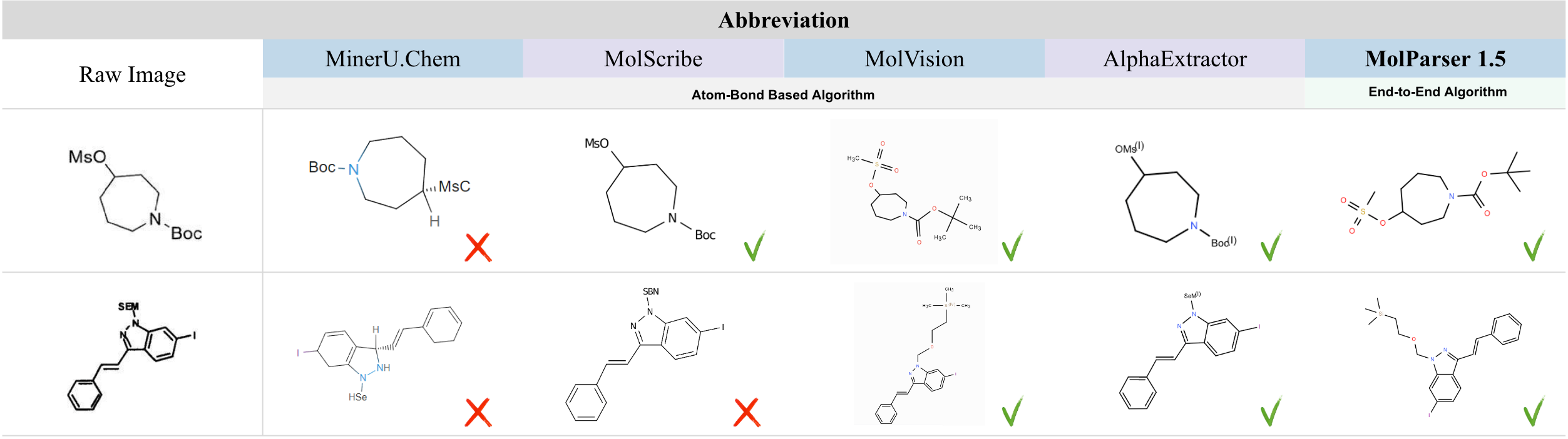}\vspace{6pt}
  \includegraphics[width=\linewidth]{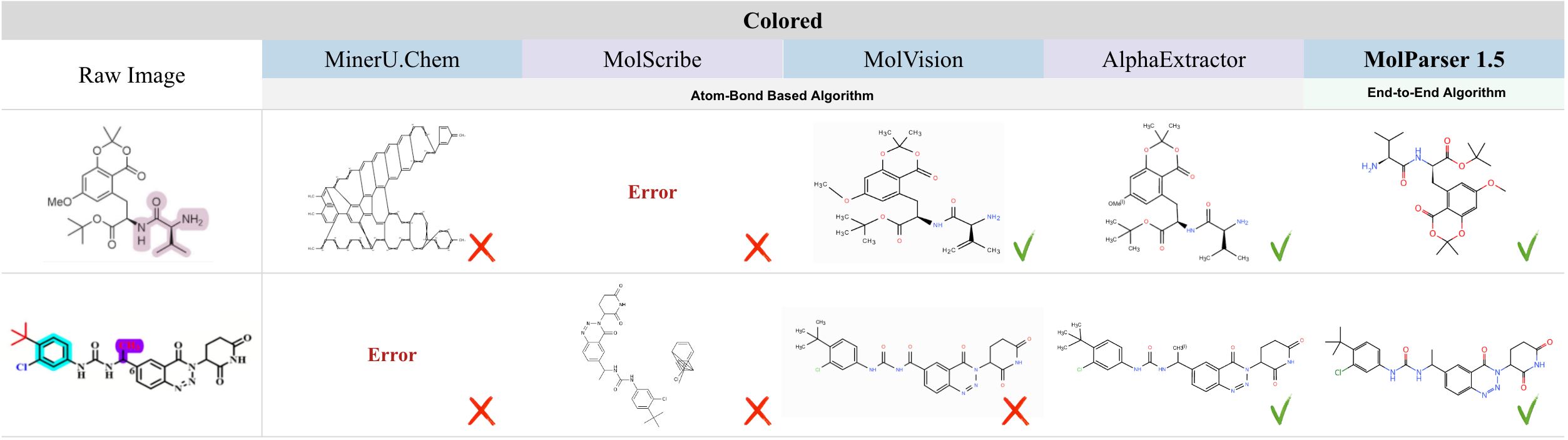}
\end{figure}

\newpage

\begin{figure}[!htpb]
  \centering
  \includegraphics[width=\linewidth]{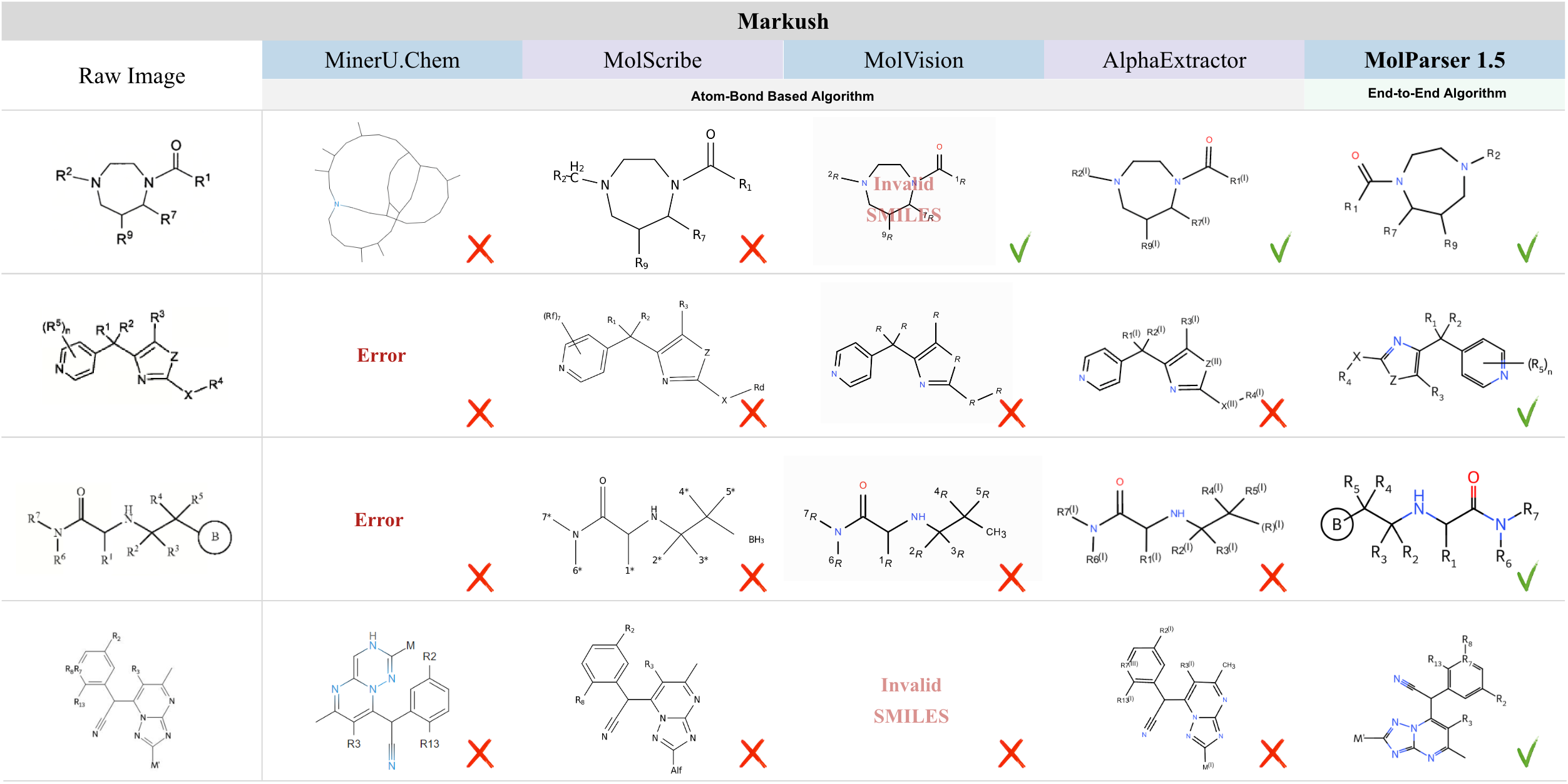}\vspace{6pt}
  \includegraphics[width=\linewidth]{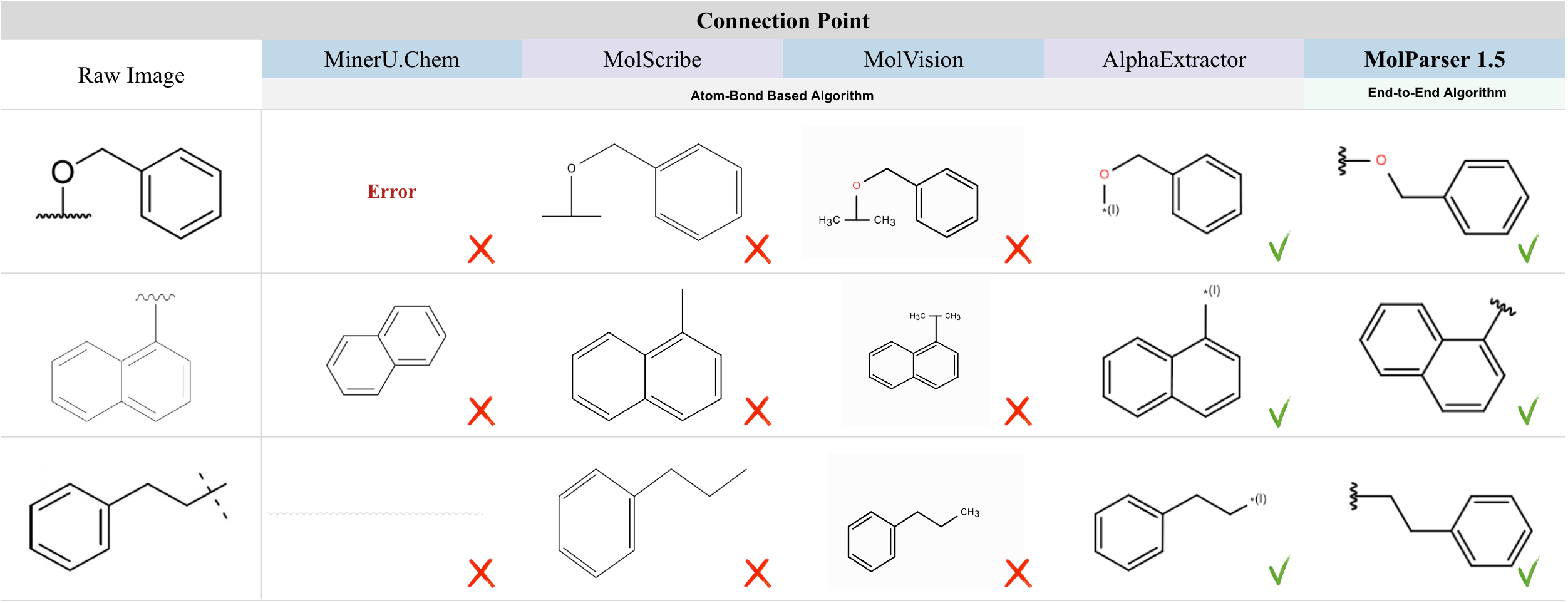}\vspace{-4pt}
  \includegraphics[width=\linewidth]{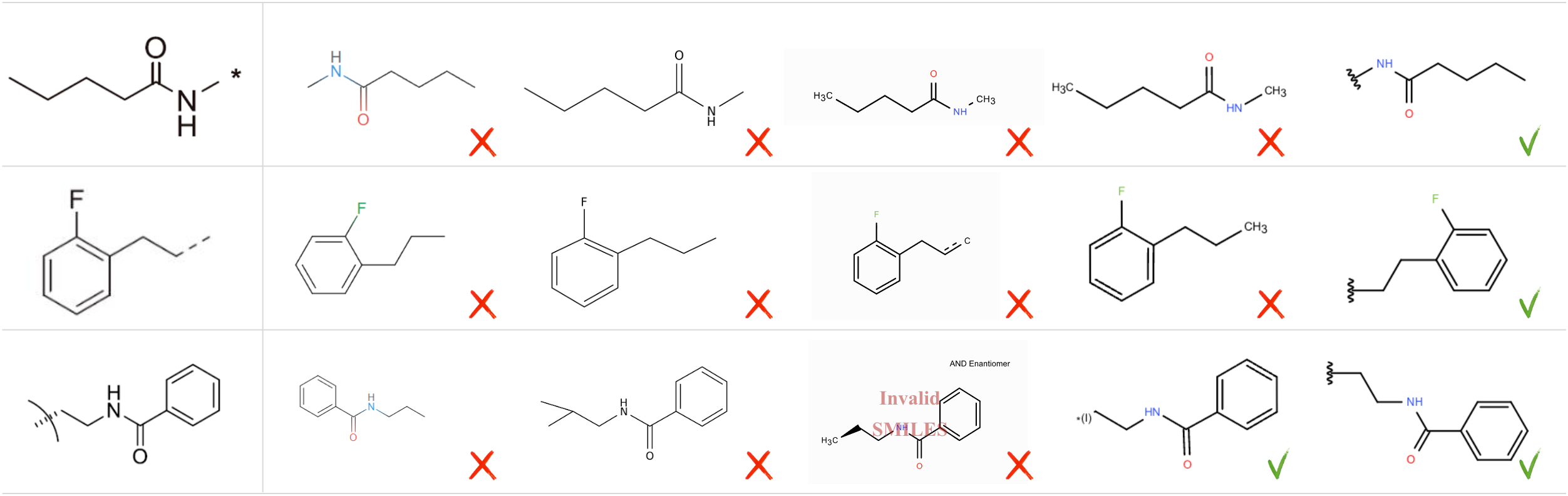}\vspace{6pt}
  \includegraphics[width=\linewidth]{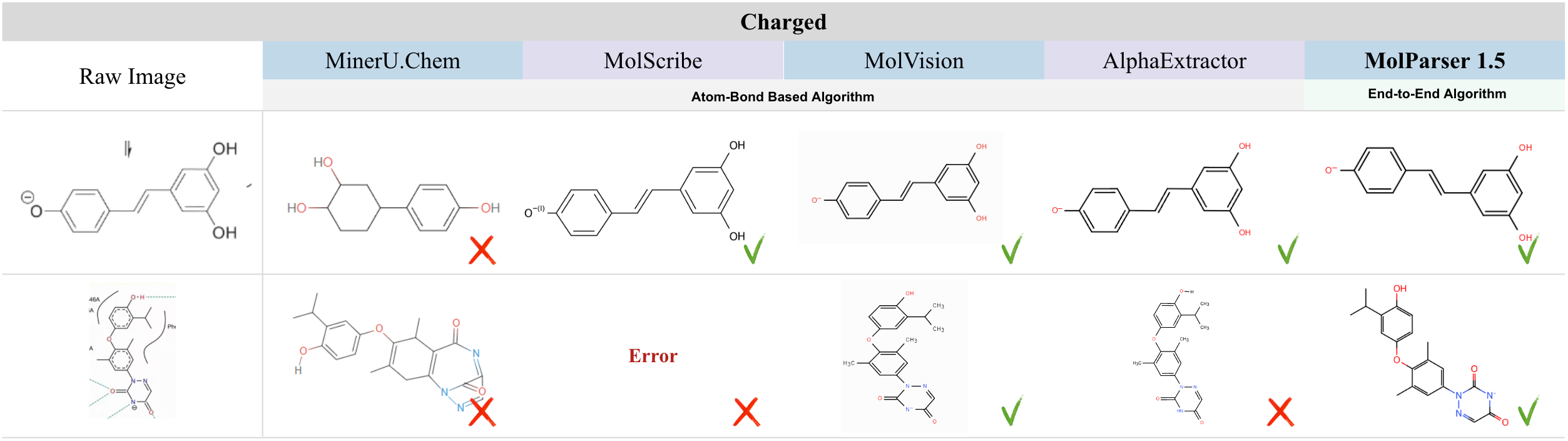}
\end{figure}

\newpage

\begin{figure}[!htpb]
  \centering
  \includegraphics[width=\linewidth]{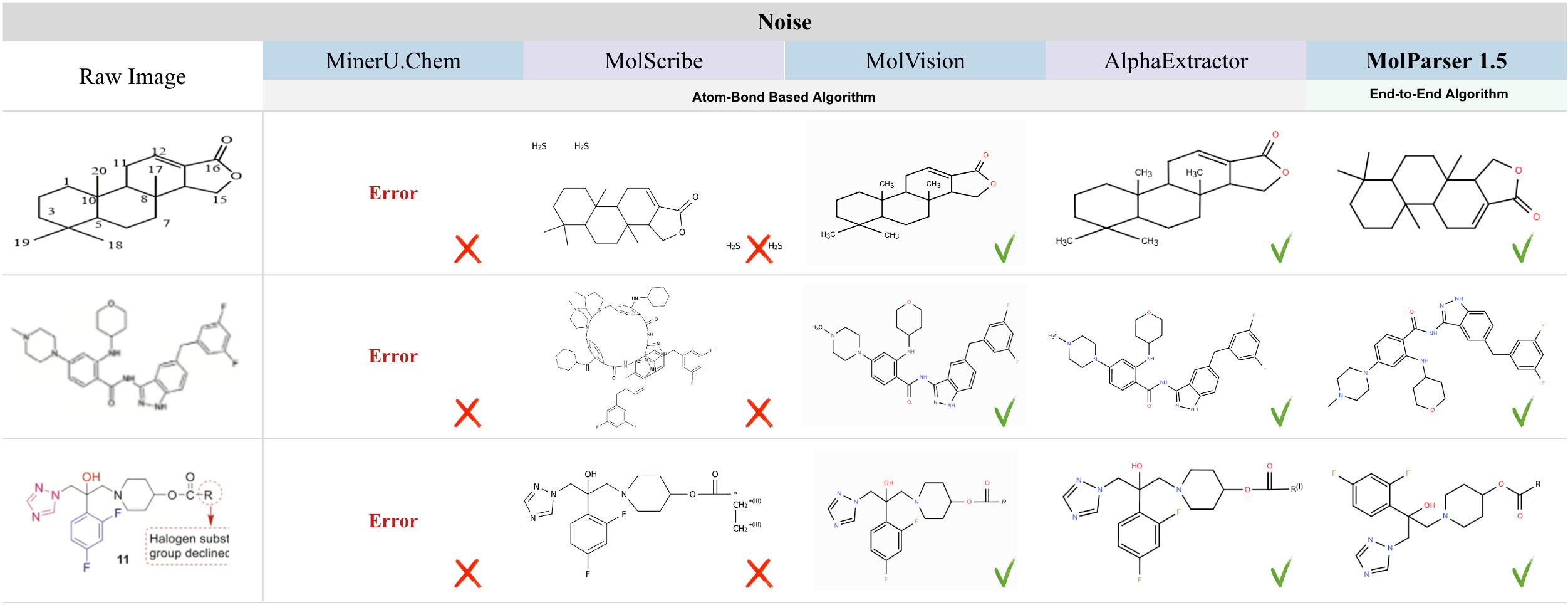}
\end{figure}

\subsection{Qualitative Examples of Cross-Page Layout Grouping}
\label{appendix:cross_example}

\begin{figure}[!htpb]
  \centering
  \includegraphics[width=0.8\linewidth]{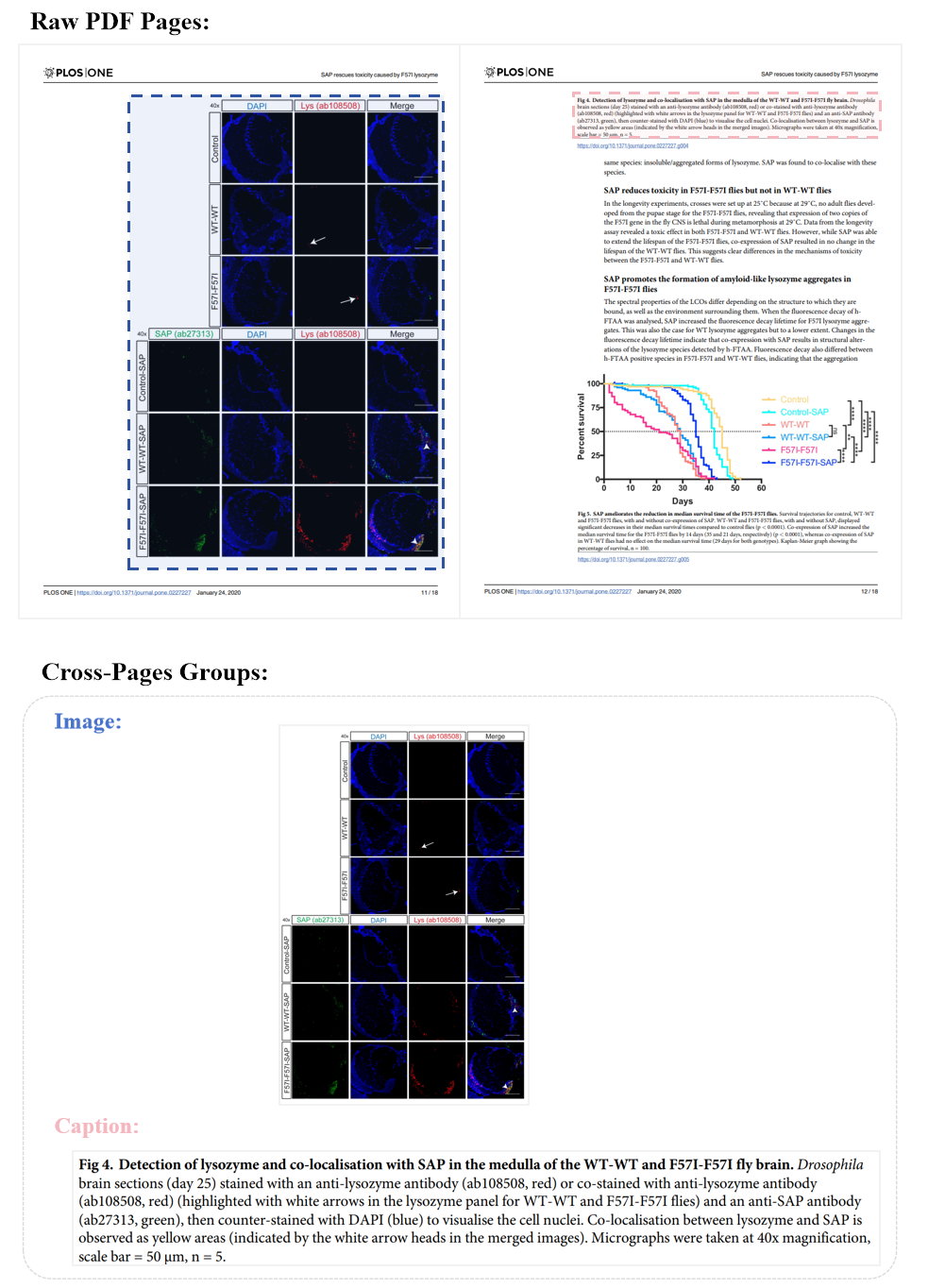}
\end{figure}

\newpage


\subsection{Layout Detection Comparison}
\label{appendix:ldcase}

\tikzset{
  gridline/.style={
    black!40,
    dashed,        
    line width=0.5pt
  },
  cellborder/.style={
    gray,
    solid,
    line width=0.8pt
  }
}

\newlength{\CellWidth}
\newlength{\CellHeight}

\setlength{\CellWidth}{0.49\textwidth}
\setlength{\CellHeight}{0.43\textheight}

\newlength{\ImageMaxHeight}
\setlength{\ImageMaxHeight}{0.90\CellHeight}

\definecolor{uniparser-color}{rgb}{0.0392157, 0.23470588, 0.45372549}

\newcommand{\ShowcaseCellOldBackup}[3]{%
\begin{minipage}[c][\CellHeight][c]{\CellWidth}
\centering

\begin{tikzpicture}
\node (img) {
    \adjustbox{
        max width=\linewidth,
        max height=\ImageMaxHeight,
        keepaspectratio,
        center
    }{\includegraphics{#1}}
};

\node[
    anchor=north west,
    fill=#3,
    text=white,
    font=\small,
    inner sep=1pt
] at (img.north west) {#2};
\end{tikzpicture}

\end{minipage}
}

\newcommand{\ShowcaseCell}[3]{%
\begin{minipage}[c][\CellHeight][c]{\CellWidth}
\centering

\begin{tikzpicture}[x=1pt,y=1pt]
    \useasboundingbox (0,0) rectangle (\CellWidth,\CellHeight);

    \node at (0.5\CellWidth,0.5\CellHeight) {
        \adjustbox{
            max width=\CellWidth,
            max height=\ImageMaxHeight,
            keepaspectratio,
            center
        }{\includegraphics{#1}}
    };

    \node[
        anchor=north west,
        fill=#3,
        text=white,
        font=\small,
        inner sep=2pt
    ] at (0,\CellHeight) {#2};

\end{tikzpicture}

\end{minipage}
}

\newcommand{\ShowcaseCellBoxed}[3]{%
\begin{minipage}[c][\CellHeight][c]{\CellWidth}
\centering

\begin{tikzpicture}[x=1pt,y=1pt]
    \useasboundingbox (0,0) rectangle (\CellWidth,\CellHeight);

    \draw[cellborder] (0,0) rectangle (\CellWidth,\CellHeight);

    \node at (0.5\CellWidth,0.5\CellHeight) {
        \adjustbox{
            max width=\CellWidth-6pt,
            max height=\ImageMaxHeight-6pt,
            keepaspectratio,
            center
        }{\includegraphics{#1}}
    };

    \node[
        anchor=north west,
        fill=#3,
        text=white,
        font=\small\bfseries,
        inner sep=2pt,
        outer sep=-4pt  
    ] at (0,\CellHeight) {#2};
    

\end{tikzpicture}

\end{minipage}
}




\pgfkeys{
    /showcasecell/.is family, /showcasecell,
    label/.store in = \scLabel,
    label/.default = {},
    bgcolor/.store in = \scBgColor,
    bgcolor/.default = blue,
    clipleft/.store in = \scClipLeft,
    clipleft/.default = 0pt,
    clipright/.store in = \scClipRight,
    clipright/.default = 0pt,
    cliptop/.store in = \scClipTop,
    cliptop/.default = 0pt,
    clipbottom/.store in = \scClipBottom,
    clipbottom/.default = 0pt,
    clip/.style = {
        clipleft=#1,
        clipright=#1,
        cliptop=#1,
        clipbottom=#1
    },
    cliph/.style = {clipleft=#1, clipright=#1},
    clipv/.style = {cliptop=#1, clipbottom=#1},
    mode/.store in = \scMode,
    mode/.default = fit, 
    align/.store in = \scAlign,
    align/.default = center, 
}

\makeatletter 

\NewDocumentCommand{\@ApplyClip}{m m m}{%
    \def\fitmode{fit}%
    \def\cropmode{crop}%
    \def\fillmode{fill}%
    \def\stretchmode{stretch}%
    
    \pgfmathsetlength{\effectivewidth}{#1 - \scClipLeft - \scClipRight}%
    \pgfmathsetlength{\effectiveheight}{#2 - \scClipTop - \scClipBottom}%
    
    \ifx\scMode\fitmode
        \adjustbox{
            max width=\effectivewidth,
            max height=\effectiveheight,
            keepaspectratio,
            center
        }{\includegraphics{#3}}%
    \fi
    \ifx\scMode\stretchmode
        \includegraphics[width=\effectivewidth,height=\effectiveheight,keepaspectratio=false]{#3}%
    \fi
    \ifx\scMode\fillmode
        \begin{adjustbox}{
            minipage=[c][\effectiveheight][c]{\effectivewidth},
            valign=c, halign=c
        }%
        \adjustbox{
            min size={\effectivewidth}{\effectiveheight},
            keepaspectratio,
            center
        }{\includegraphics{#3}}%
        \end{adjustbox}%
    \fi
    \ifx\scMode\cropmode
        \pgfmathsetlength{\@trimleft}{max(0pt, -\scClipLeft)}%
        \pgfmathsetlength{\@trimright}{max(0pt, -\scClipRight)}%
        \pgfmathsetlength{\@trimtop}{max(0pt, -\scClipTop)}%
        \pgfmathsetlength{\@trimbottom}{max(0pt, -\scClipBottom)}%
        
        \ifdim\@trimleft=0pt
            \ifdim\@trimright=0pt
                \ifdim\@trimtop=0pt
                    \ifdim\@trimbottom=0pt
                        \adjustbox{
                            max width=\effectivewidth,
                            max height=\effectiveheight,
                            keepaspectratio,
                            center
                        }{\includegraphics{#3}}%
                    \else
                        \@DoTrim{#3}%
                    \fi
                \else
                    \@DoTrim{#3}%
                \fi
            \else
                \@DoTrim{#3}%
            \fi
        \else
            \@DoTrim{#3}%
        \fi
    \fi
}

\NewDocumentCommand{\@DoTrim}{m}{%
    \edef\@trimparams{\@trimleft\space \@trimbottom\space \@trimright\space \@trimtop}%
    \begin{adjustbox}{
        trim=\@trimparams,
        clip,
        center
    }{\includegraphics{#1}}%
    \end{adjustbox}%
}

\NewDocumentCommand{\ShowcaseCellBoxedCrop}{O{} m}{%
    \pgfkeys{/showcasecell/.cd, 
        label={}, bgcolor=blue, 
        clipleft=0pt, clipright=0pt, cliptop=0pt, clipbottom=0pt,
        mode=fit, align=center,
        #1 
    }%
    
    \begin{minipage}[c][\CellHeight][c]{\CellWidth}
    \centering
    
    \begin{tikzpicture}[x=1pt,y=1pt]
        \useasboundingbox (0,0) rectangle (\CellWidth,\CellHeight);
        
        \draw[cellborder] (0,0) rectangle (\CellWidth,\CellHeight);
        
        \pgfmathsetlength{\@availwidth}{\CellWidth-6pt}%
        \pgfmathsetlength{\@availheight}{\ImageMaxHeight-6pt}%
        
        \node[inner sep=0pt] at (0.5\CellWidth,0.5\CellHeight) {
            \@ApplyClip{\@availwidth}{\@availheight}{#2}
        };
        
        \node[
            anchor=north west,
            fill=\scBgColor,
            text=white,
            font=\small\bfseries,
            inner sep=2pt,
            outer sep=-4pt
        ] at (0,\CellHeight) {\scLabel};
        
    \end{tikzpicture}
    
    \end{minipage}
}

\begin{figure}[htpb]
\centering

\begin{tikzpicture}[remember picture]

\node (grid)[inner sep=0pt] {
\begin{minipage}{\textwidth}
\centering

\ShowcaseCellBoxed{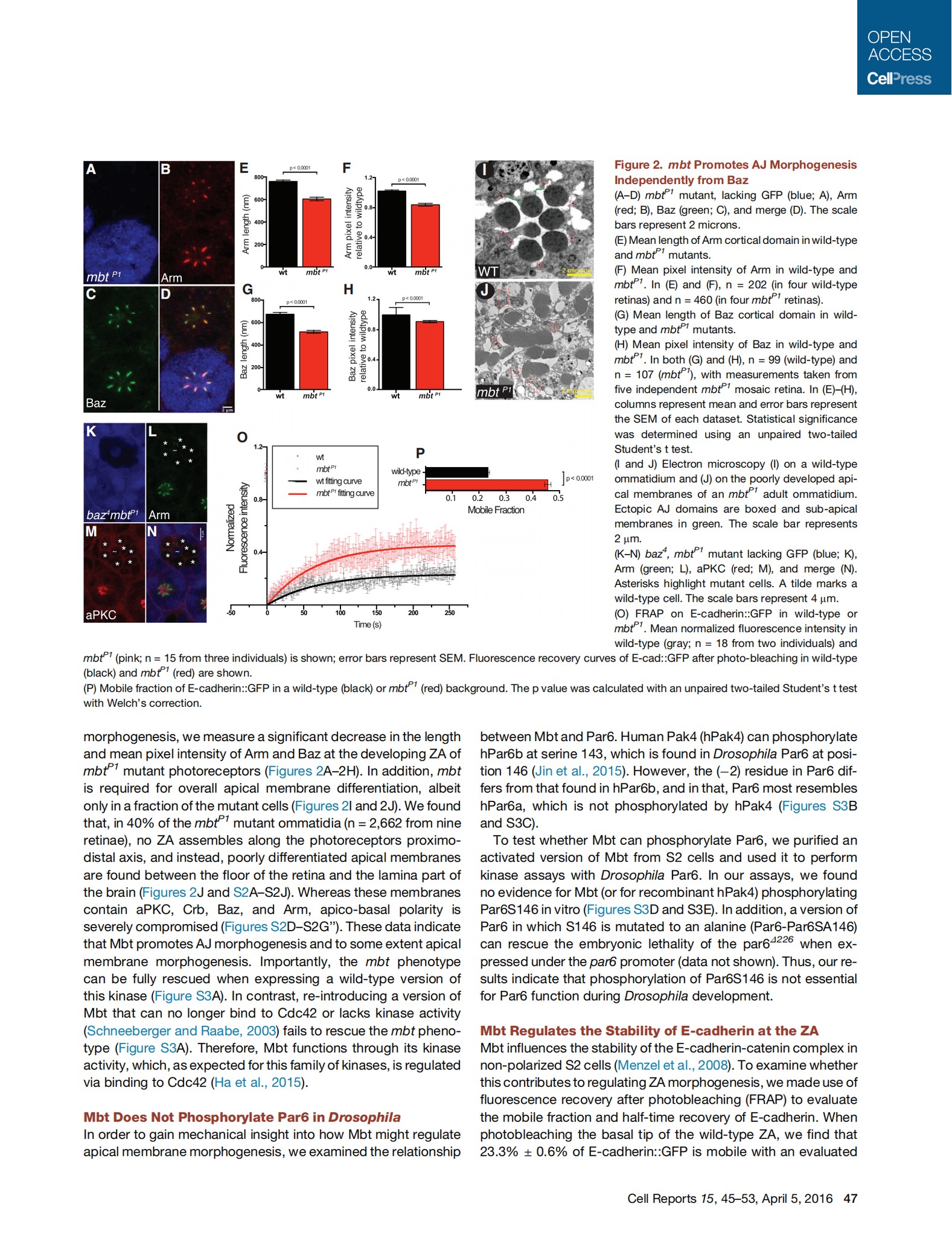}{Original Document}{black!80}
\hfill
\ShowcaseCellBoxed{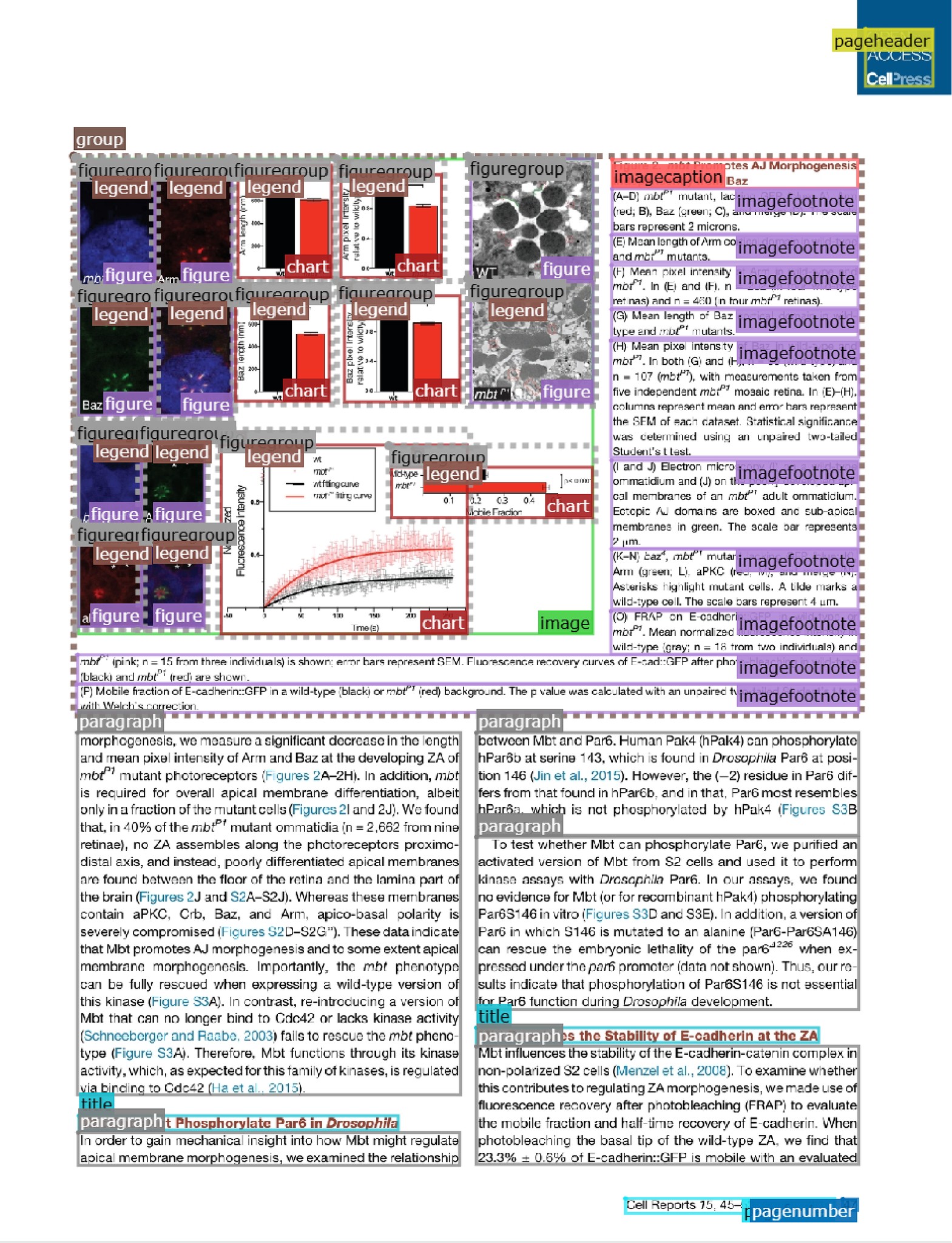}{Uni-Parser}{uniparser-color}

\vspace{12pt} 

\ShowcaseCellBoxed{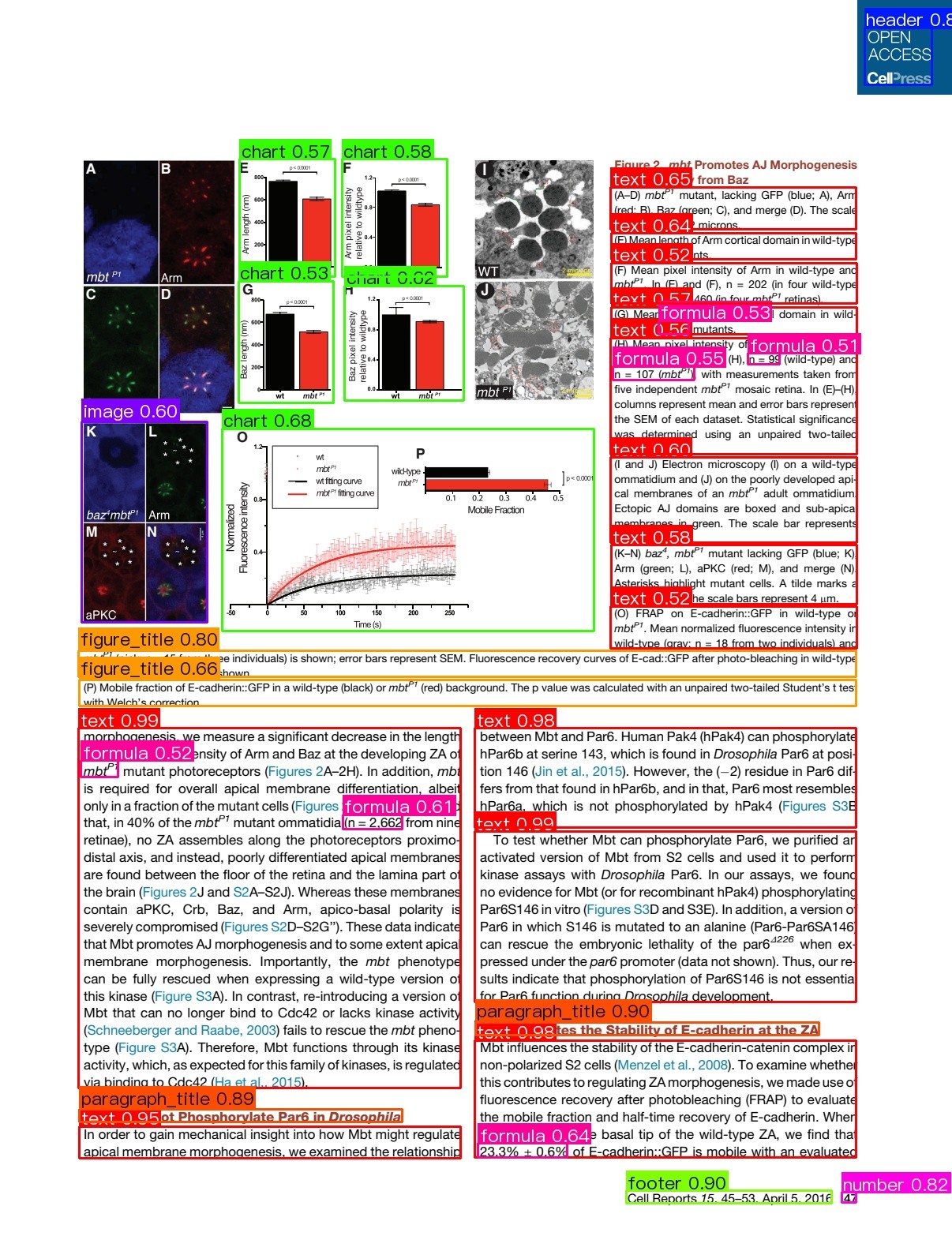}{PP-StructureV3}{gray!80}
\hfill
\ShowcaseCellBoxed{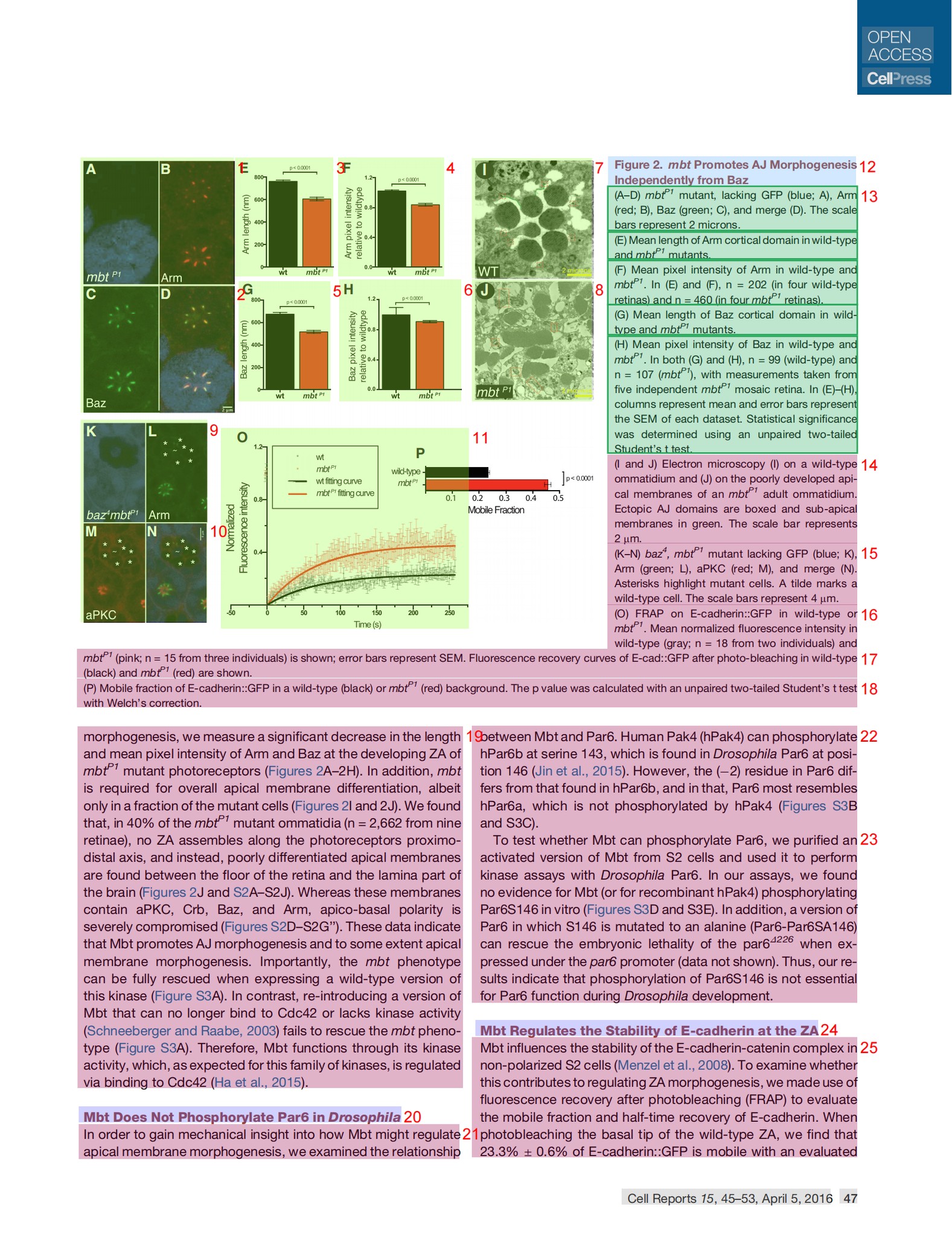}{MinerU2.5}{gray!80}

\end{minipage}
};



\end{tikzpicture}

\end{figure}


\begin{figure}[htpb]
\centering

\begin{tikzpicture}[remember picture]

\node (grid)[inner sep=0pt] {
\begin{minipage}{\textwidth}
\centering

\ShowcaseCellBoxed{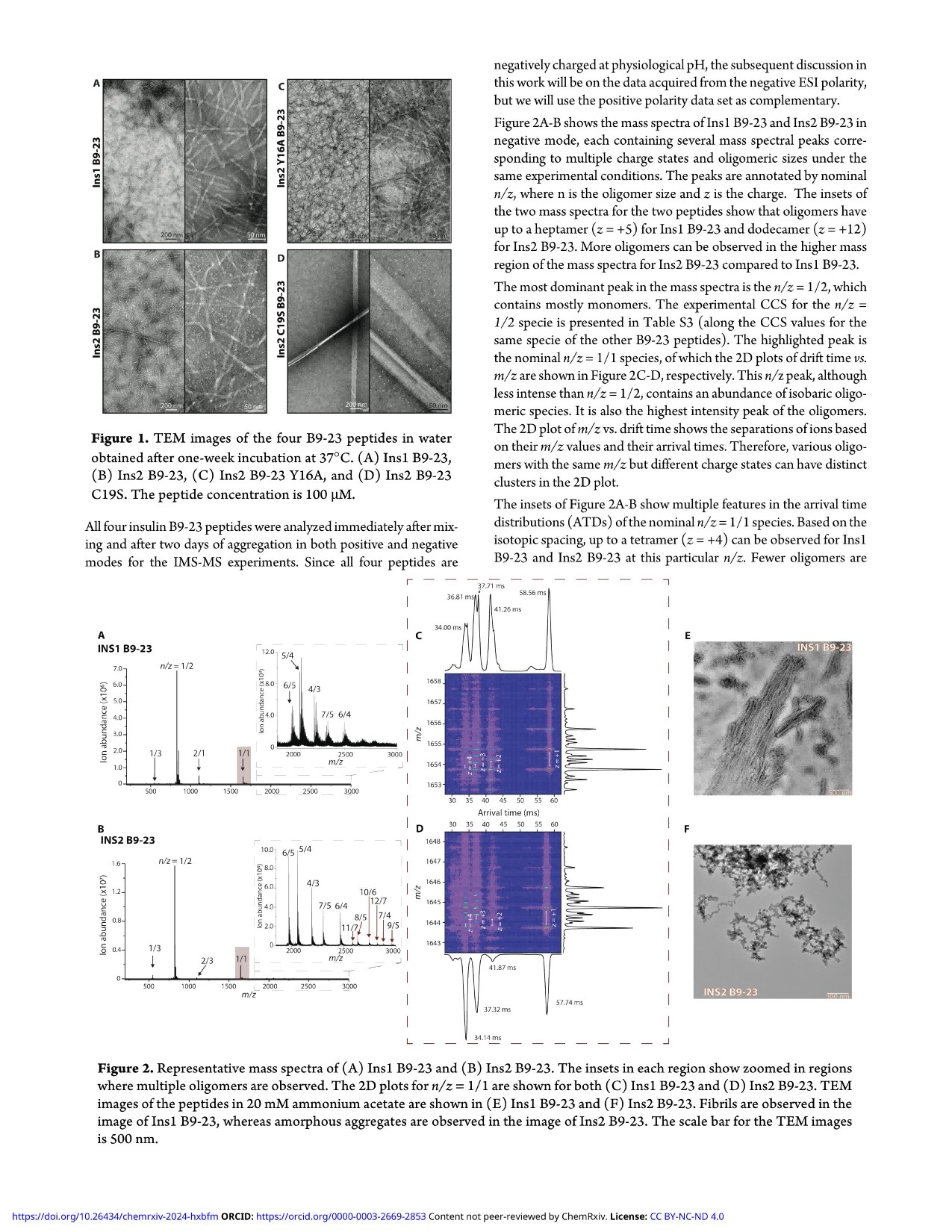}{Original Document}{black!80}
\hfill
\ShowcaseCellBoxed{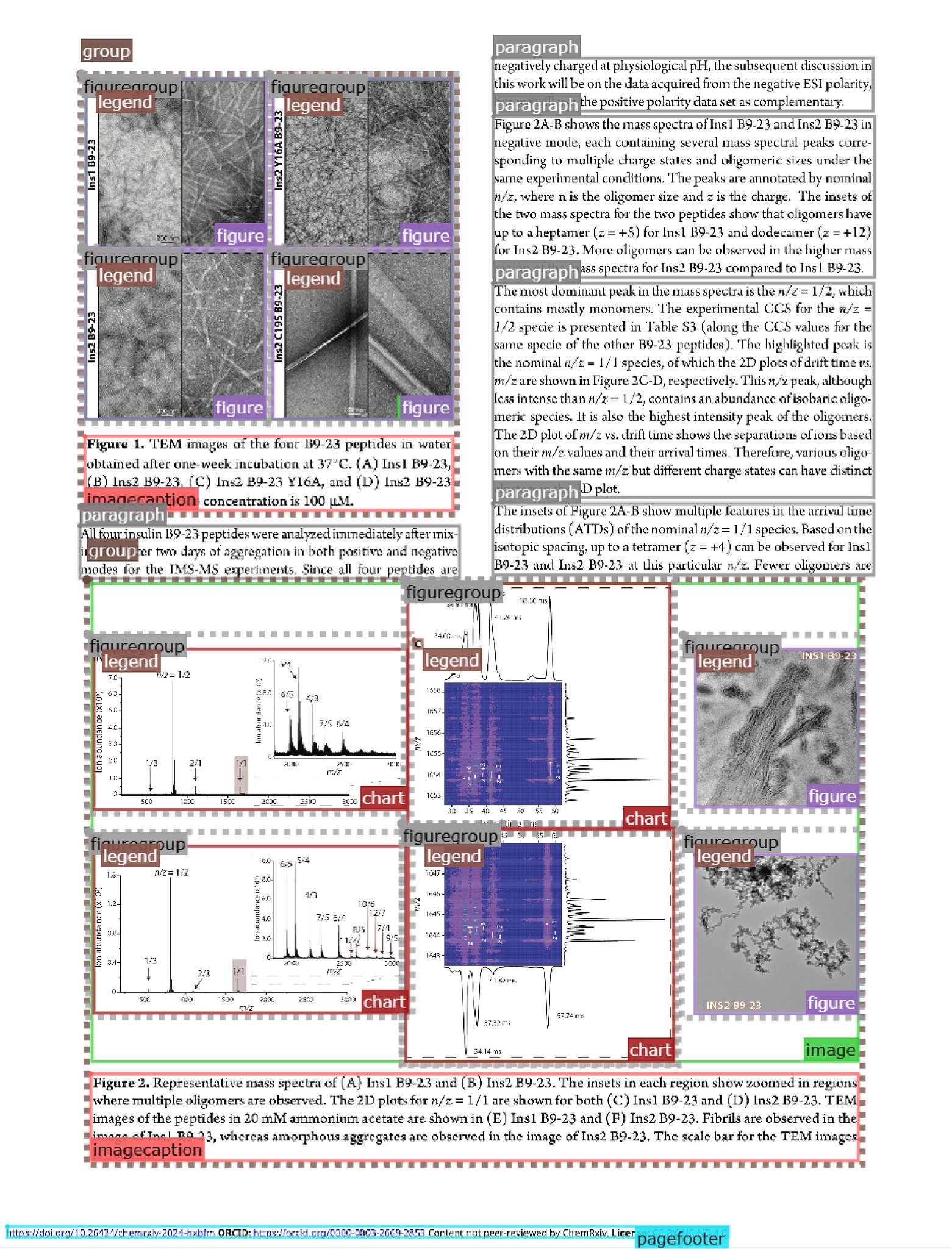}{Uni-Parser}{uniparser-color}

\vspace{12pt} 

\ShowcaseCellBoxed{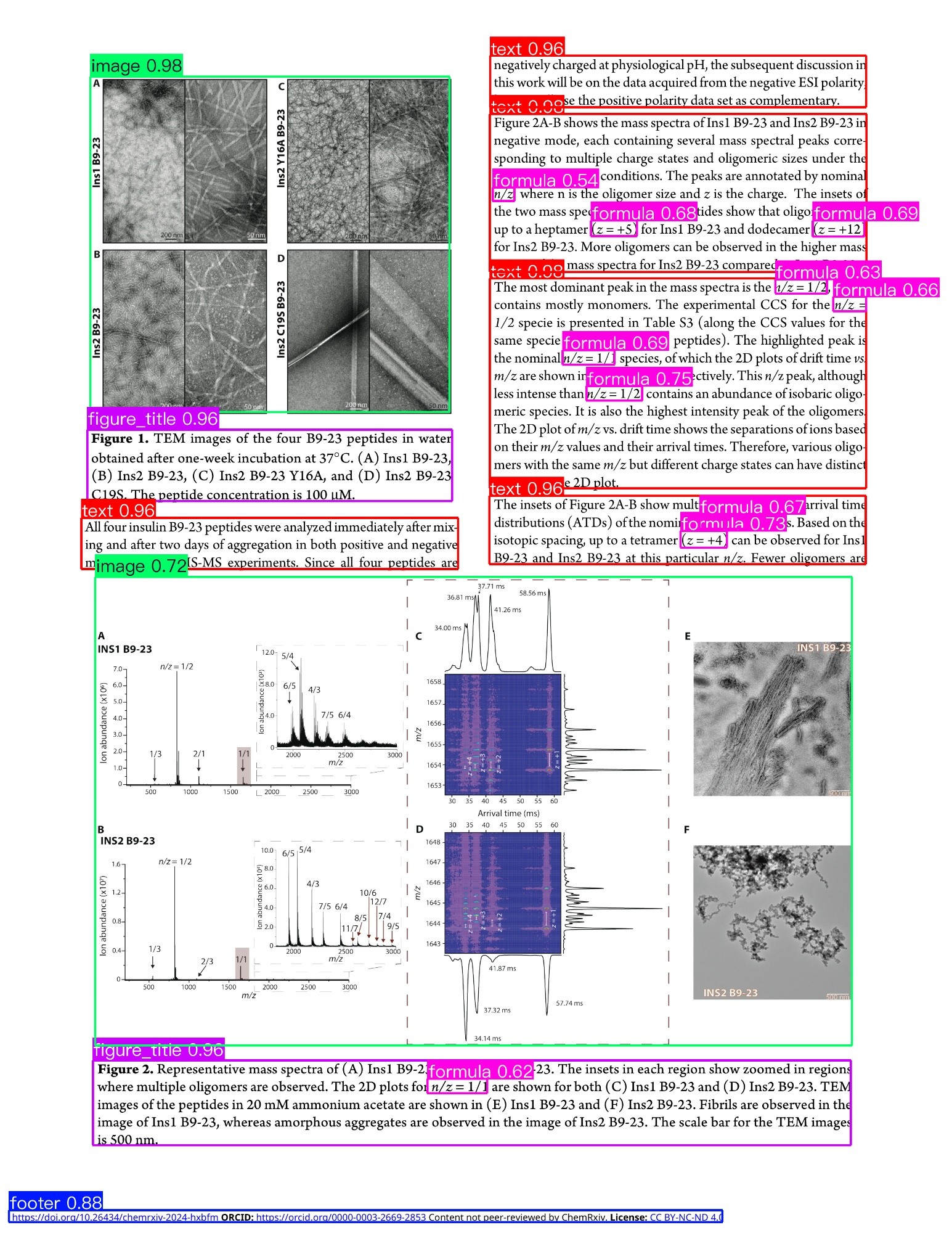}{PP-StructureV3}{gray!80}
\hfill
\ShowcaseCellBoxed{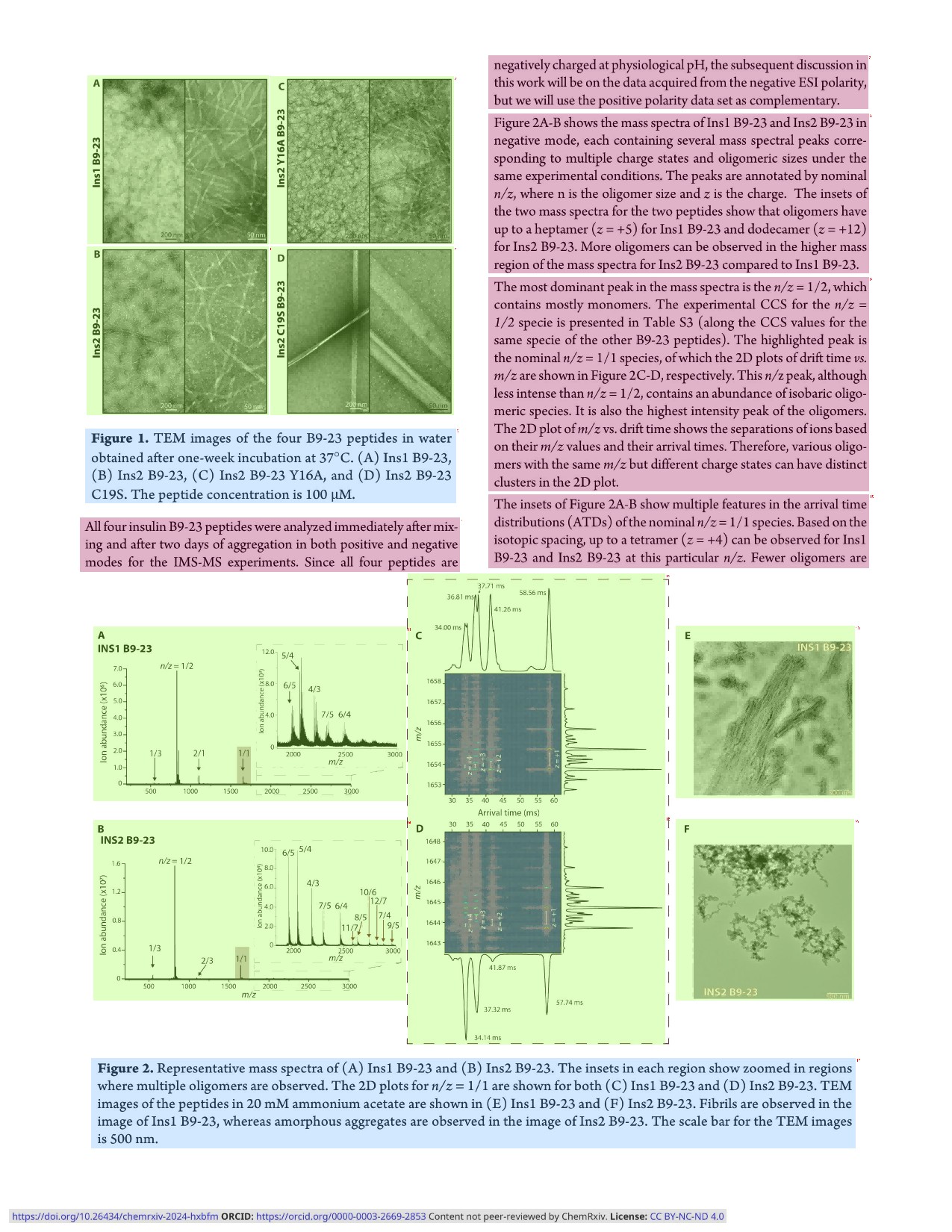}{MinerU2.5}{gray!80}

\end{minipage}
};



\end{tikzpicture}

\end{figure}


\begin{figure}[htpb]
\centering

\begin{tikzpicture}[remember picture]

\node (grid)[inner sep=0pt] {
\begin{minipage}{\textwidth}
\centering

\ShowcaseCellBoxed{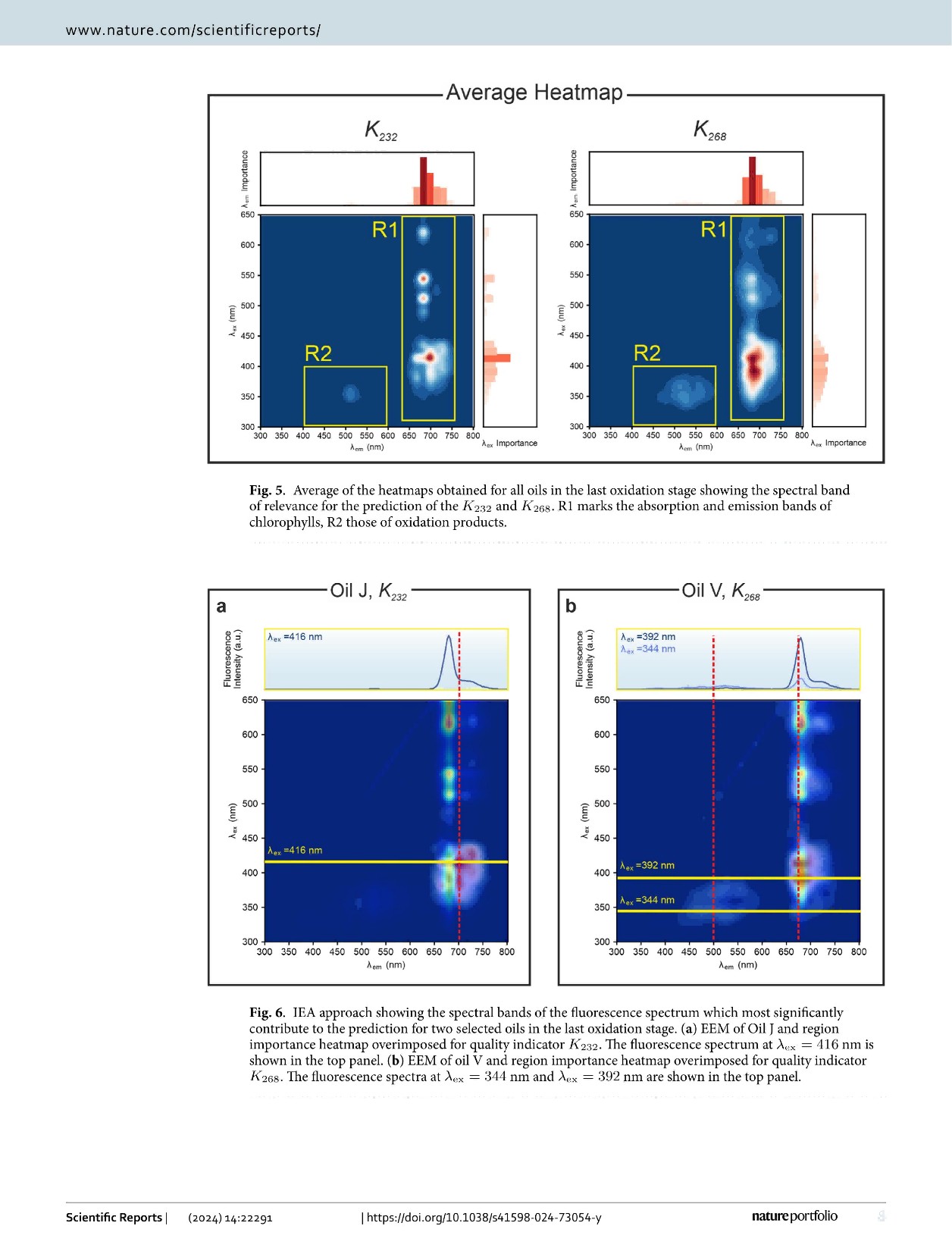}{Original Document}{black!80}
\hfill
\ShowcaseCellBoxed{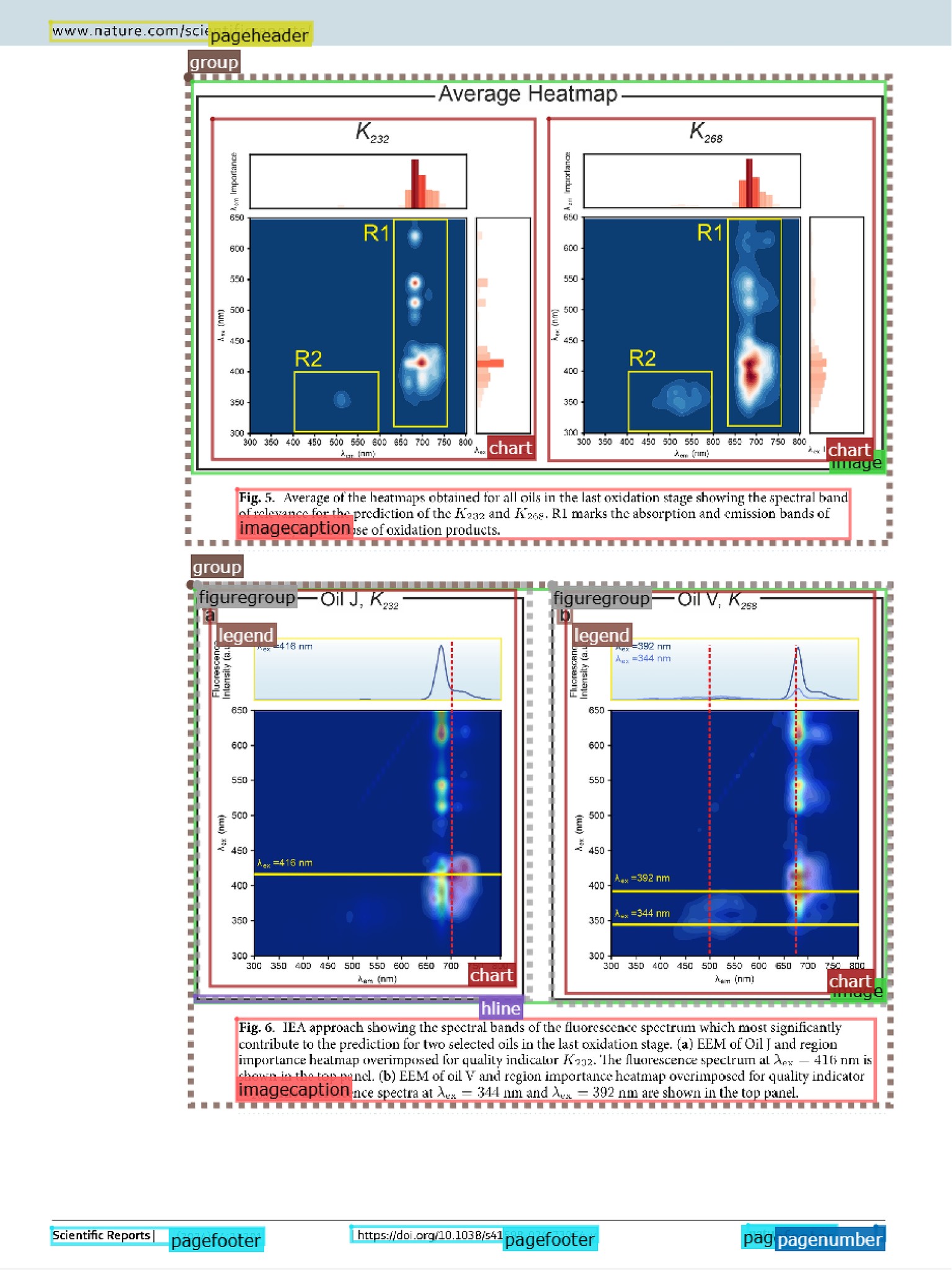}{Uni-Parser}{uniparser-color}

\vspace{12pt} 

\ShowcaseCellBoxed{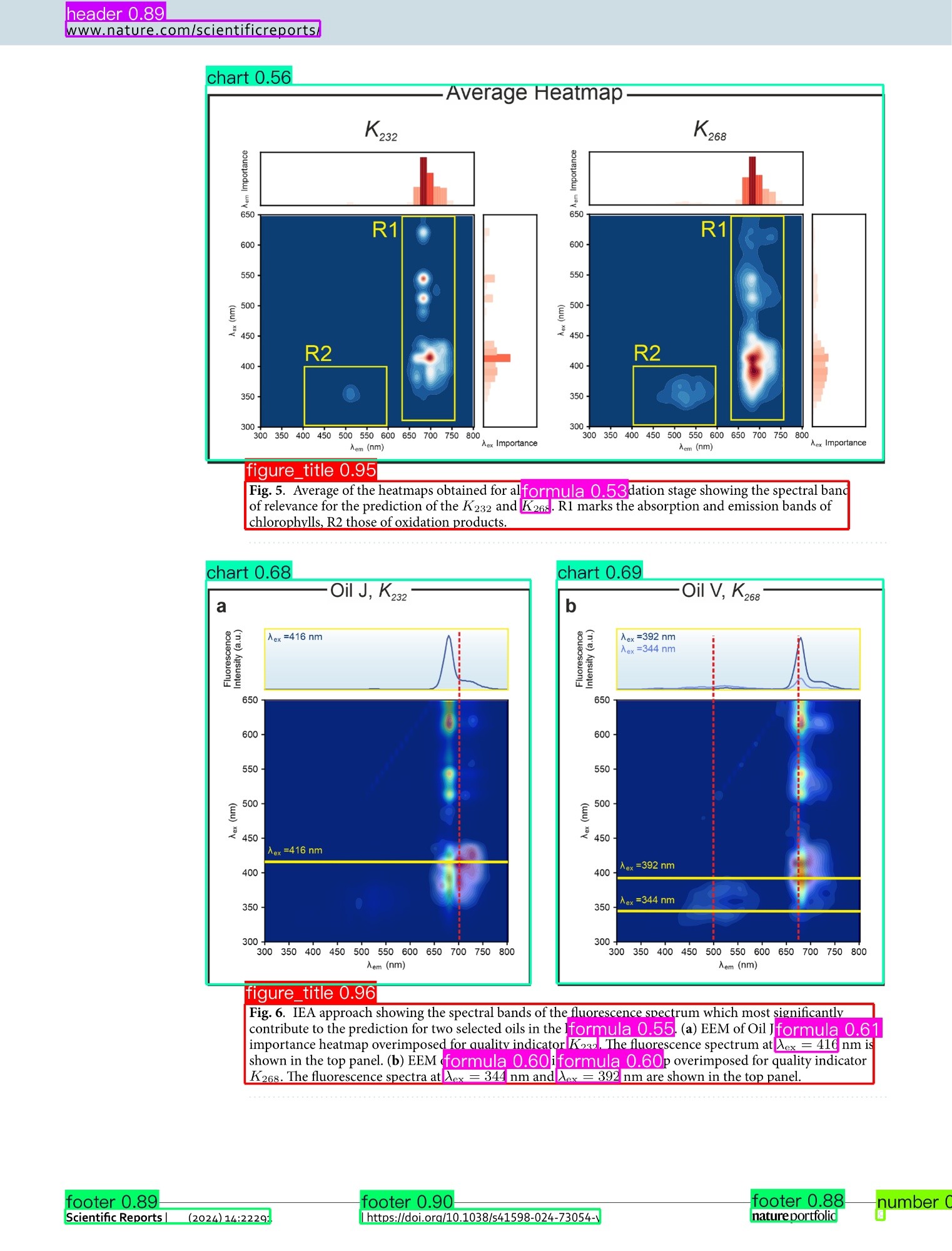}{PP-StructureV3}{gray!80}
\hfill
\ShowcaseCellBoxed{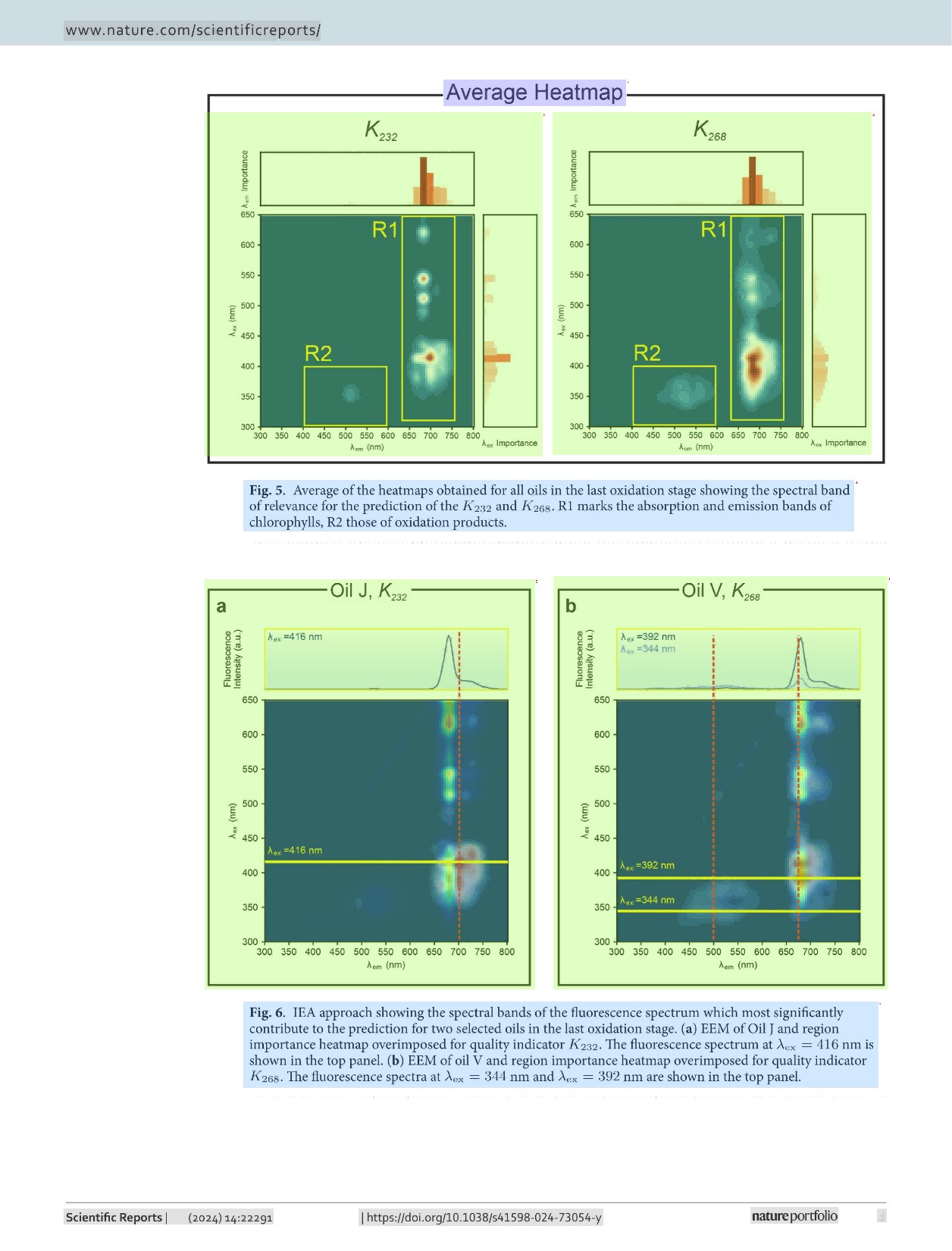}{MinerU2.5}{gray!80}

\end{minipage}
};



\end{tikzpicture}

\end{figure}


\begin{figure}[htpb]
\centering

\begin{tikzpicture}[remember picture]

\node (grid)[inner sep=0pt] {
\begin{minipage}{\textwidth}
\centering

\ShowcaseCellBoxed{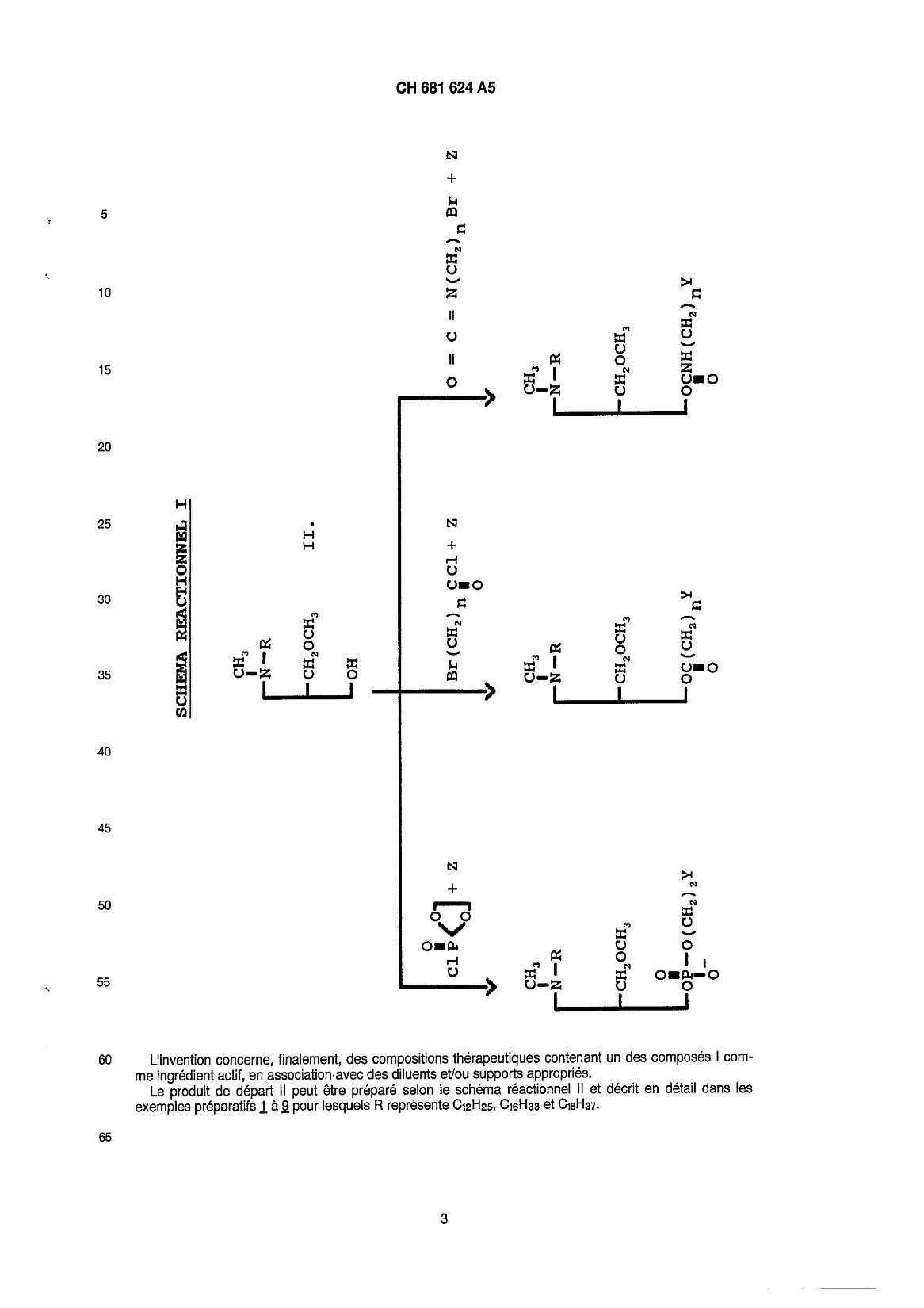}{Original Document}{black!80}
\hfill
\ShowcaseCellBoxed{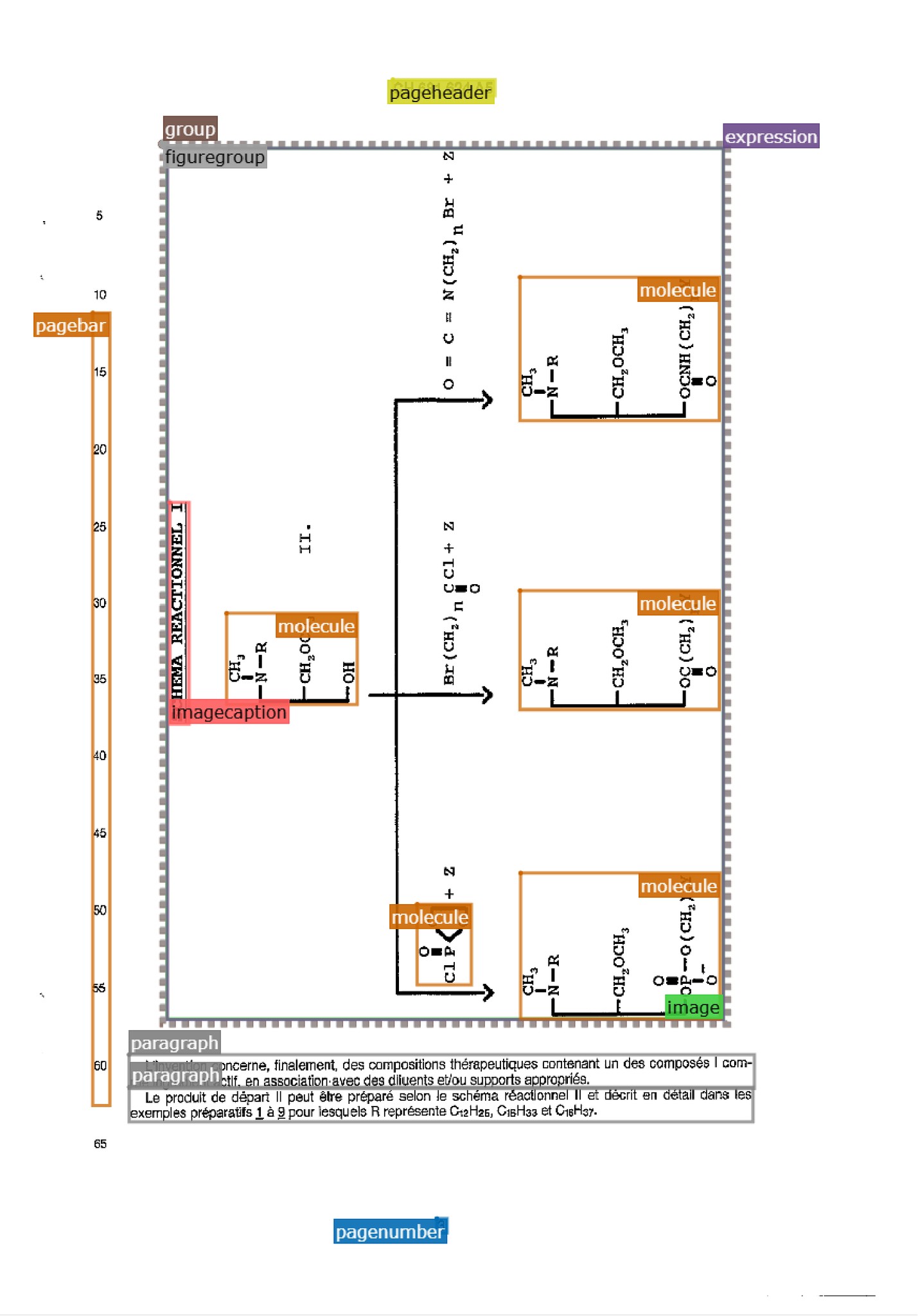}{Uni-Parser}{uniparser-color}

\vspace{12pt} 

\ShowcaseCellBoxed{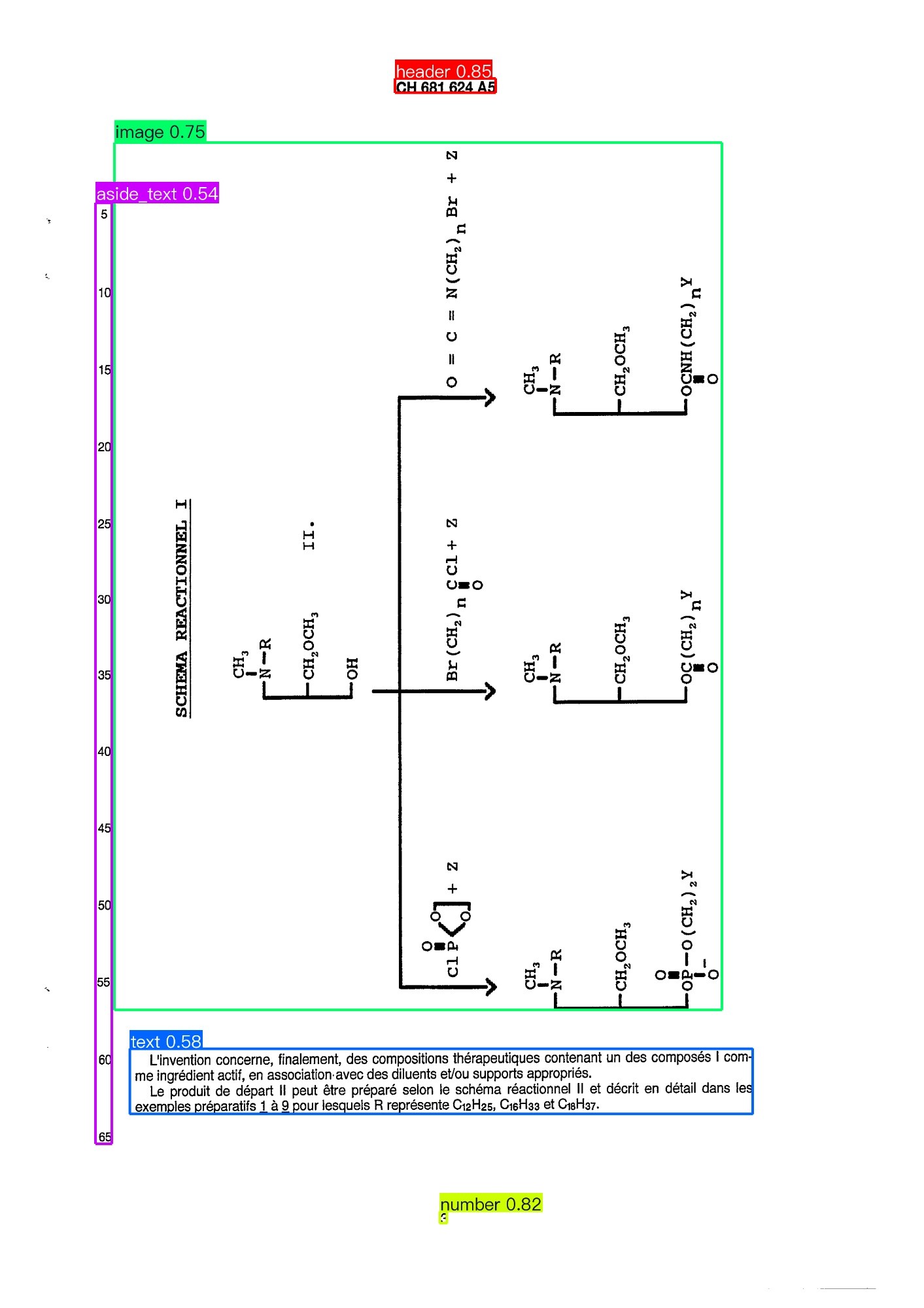}{PP-StructureV3}{gray!80}
\hfill
\ShowcaseCellBoxed{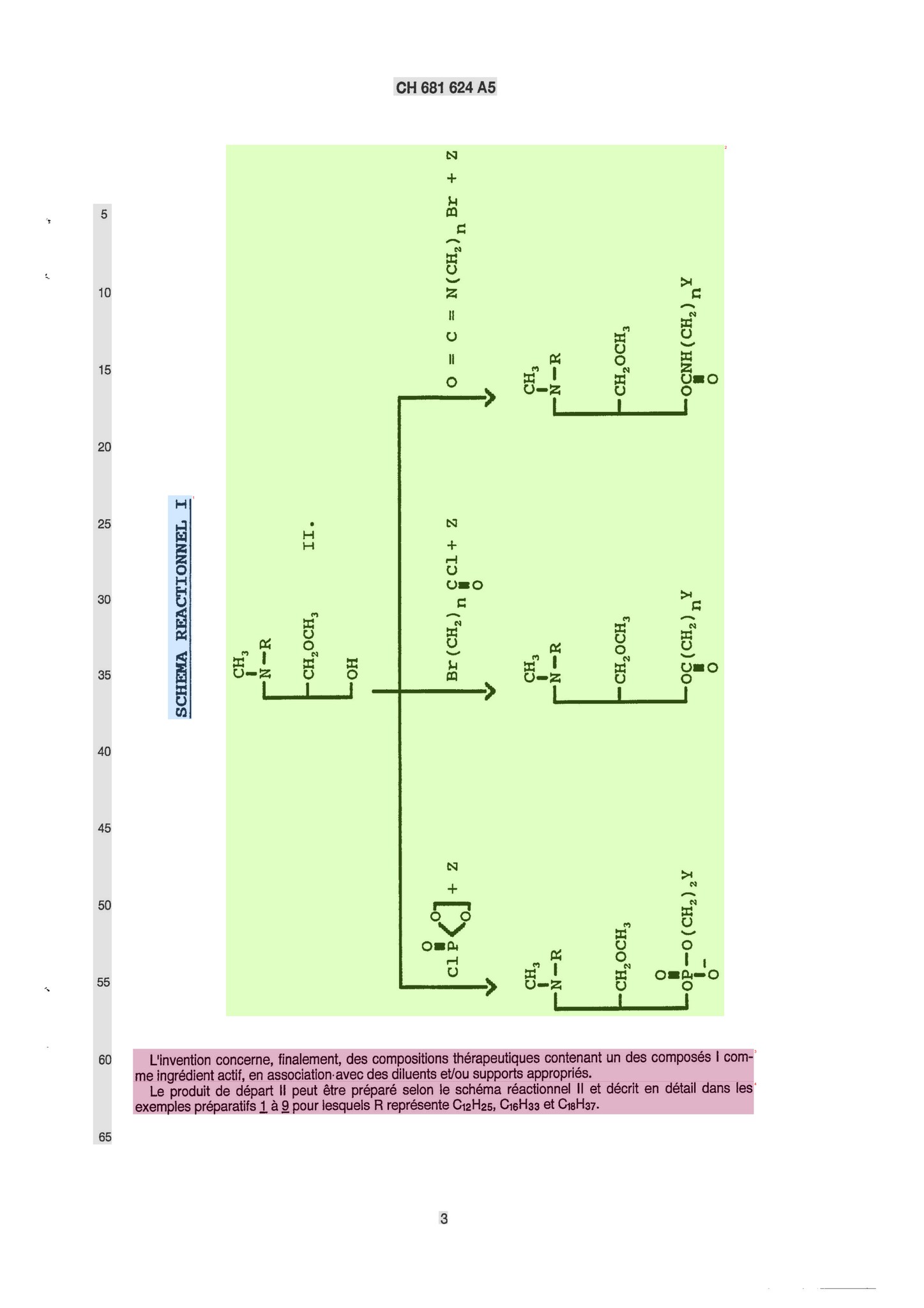}{MinerU2.5}{gray!80}

\end{minipage}
};



\end{tikzpicture}

\end{figure}


\begin{figure}[htpb]
\centering

\begin{tikzpicture}[remember picture]

\node (grid)[inner sep=0pt] {
\begin{minipage}{\textwidth}
\centering

\ShowcaseCellBoxed{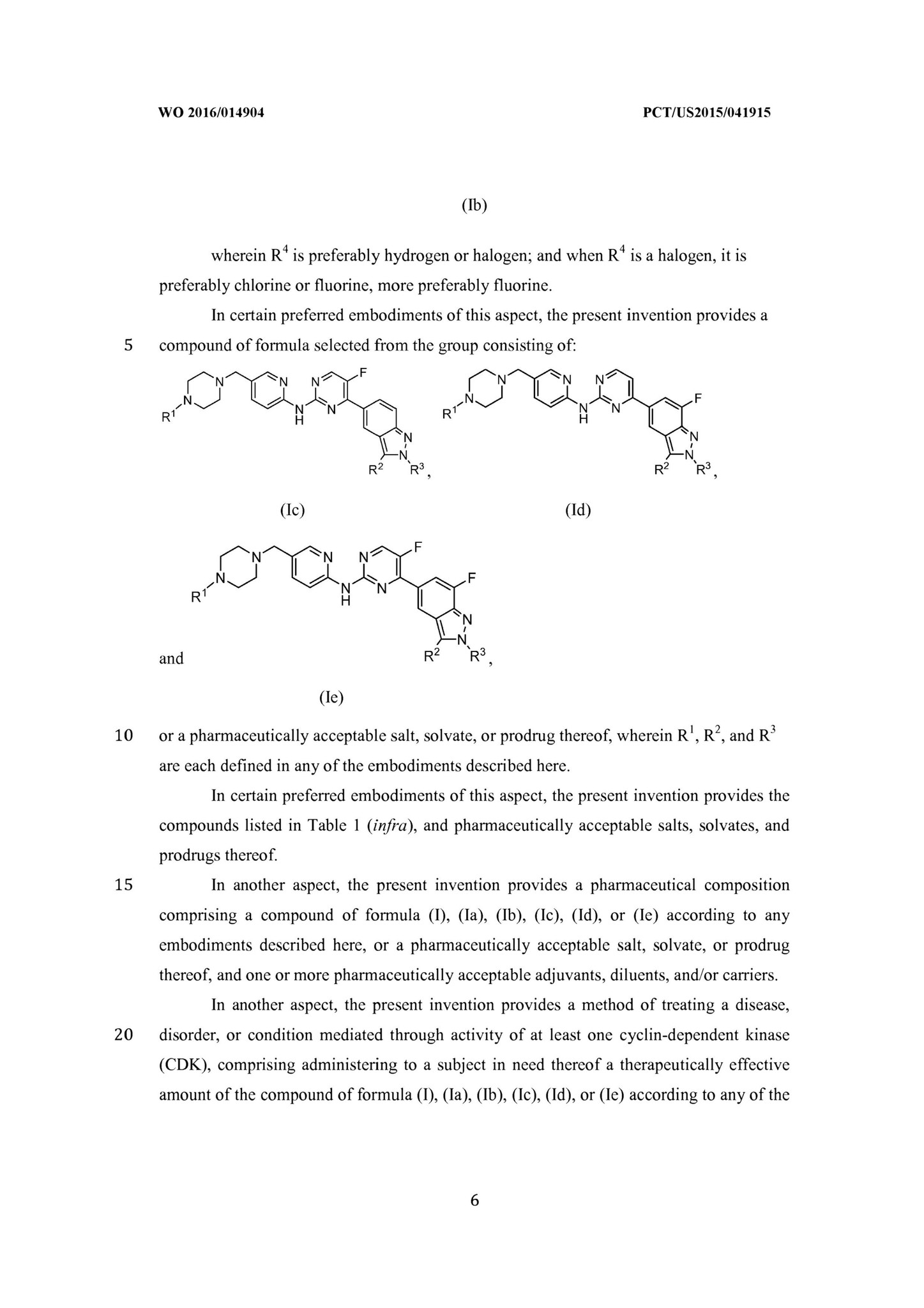}{Original Document}{black!80}
\hfill
\ShowcaseCellBoxed{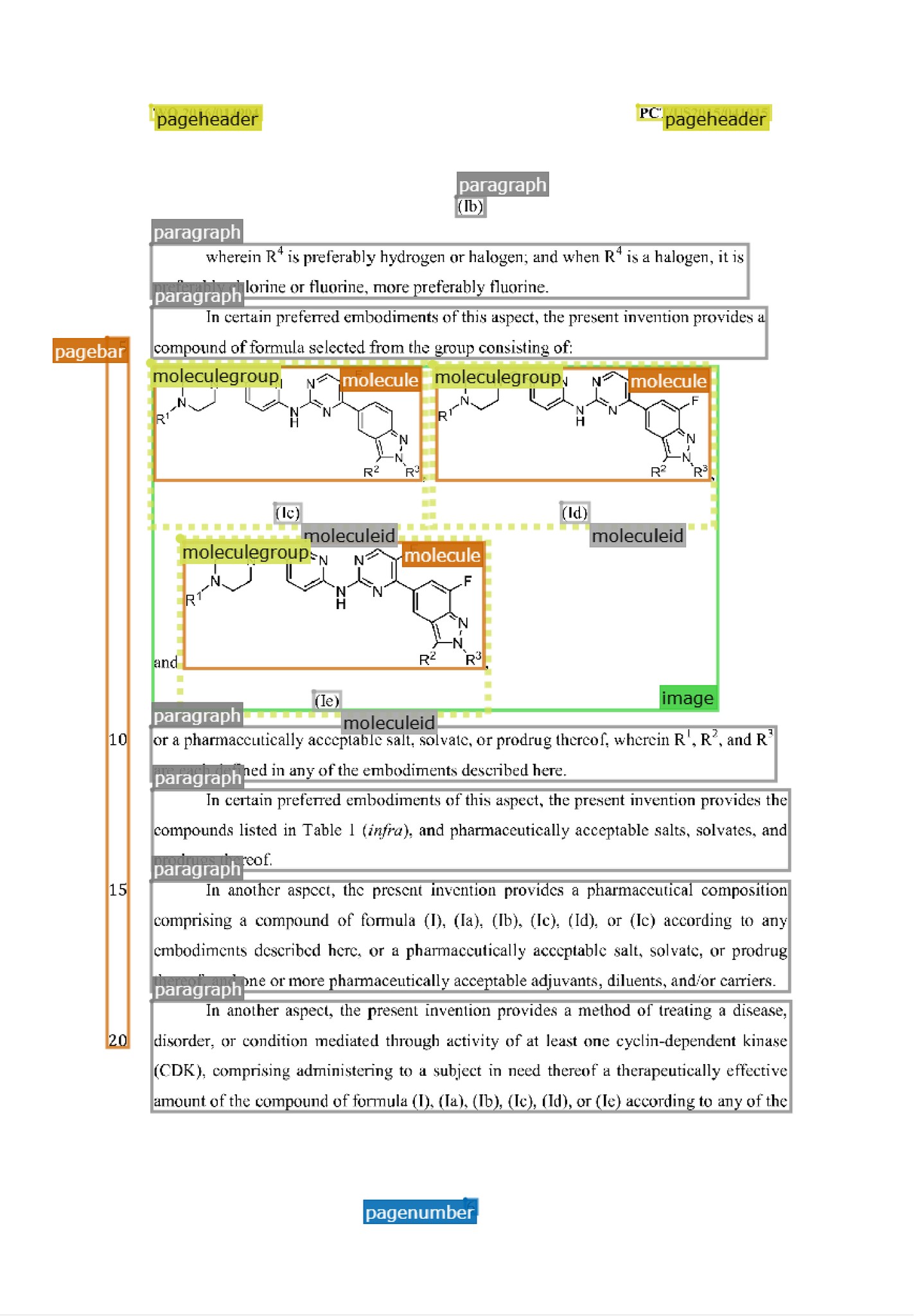}{Uni-Parser}{uniparser-color}

\vspace{12pt} 

\ShowcaseCellBoxed{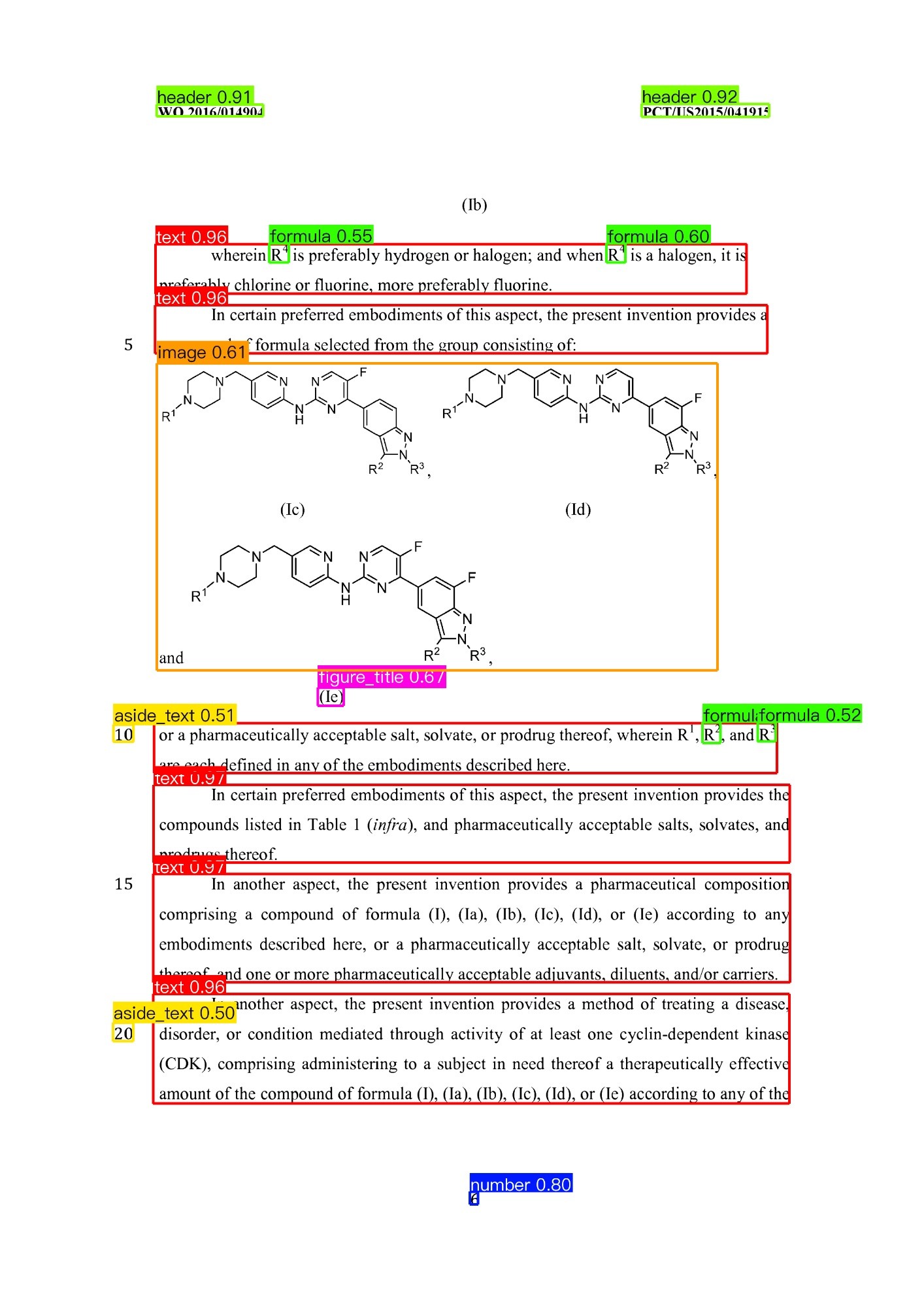}{PP-StructureV3}{gray!80}
\hfill
\ShowcaseCellBoxed{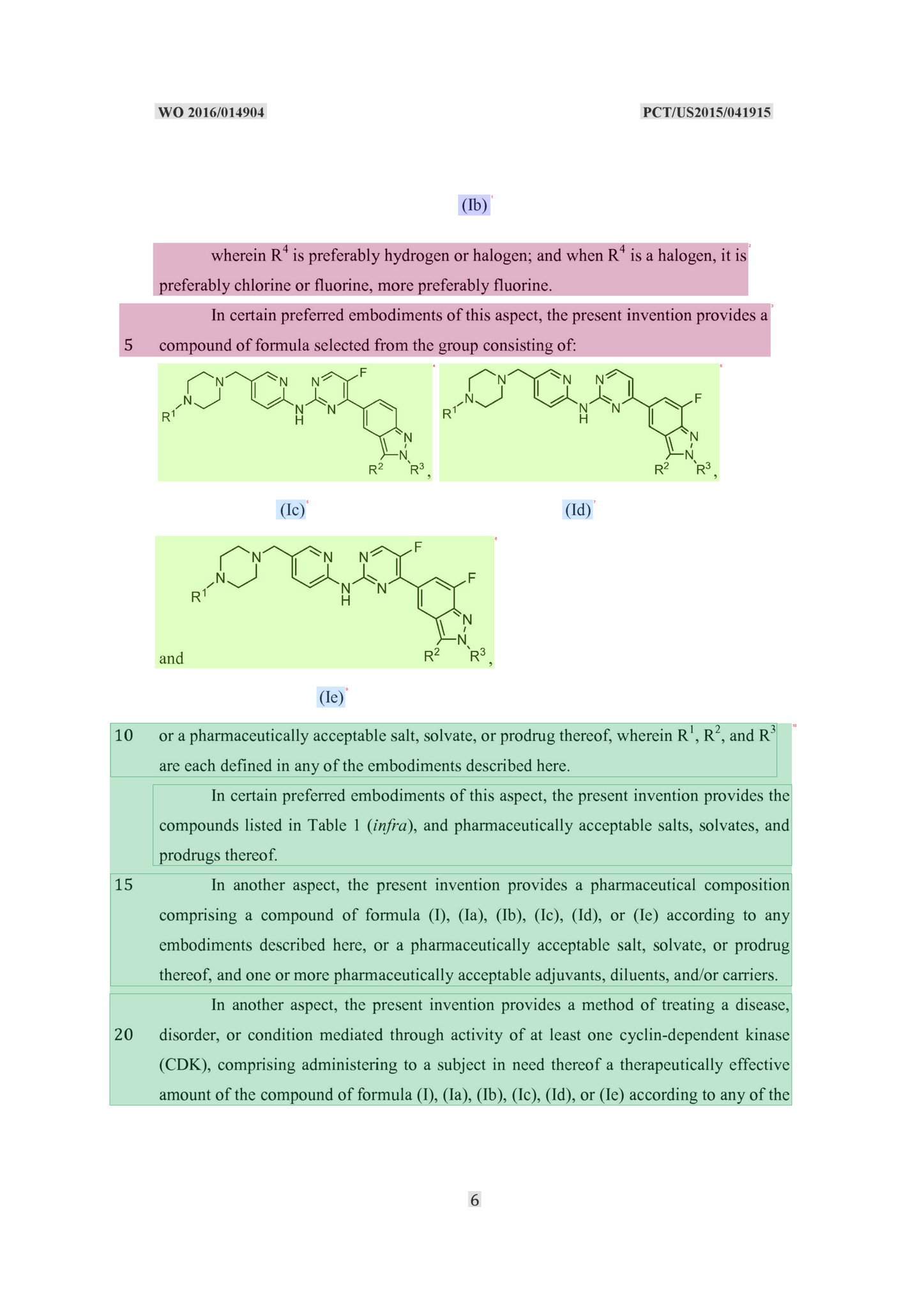}{MinerU2.5}{gray!80}

\end{minipage}
};



\end{tikzpicture}

\end{figure}


\newpage
\subsection{Qualitative Examples of Reading Order}
\label{appendix:rocase}

\begin{figure}[htpb]
\centering

\begin{tikzpicture}[remember picture]

\node (grid)[inner sep=0pt] {
\begin{minipage}{\textwidth}
\centering

\ShowcaseCellBoxed{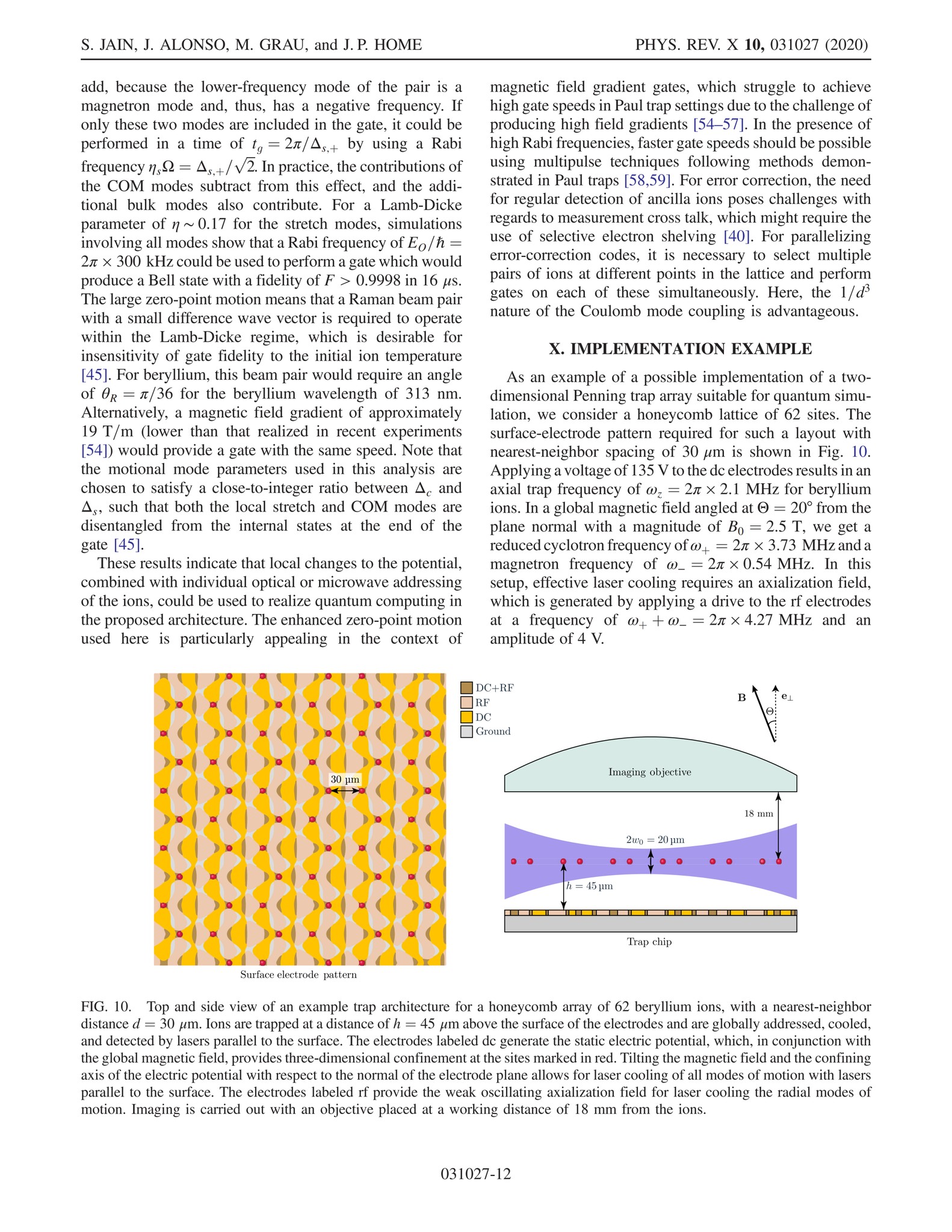}{Original Document}{black!80}
\hfill
\ShowcaseCellBoxed{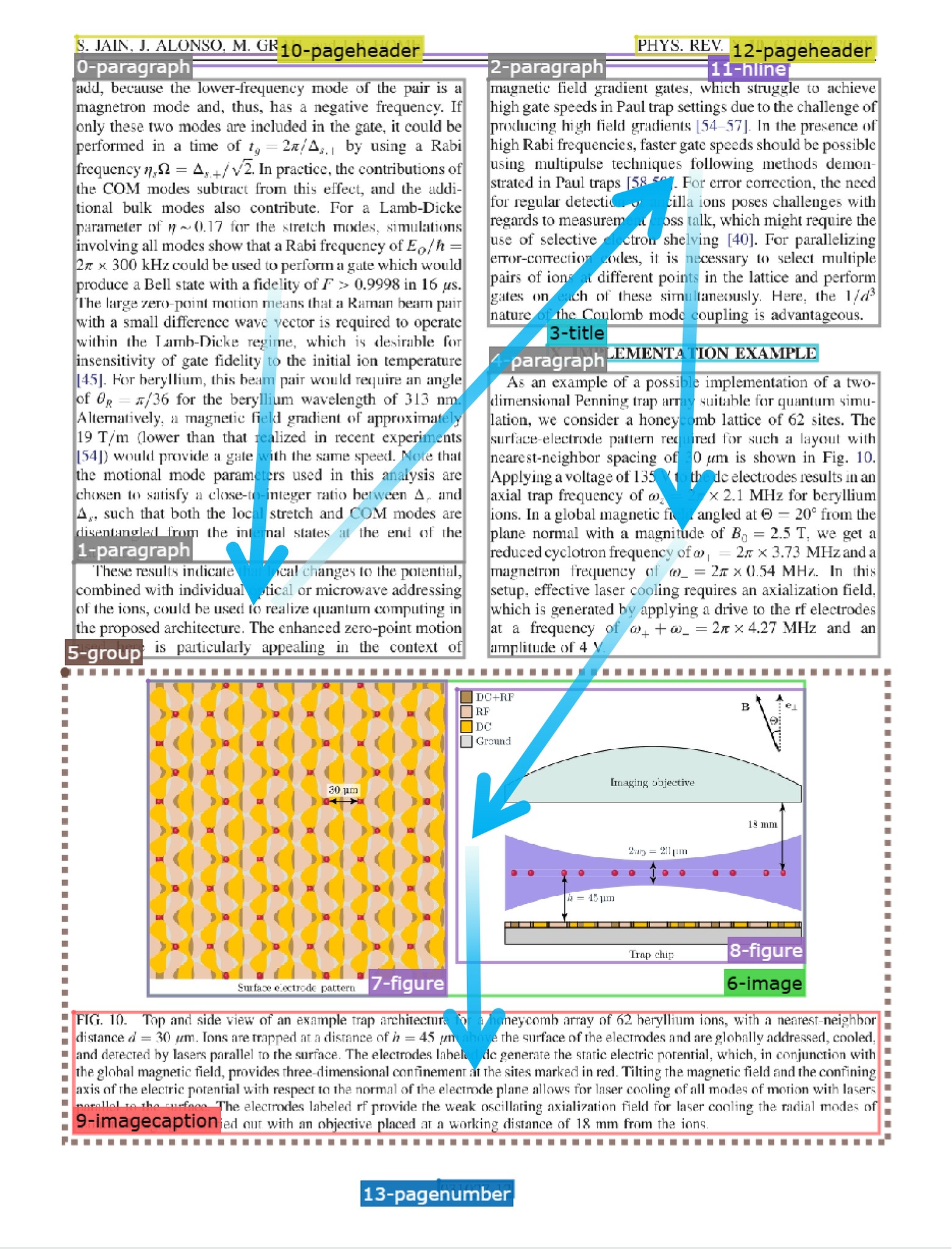}{Uni-Parser}{uniparser-color}

\vspace{12pt} 

\ShowcaseCellBoxed{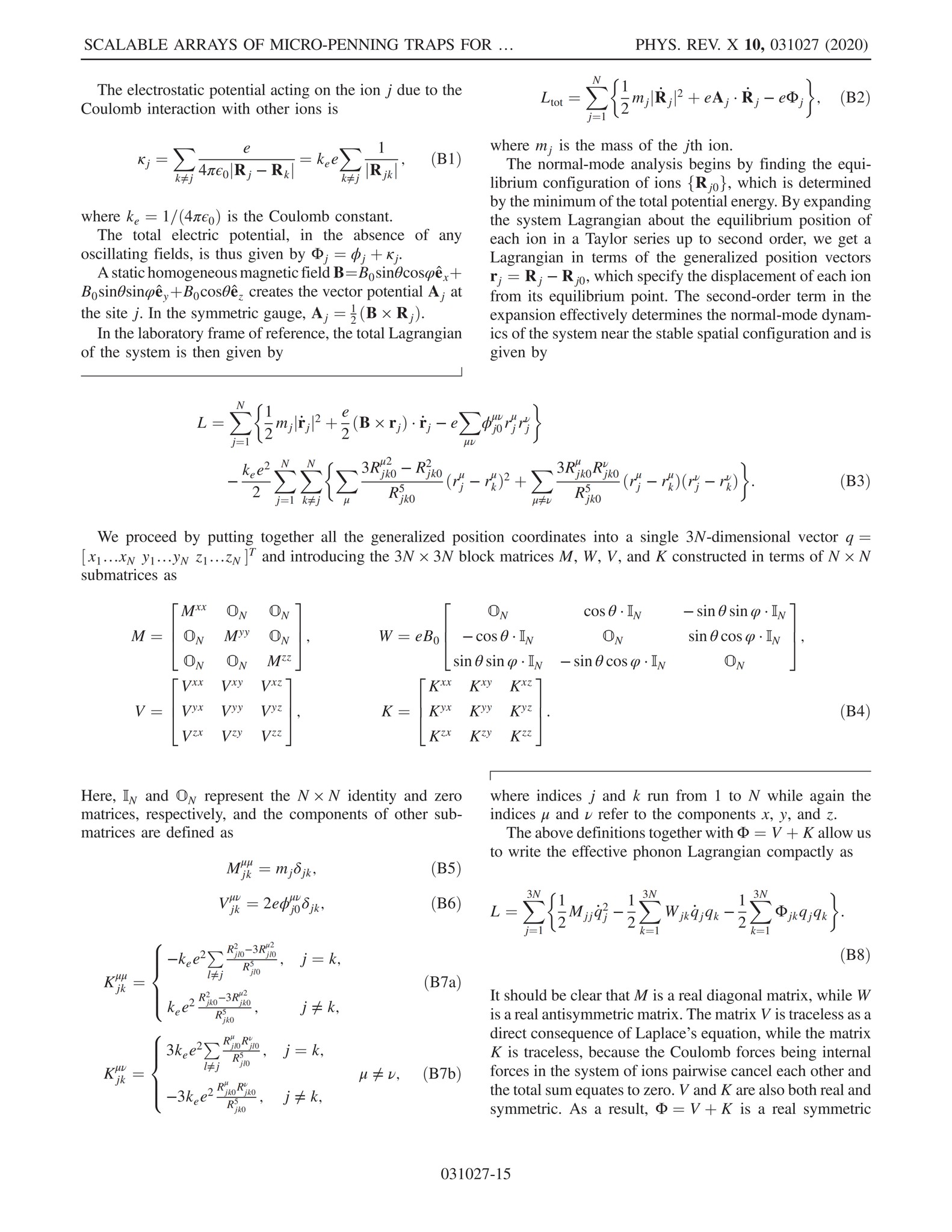}{Original Document}{black!80}
\hfill
\ShowcaseCellBoxed{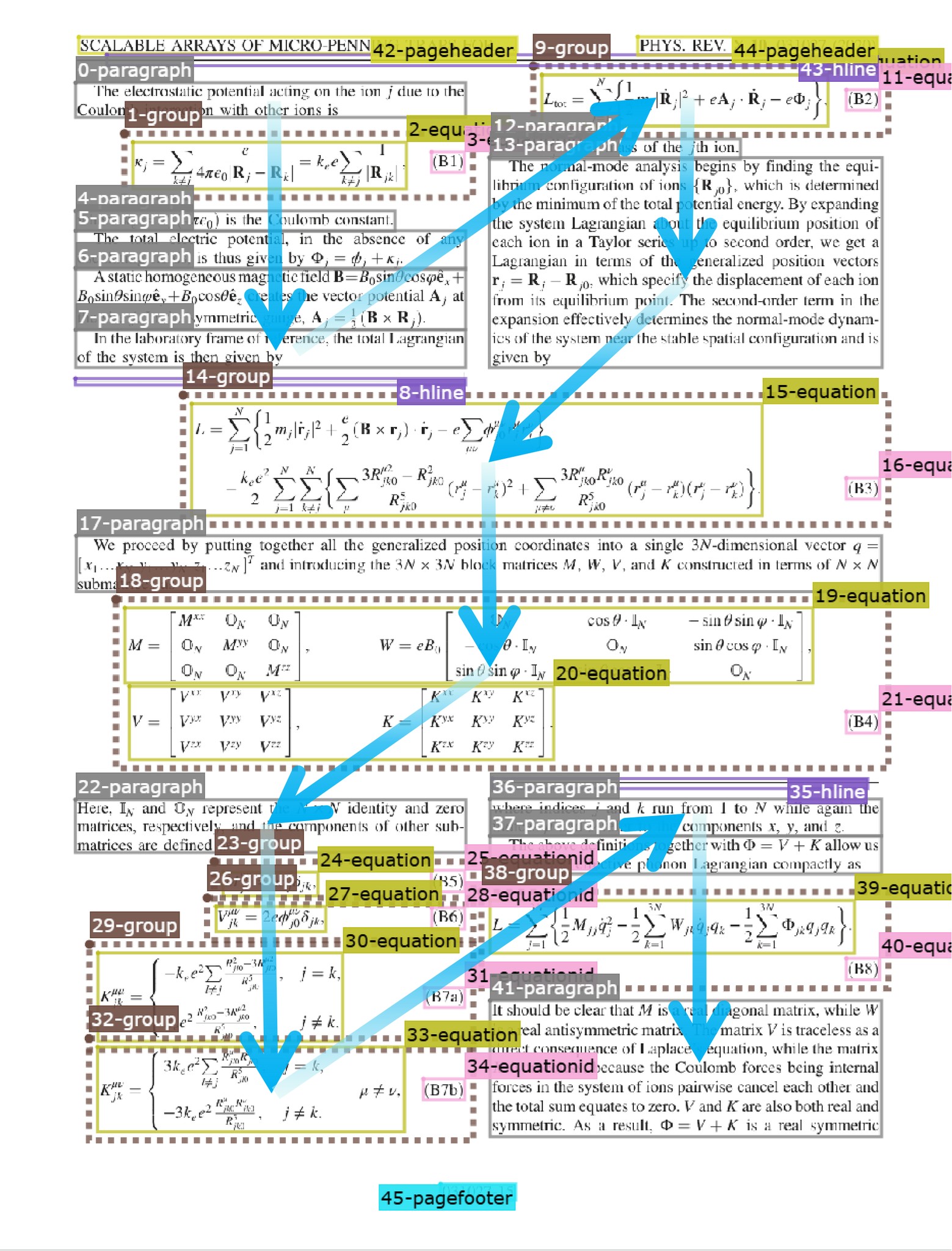}{Uni-Parser}{uniparser-color}

\end{minipage}
};

\draw[gridline] (grid.west) -- (grid.east);

\end{tikzpicture}

\end{figure}


\begin{figure}[htpb]
\centering

\begin{tikzpicture}[remember picture]

\node (grid)[inner sep=0pt] {
\begin{minipage}{\textwidth}
\centering

\ShowcaseCellBoxed{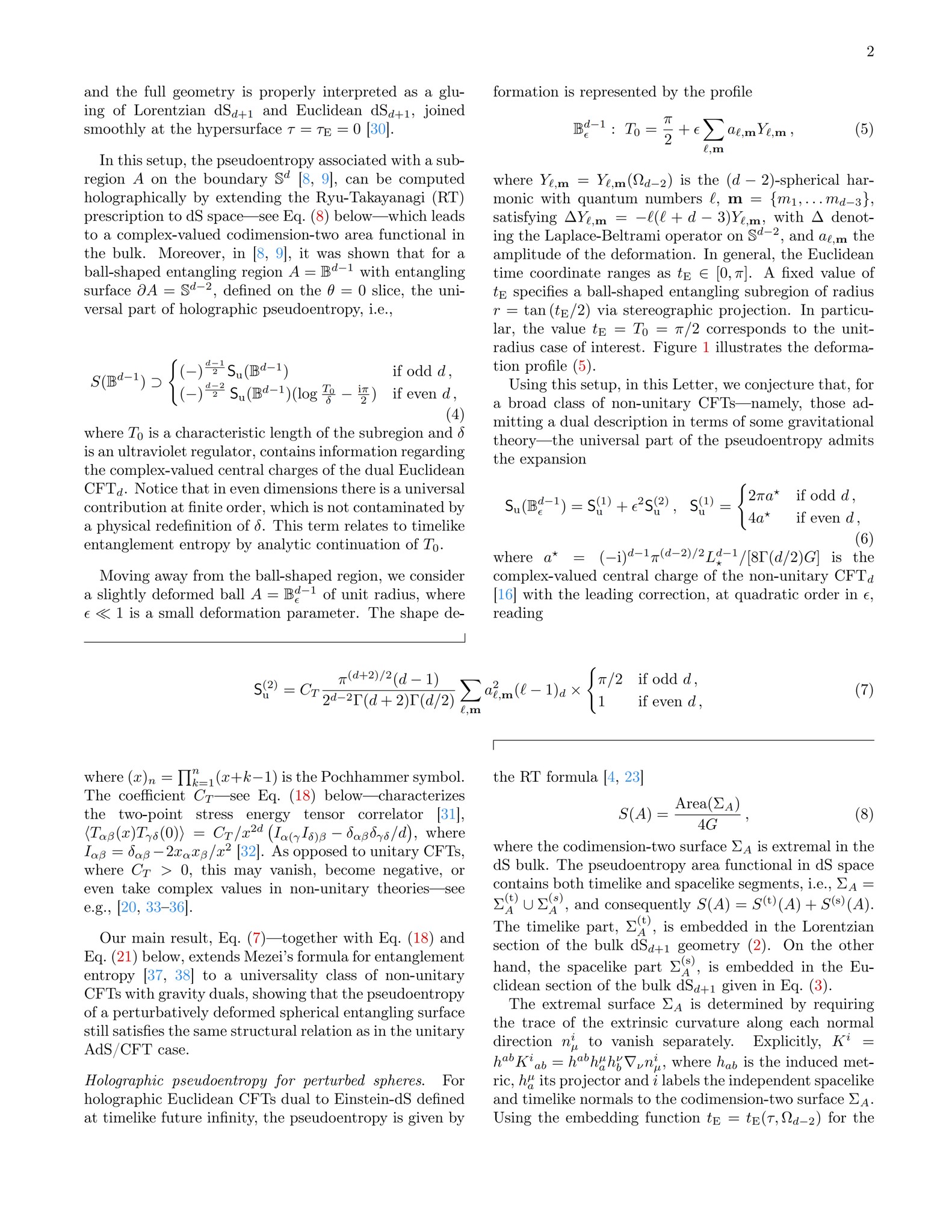}{Original Document}{black!80}
\hfill
\ShowcaseCellBoxed{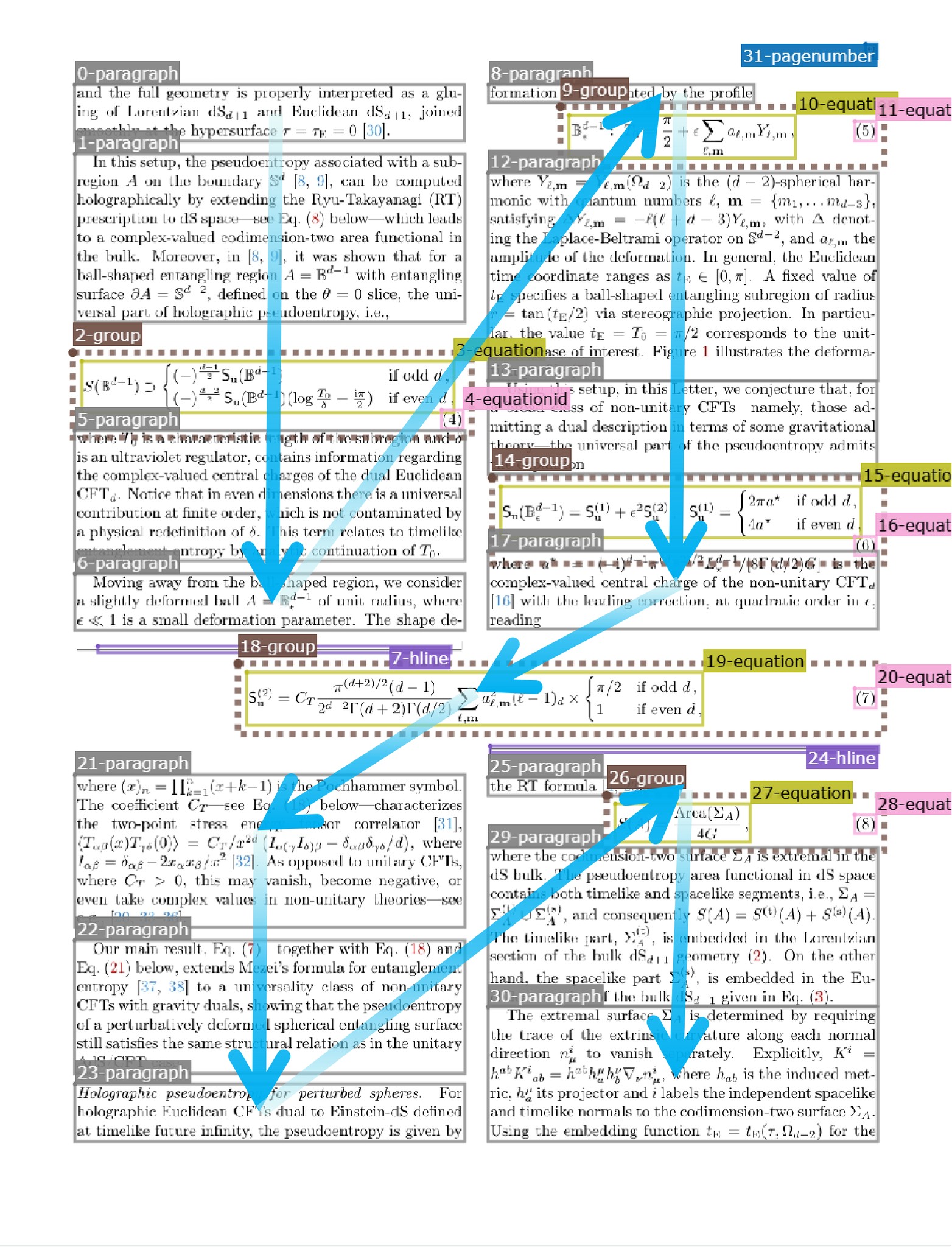}{Uni-Parser}{uniparser-color}

\vspace{12pt} 

\ShowcaseCellBoxed{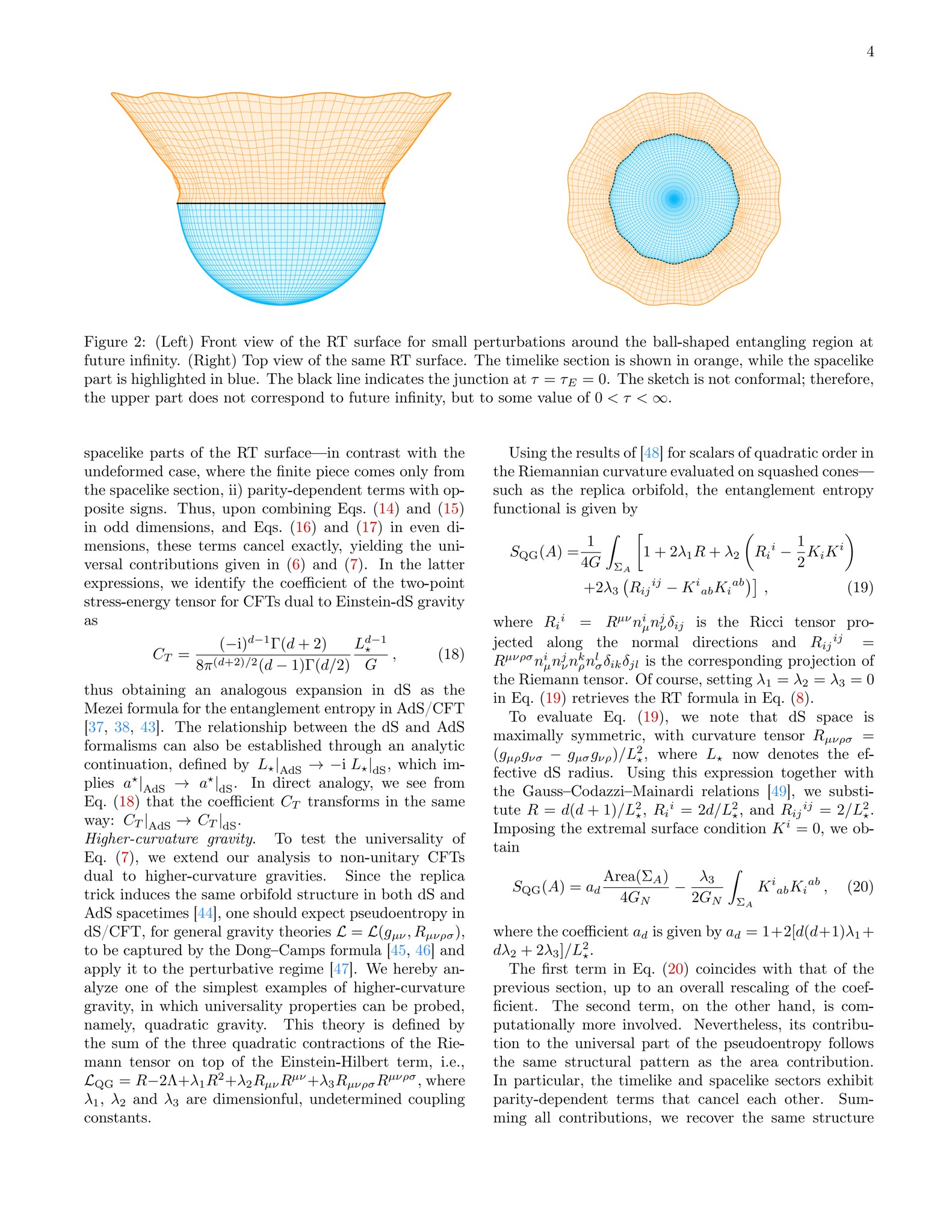}{Original Document}{black!80}
\hfill
\ShowcaseCellBoxed{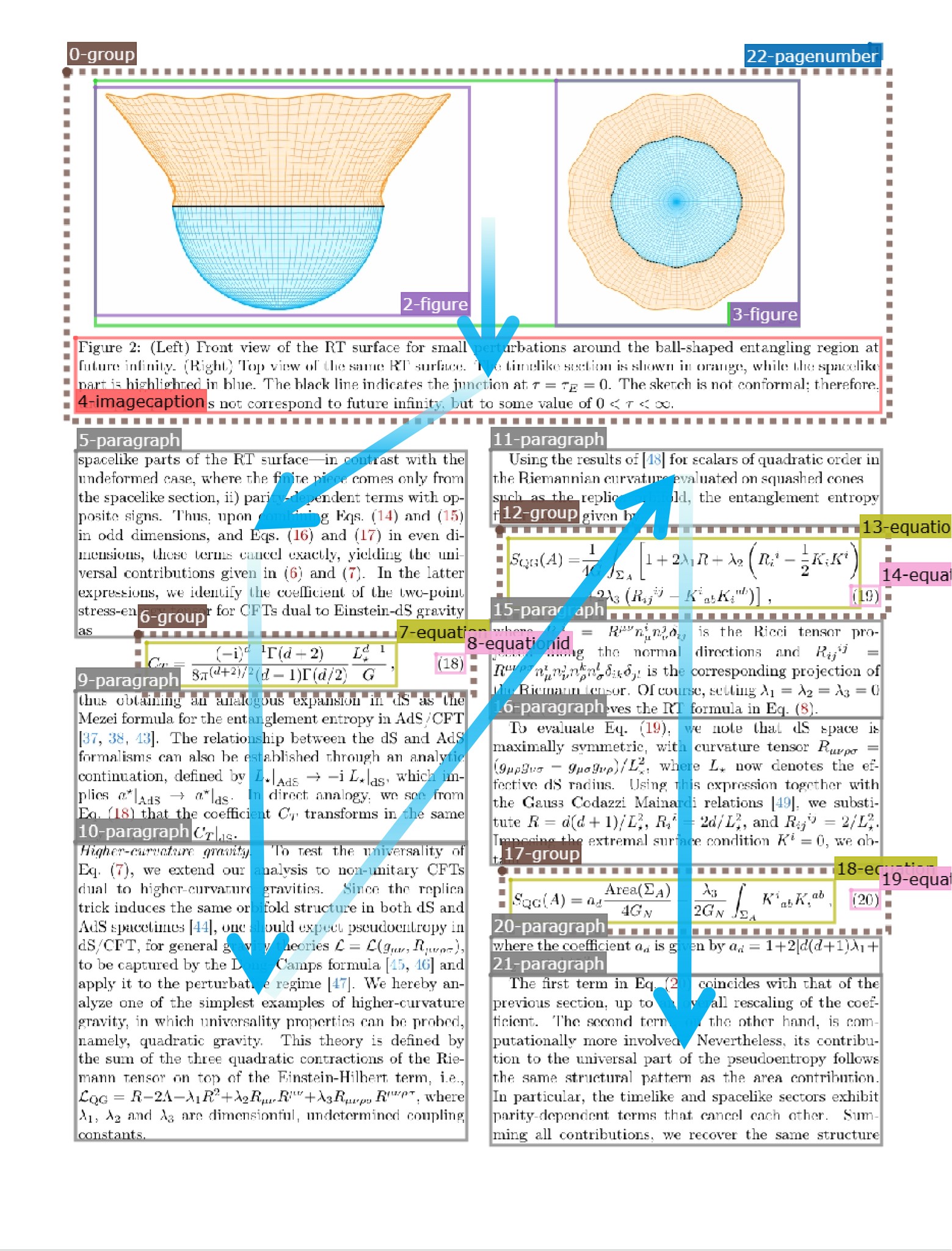}{Uni-Parser}{uniparser-color}

\end{minipage}
};

\draw[gridline] (grid.west) -- (grid.east);

\end{tikzpicture}

\end{figure}


\newpage
\subsection{Uni-Parser Case Study}
\label{appendix:case}

\begin{figure}[htpb]
\centering

\begin{tikzpicture}[remember picture]

\node (grid)[inner sep=0pt] {
\begin{minipage}{\textwidth}
\centering

\ShowcaseCellBoxed{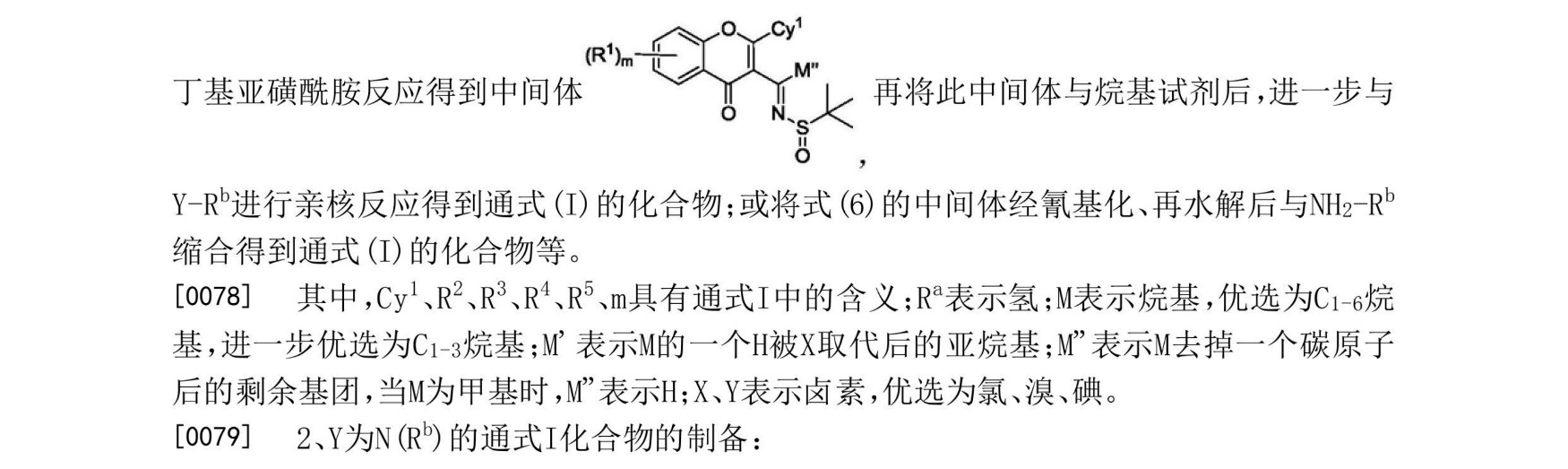}{Original Document}{black!80}
\hfill
\ShowcaseCellBoxed{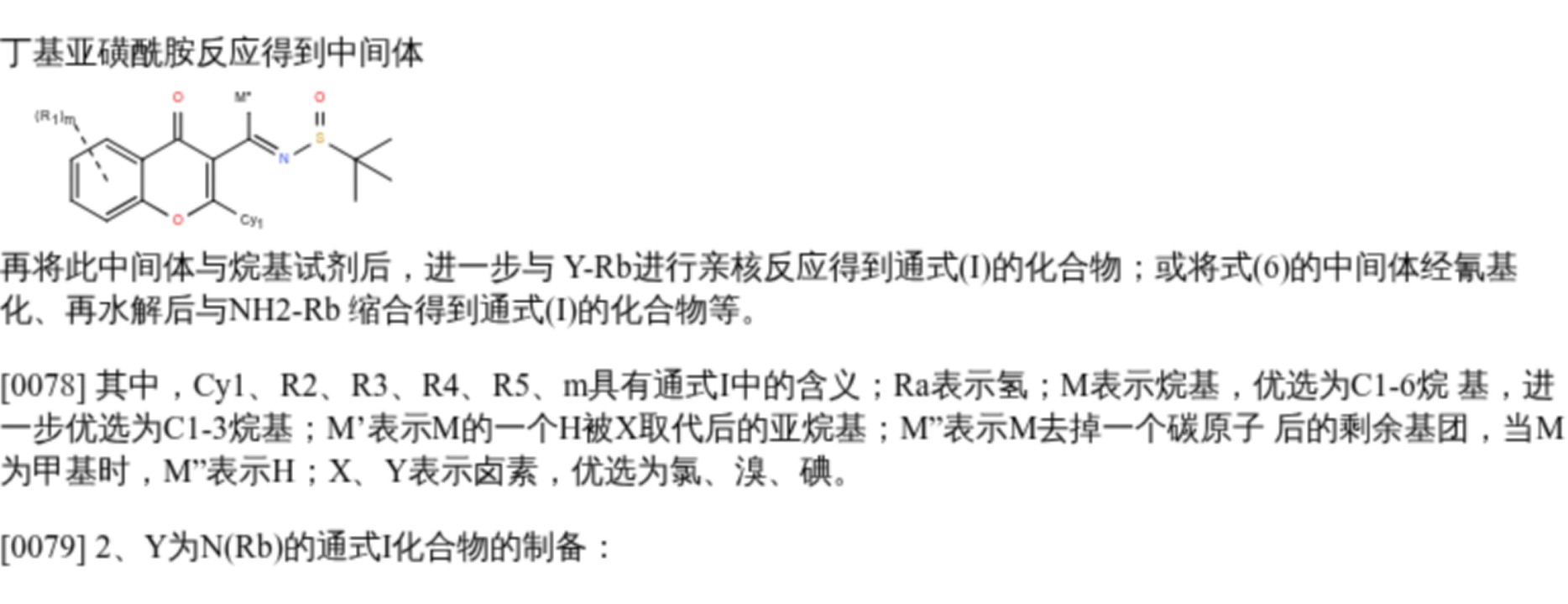}{Uni-Parser}{uniparser-color}

\vspace{12pt} 

\ShowcaseCellBoxed{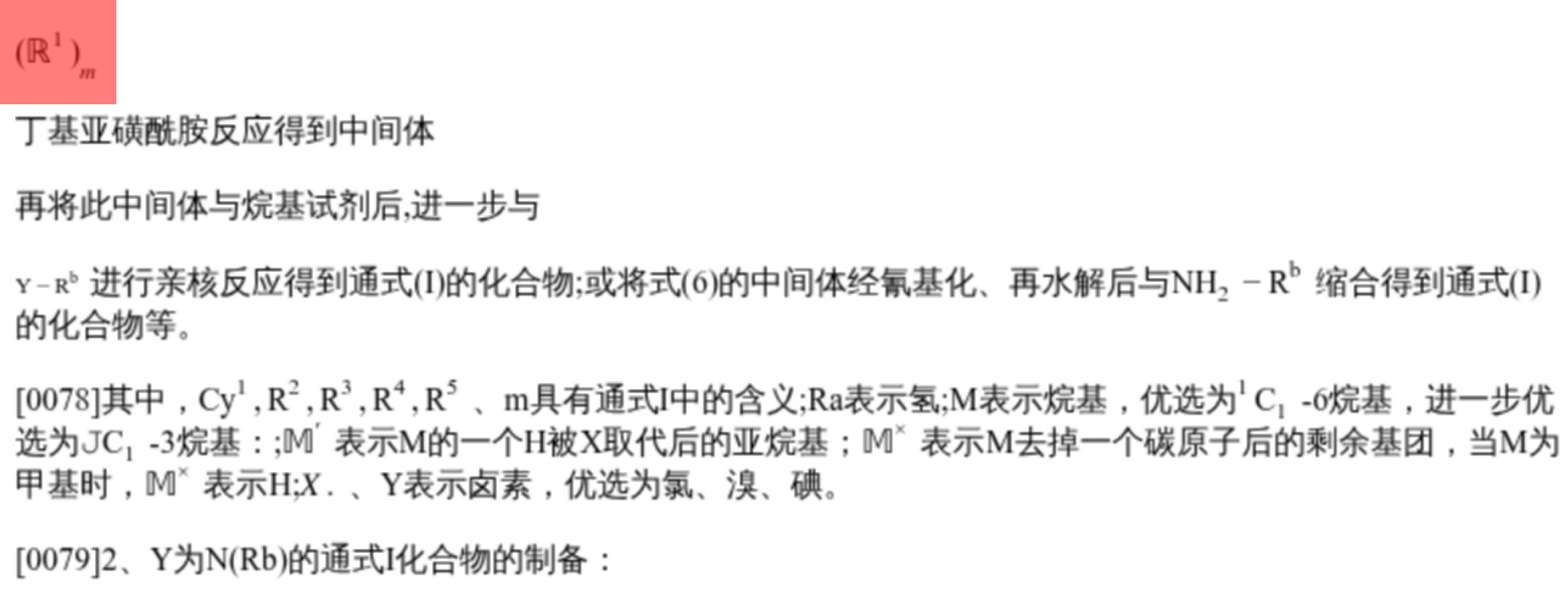}{PP-StructureV3}{gray!80}
\hfill
\ShowcaseCellBoxed{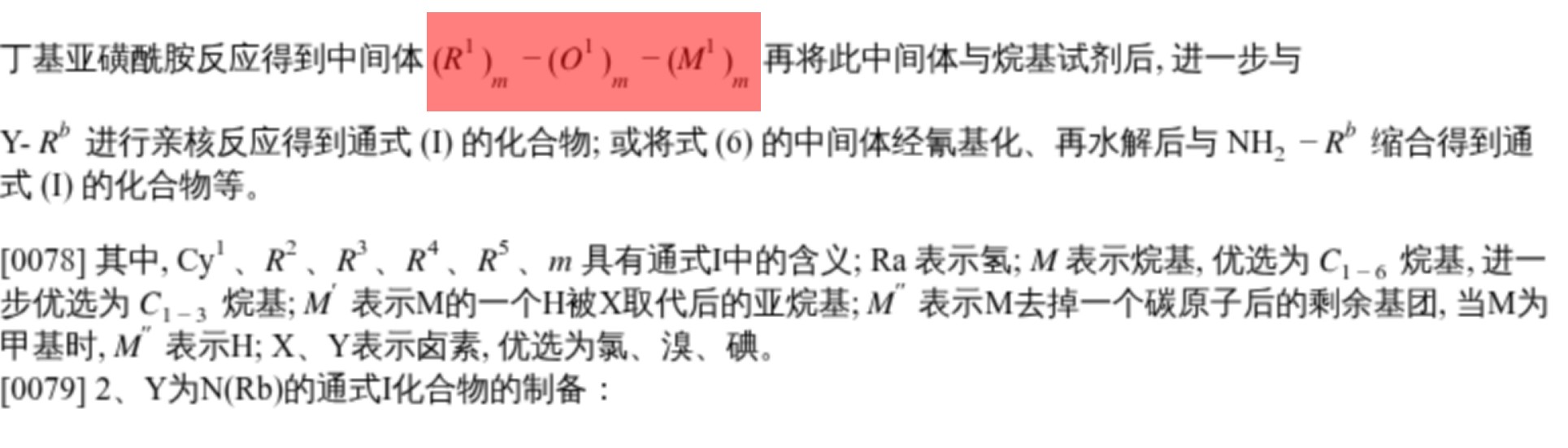}{MinerU2.5}{gray!80}

\end{minipage}
};



\end{tikzpicture}

\end{figure}


\begin{figure}[htpb]
\centering

\begin{tikzpicture}[remember picture]

\node (grid)[inner sep=0pt] {
\begin{minipage}{\textwidth}
\centering

\ShowcaseCellBoxed{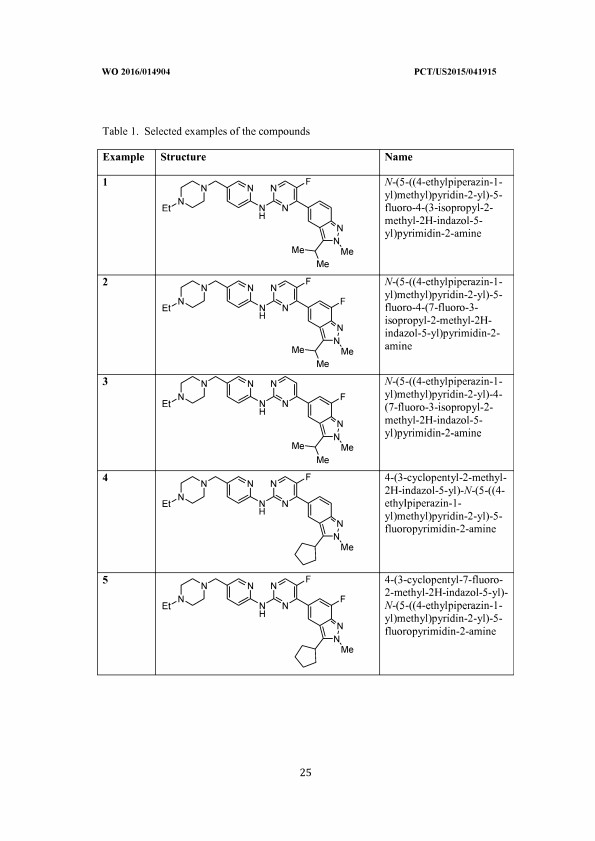}{Original Document}{black!80}
\hfill
\ShowcaseCellBoxed{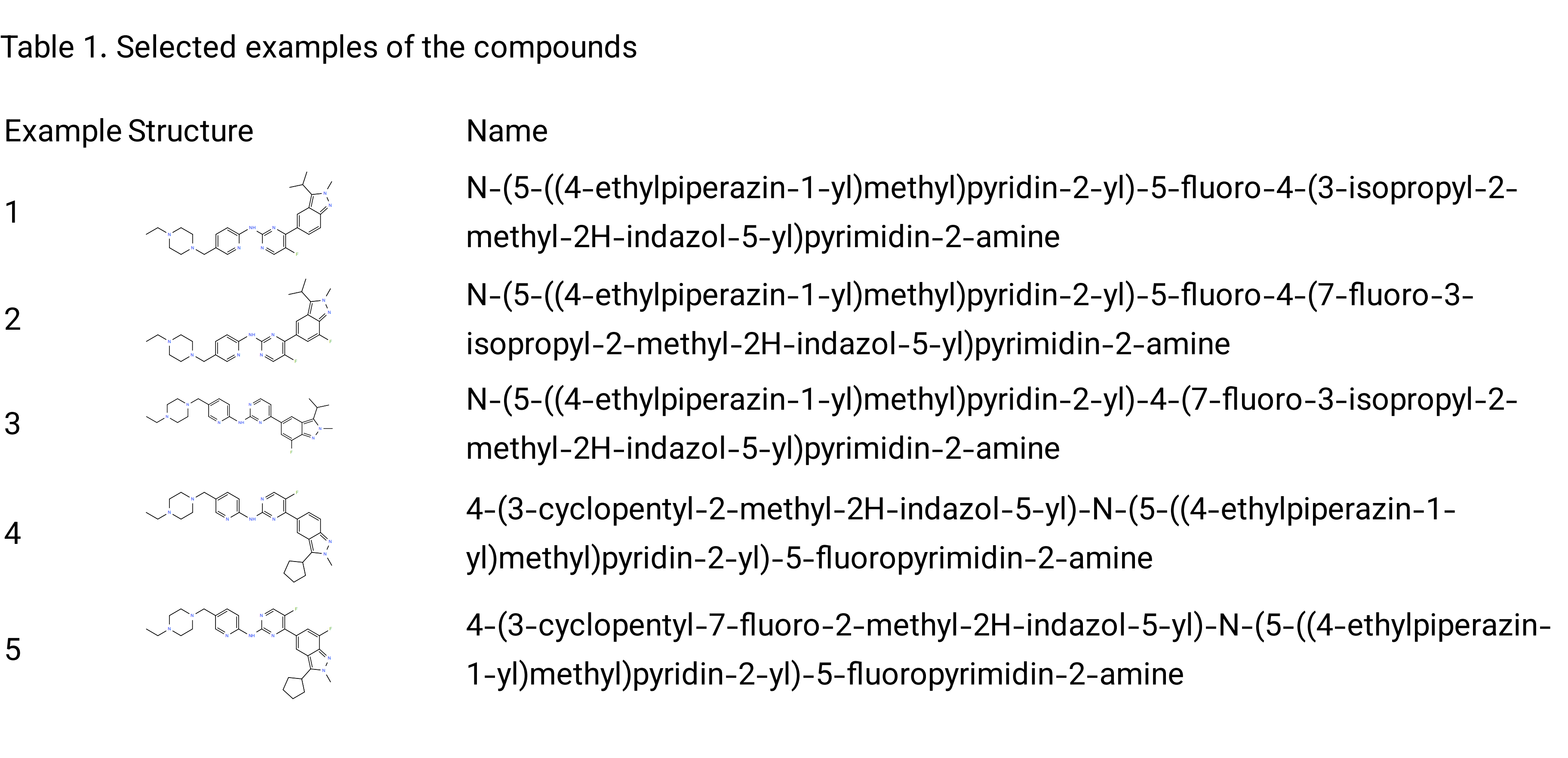}{Uni-Parser}{uniparser-color}

\vspace{12pt} 

\ShowcaseCellBoxed{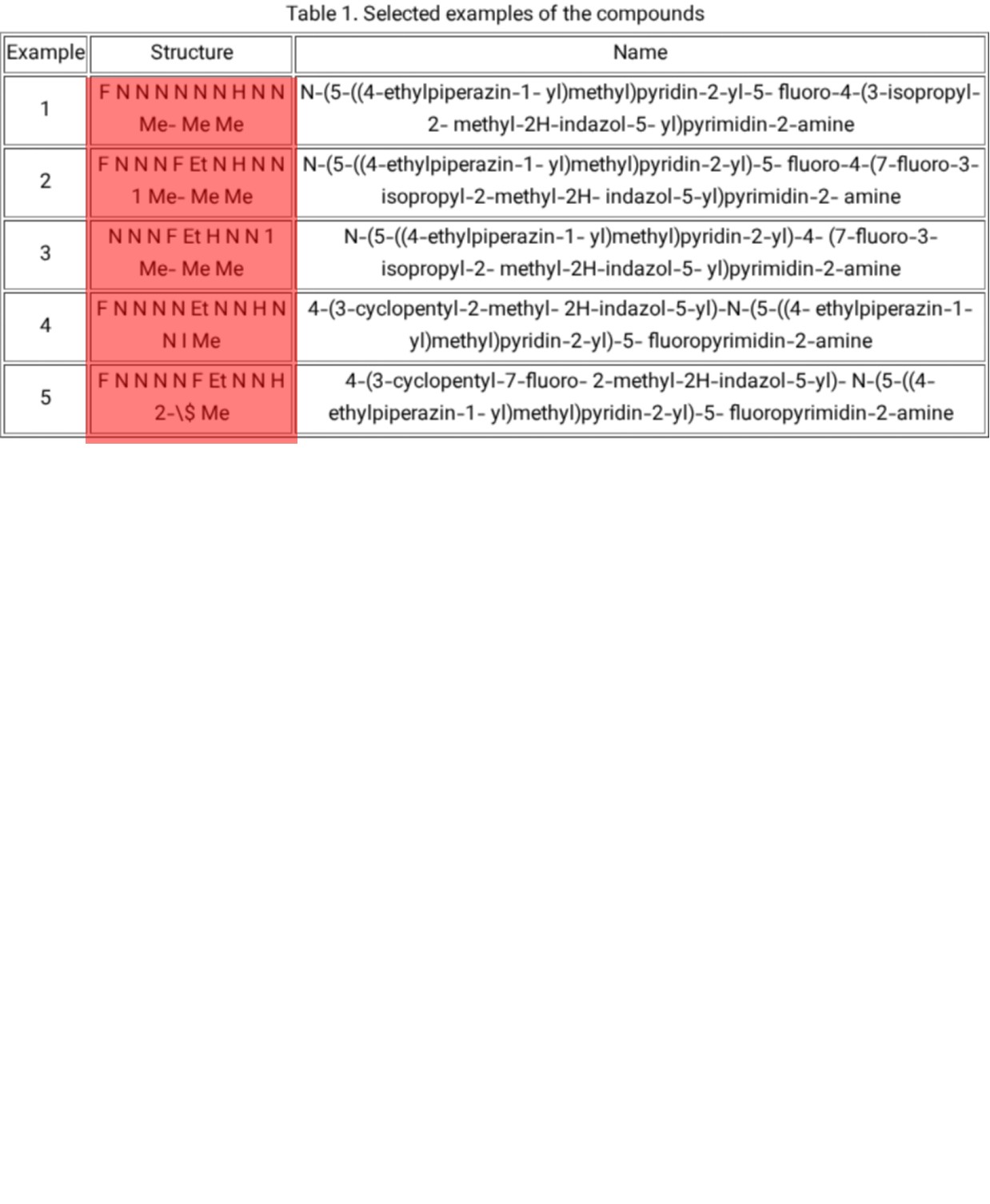}{PP-StructureV3}{gray!80}
\hfill
\ShowcaseCellBoxed{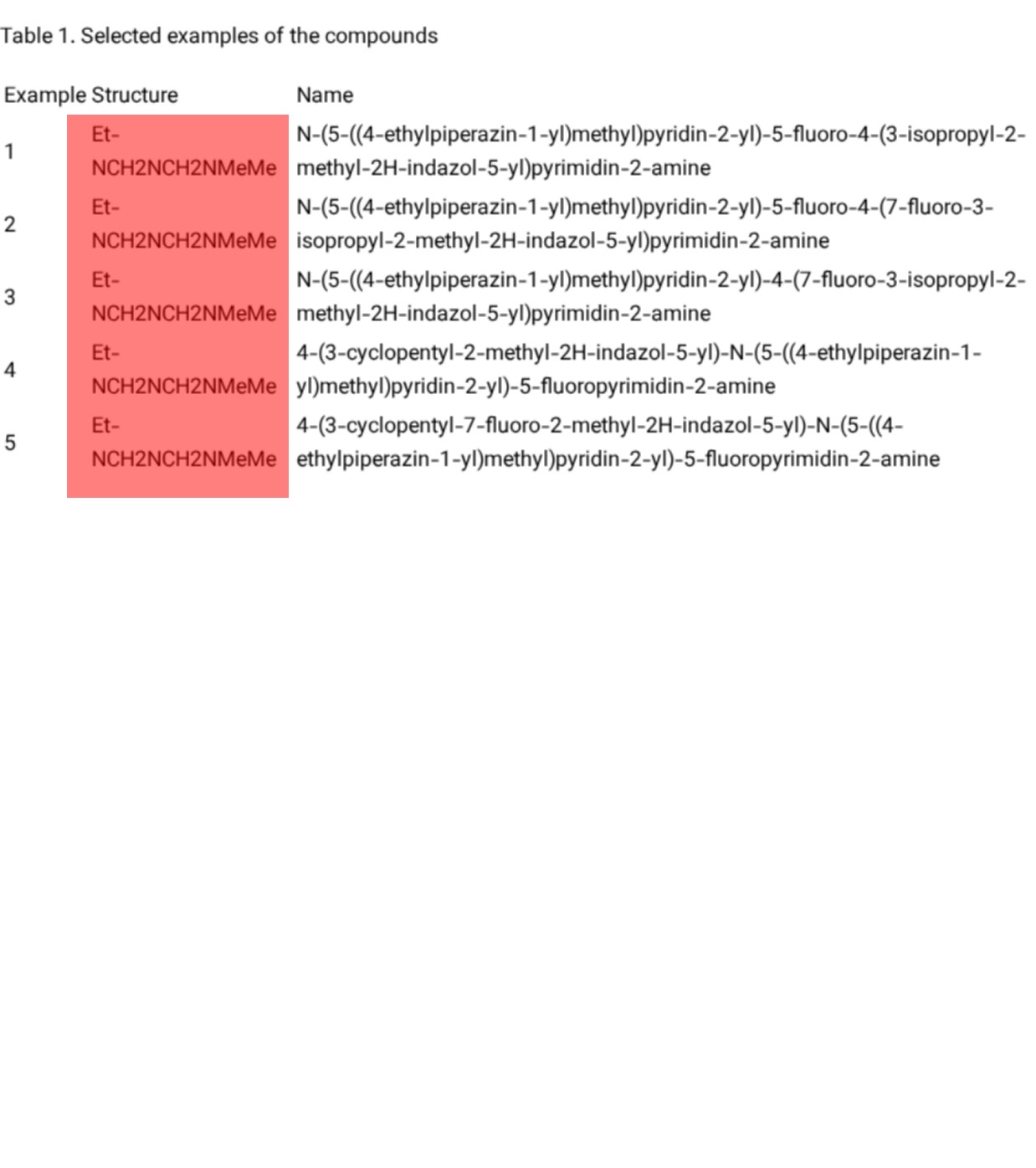}{MinerU2.5}{gray!80}

\end{minipage}
};



\end{tikzpicture}

\end{figure}


\begin{figure}[htpb]
\centering

\begin{tikzpicture}[remember picture]

\node (grid)[inner sep=0pt] {
\begin{minipage}{\textwidth}
\centering

\ShowcaseCellBoxed{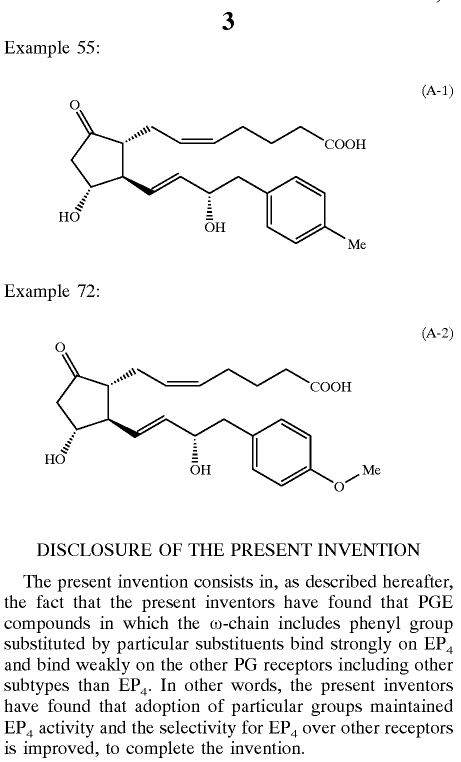}{Original Document}{black!80}
\hfill
\ShowcaseCellBoxed{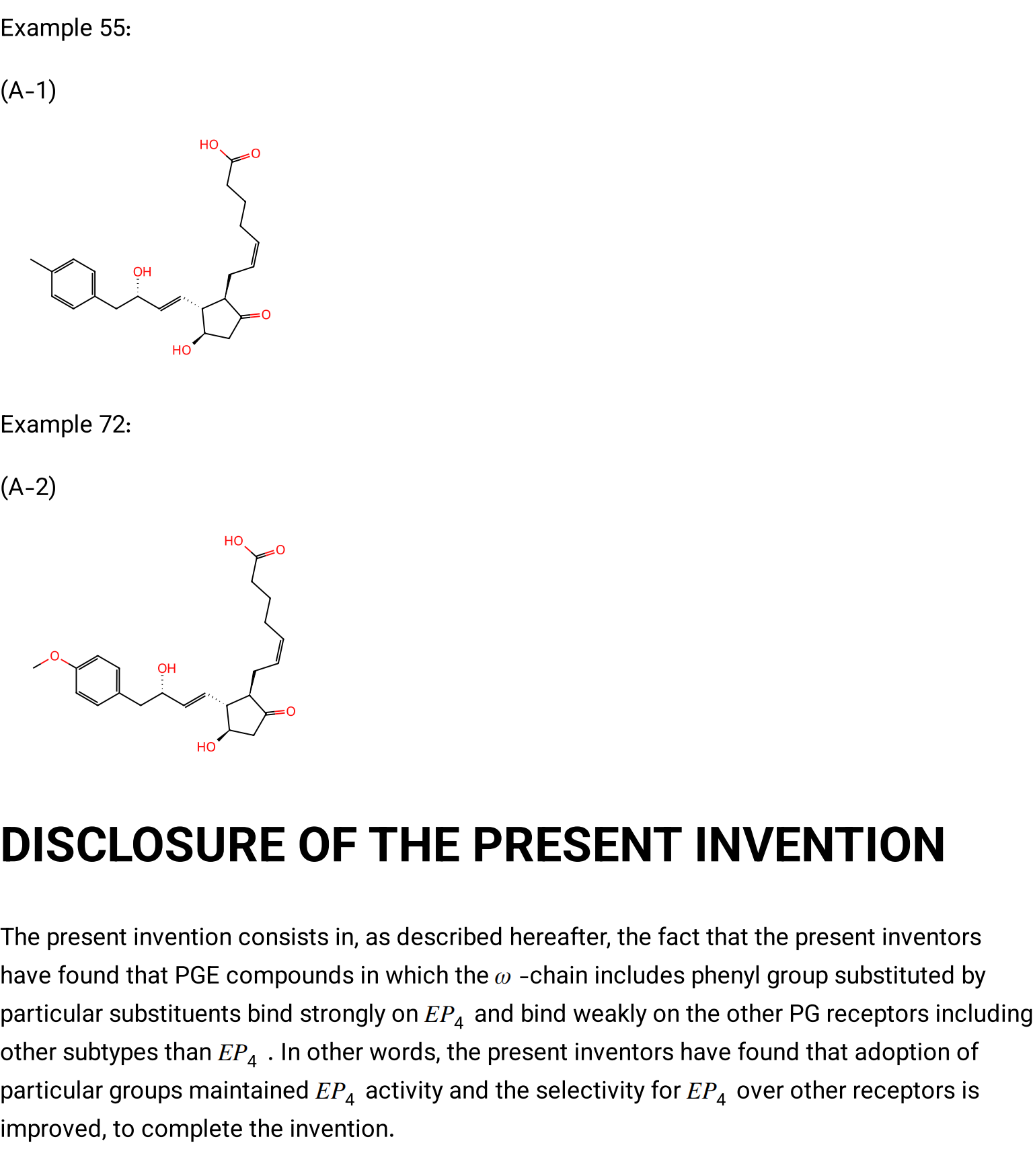}{Uni-Parser}{uniparser-color}

\vspace{12pt} 

\ShowcaseCellBoxed{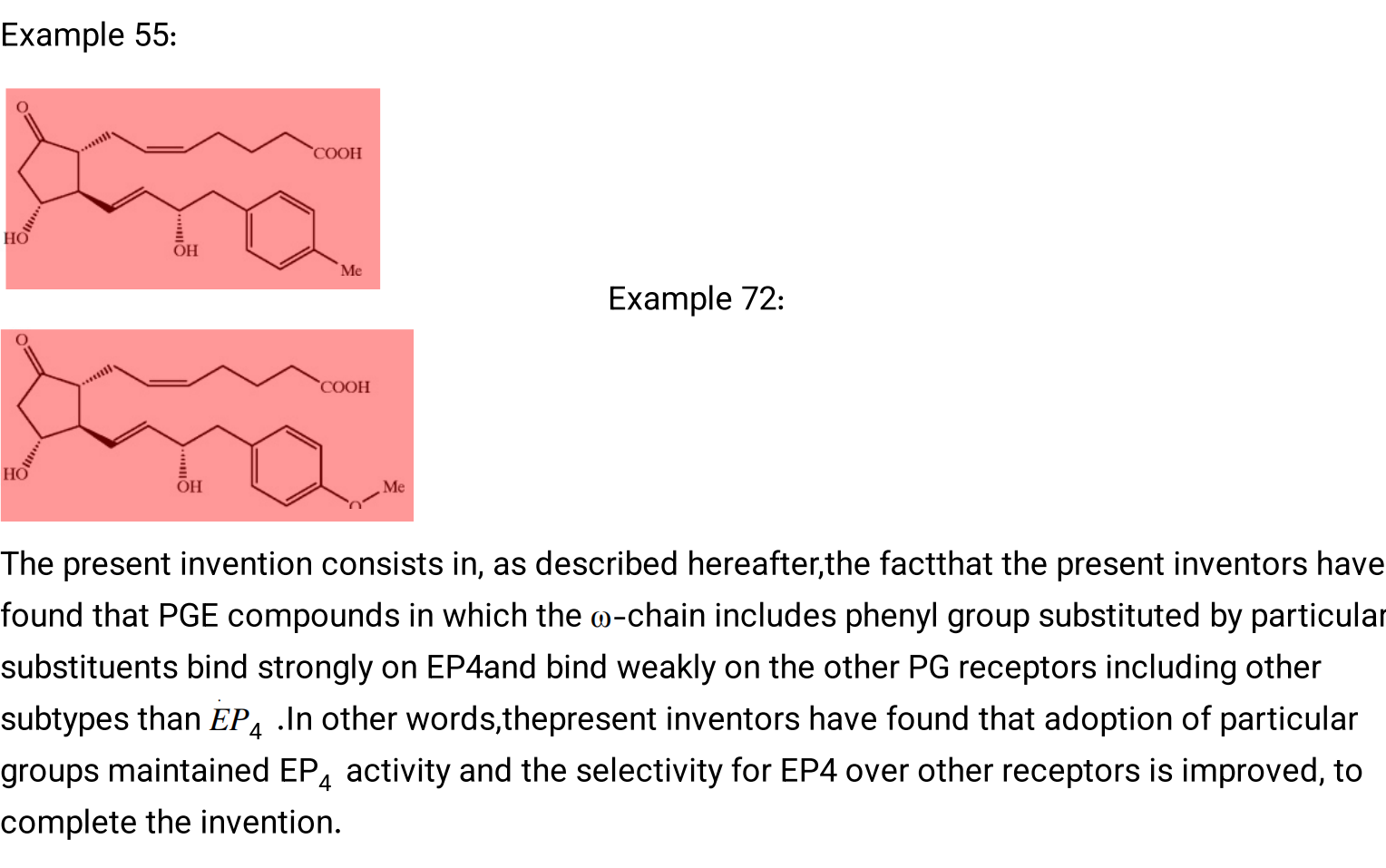}{PP-StructureV3}{gray!80}
\hfill
\ShowcaseCellBoxed{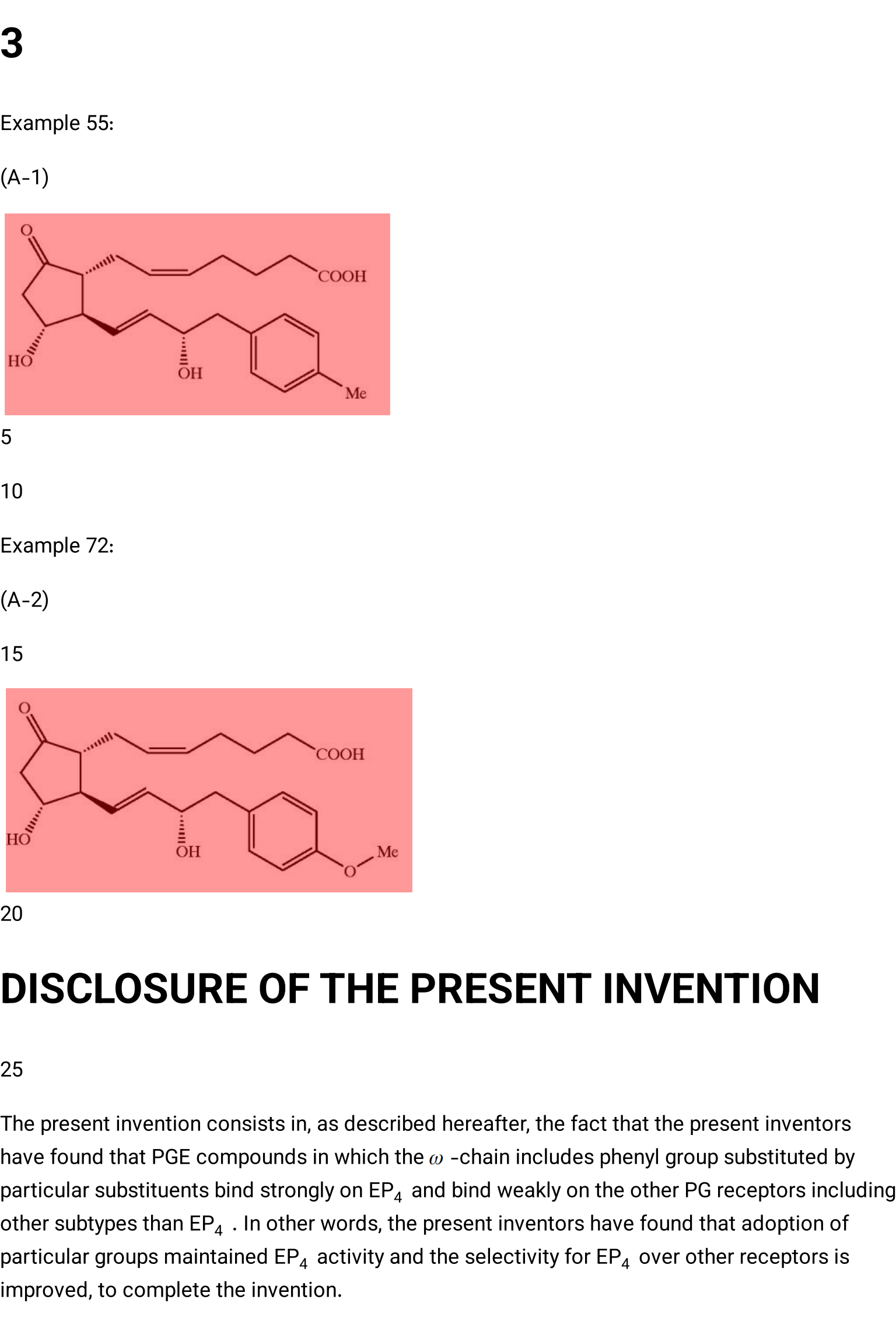}{MinerU2.5}{gray!80}

\end{minipage}
};



\end{tikzpicture}

\end{figure}


\begin{figure}[htpb]
\centering

\begin{tikzpicture}[remember picture]

\node (grid)[inner sep=0pt] {
\begin{minipage}{\textwidth}
\centering

\ShowcaseCellBoxed{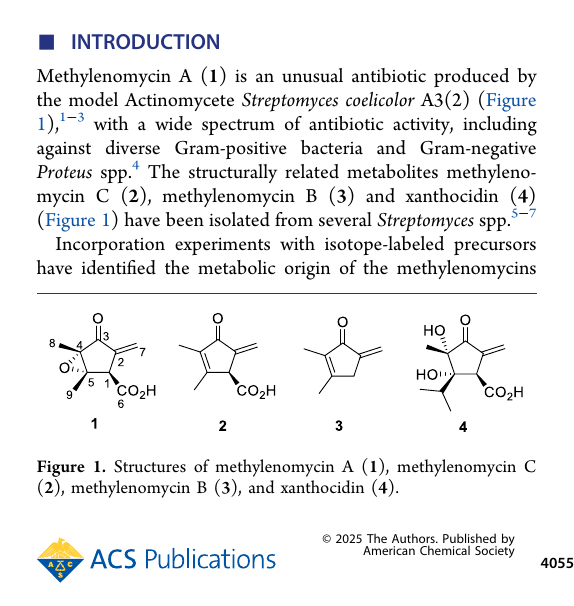}{Original Document}{black!80}
\hfill
\ShowcaseCellBoxed{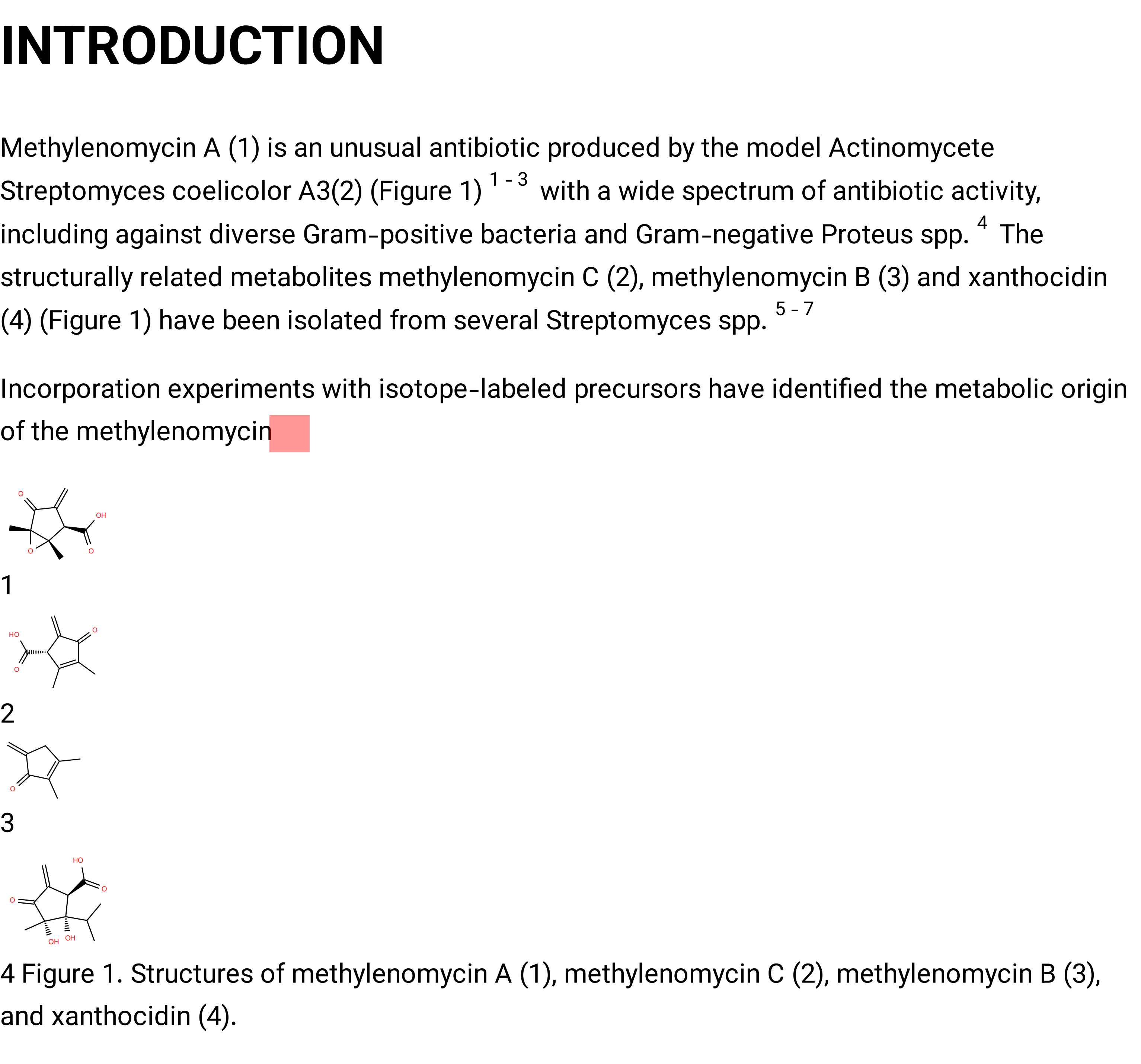}{Uni-Parser}{uniparser-color}

\vspace{12pt} 

\ShowcaseCellBoxed{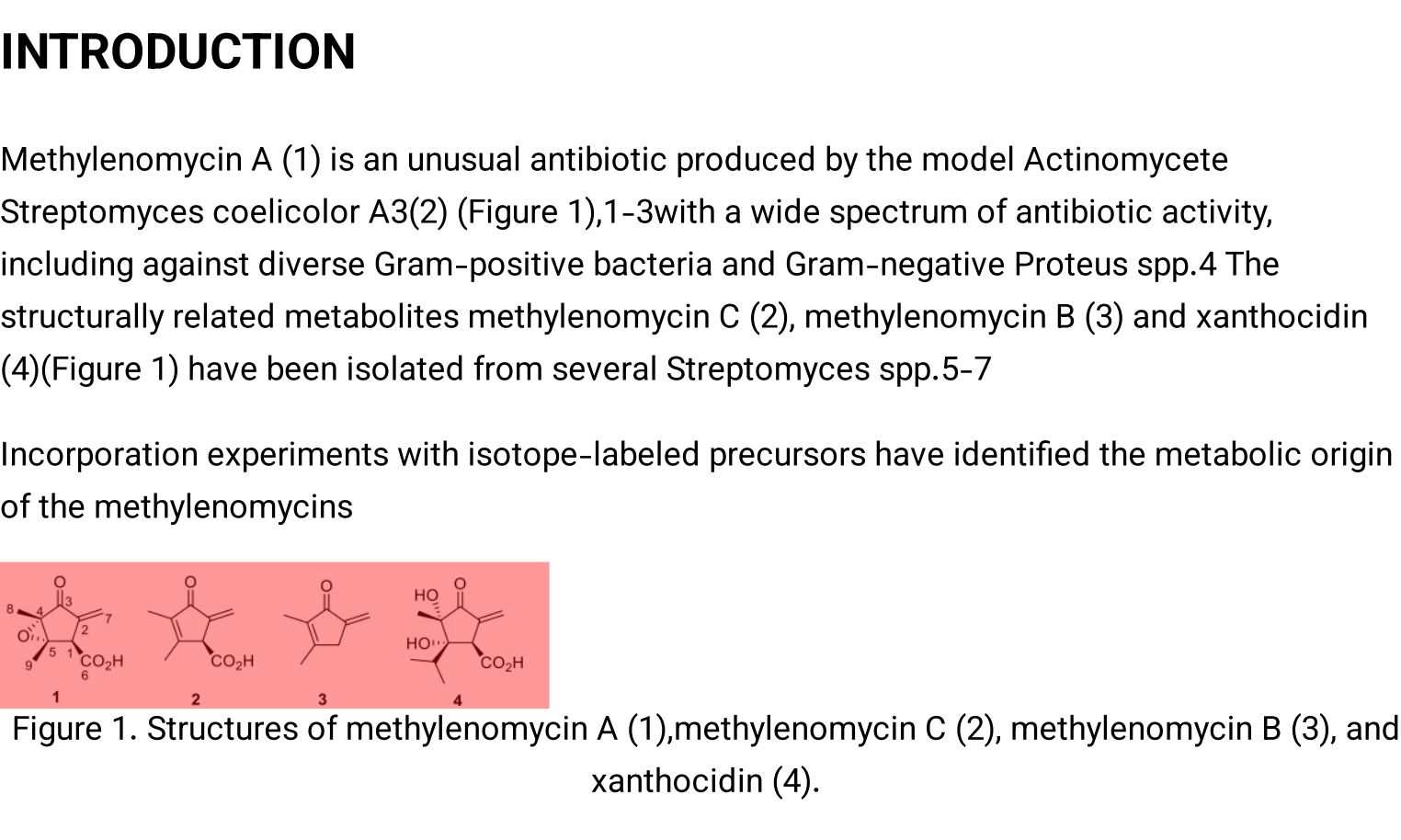}{PP-StructureV3}{gray!80}
\hfill
\ShowcaseCellBoxed{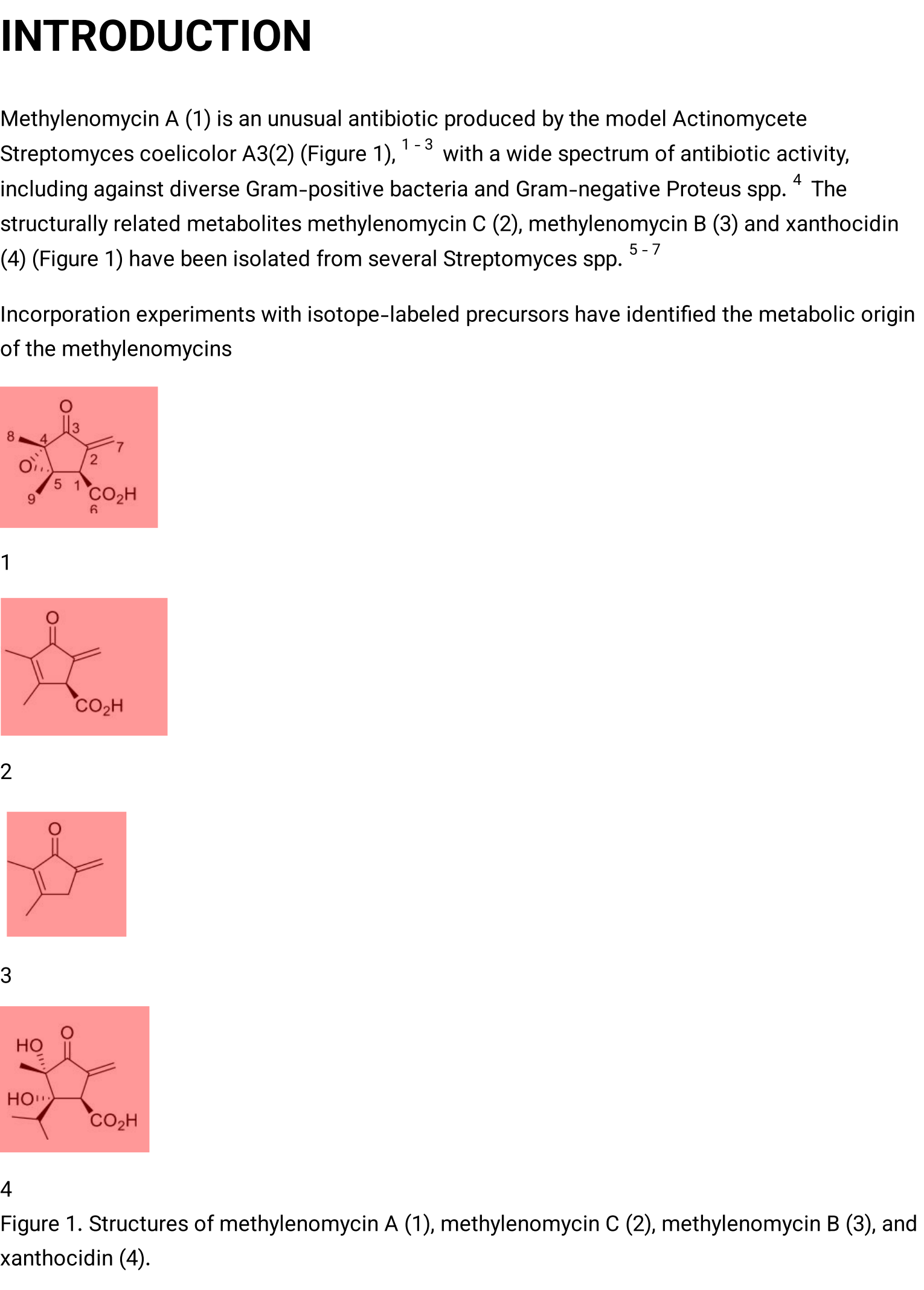}{MinerU2.5}{gray!80}

\end{minipage}
};



\end{tikzpicture}

\end{figure}


\begin{figure}[htpb]
\centering

\begin{tikzpicture}[remember picture]

\node (grid)[inner sep=0pt] {
\begin{minipage}{\textwidth}
\centering

\ShowcaseCellBoxed{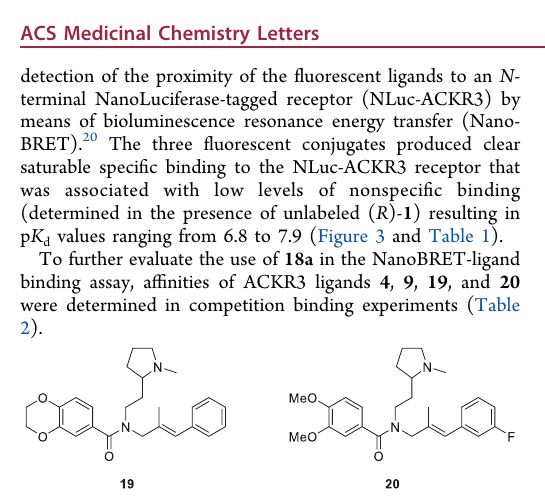}{Original Document}{black!80}
\hfill
\ShowcaseCellBoxed{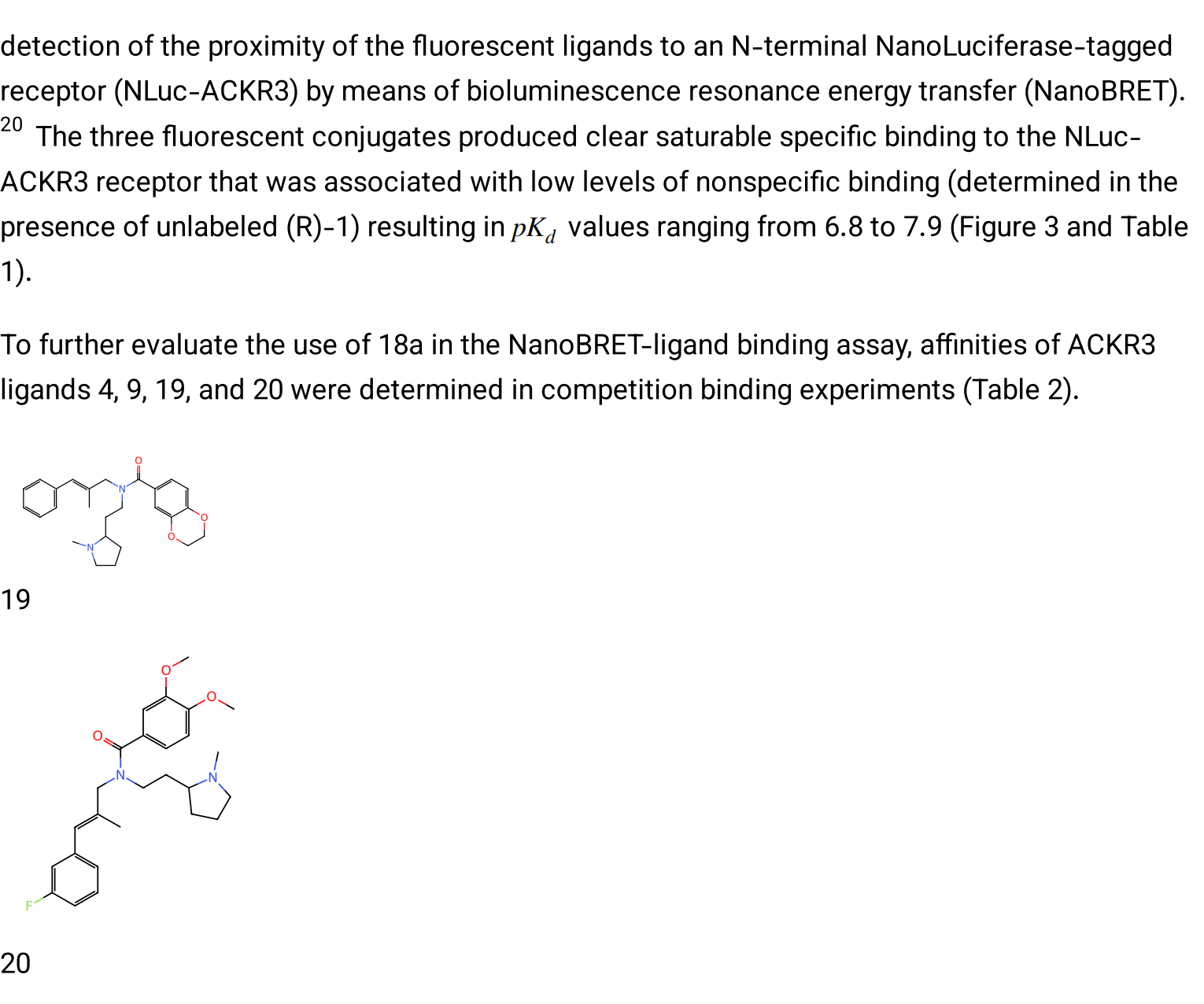}{Uni-Parser}{uniparser-color}

\vspace{12pt} 

\ShowcaseCellBoxed{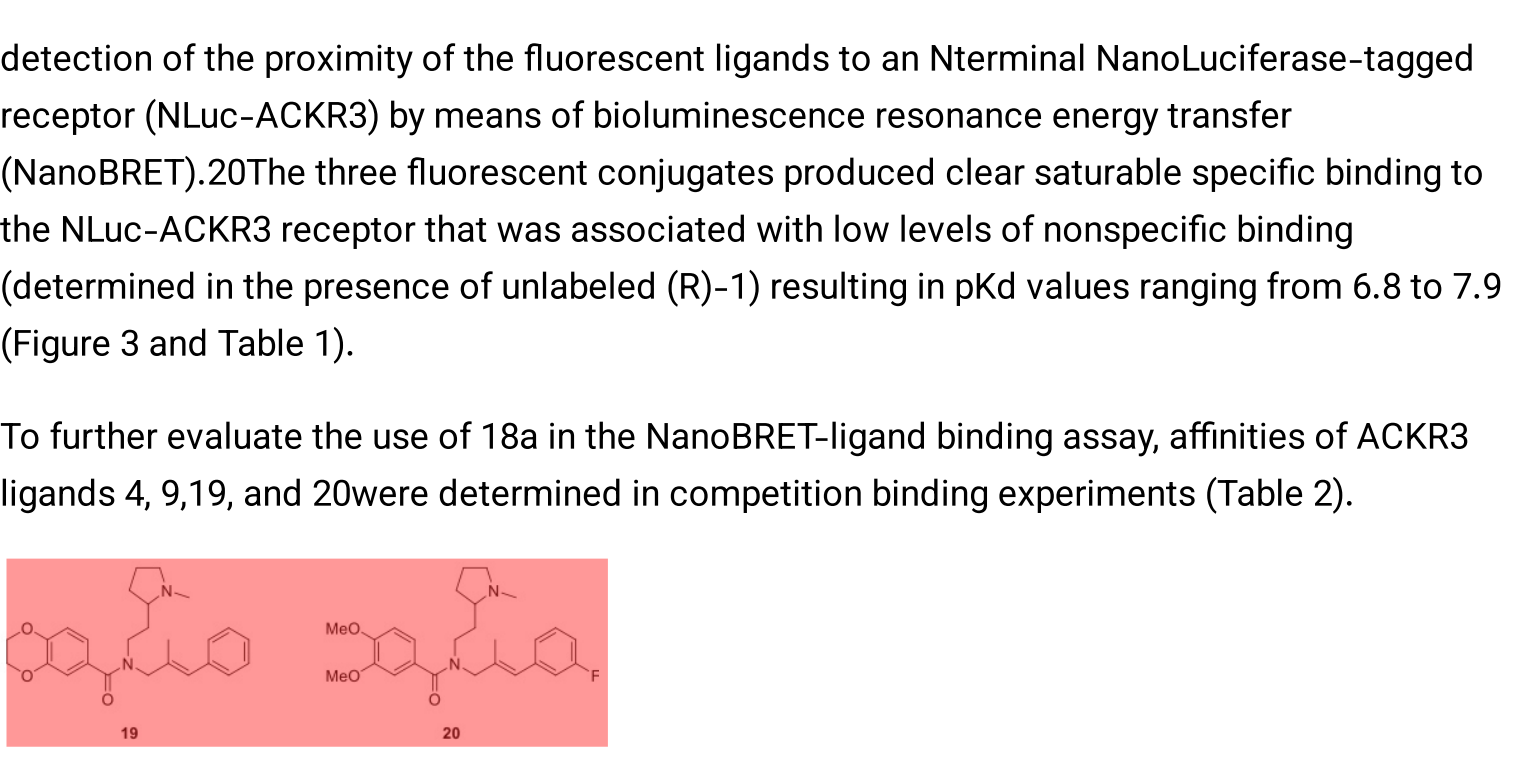}{PP-StructureV3}{gray!80}
\hfill
\ShowcaseCellBoxed{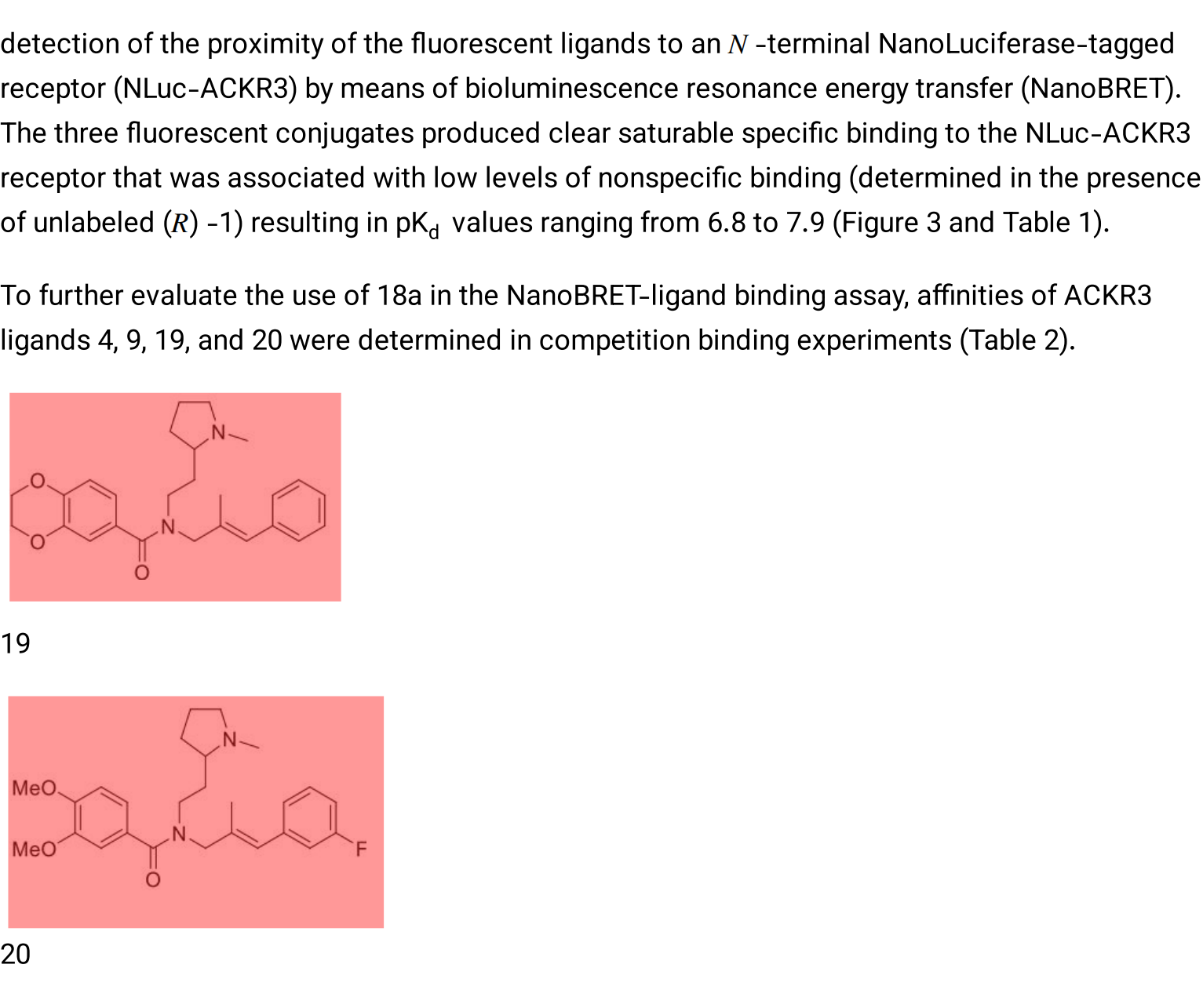}{MinerU2.5}{gray!80}

\end{minipage}
};



\end{tikzpicture}

\end{figure}


\begin{figure}[htpb]
\centering

\begin{tikzpicture}[remember picture]

\node (grid)[inner sep=0pt] {
\begin{minipage}{\textwidth}
\centering

\ShowcaseCellBoxed{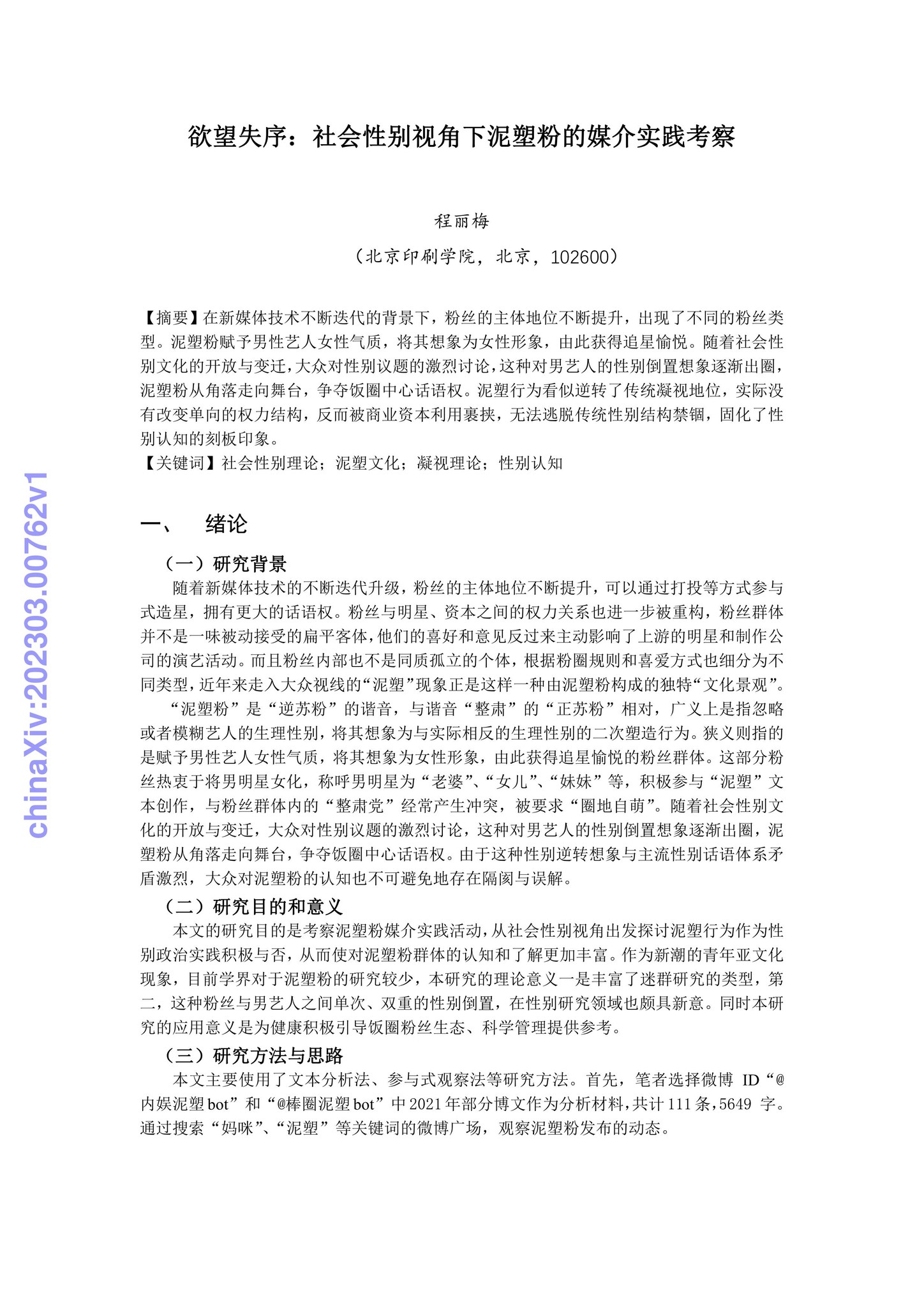}{Original Document}{black!80}
\hfill
\ShowcaseCellBoxed{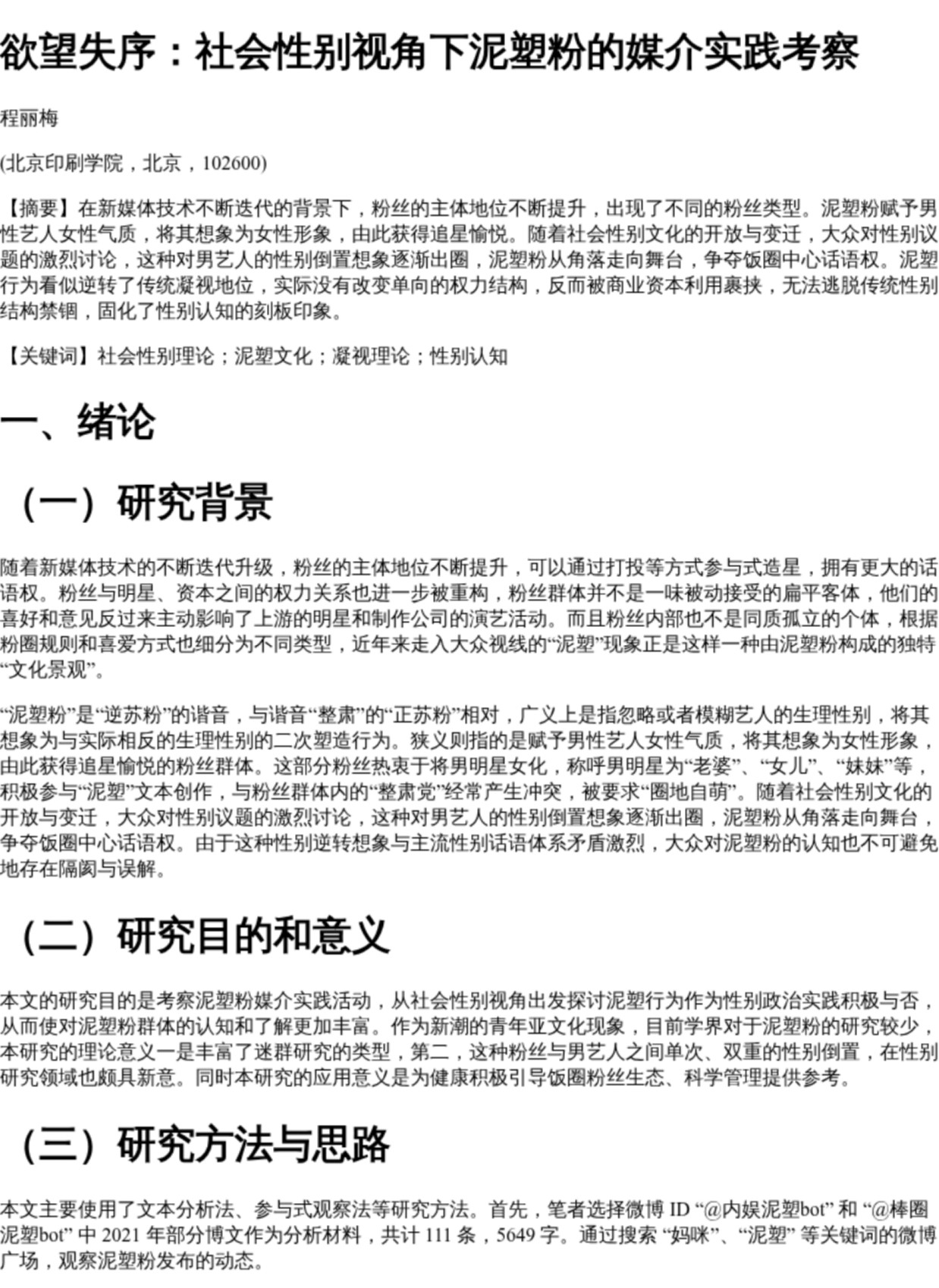}{Uni-Parser}{uniparser-color}

\vspace{12pt} 

\ShowcaseCellBoxed{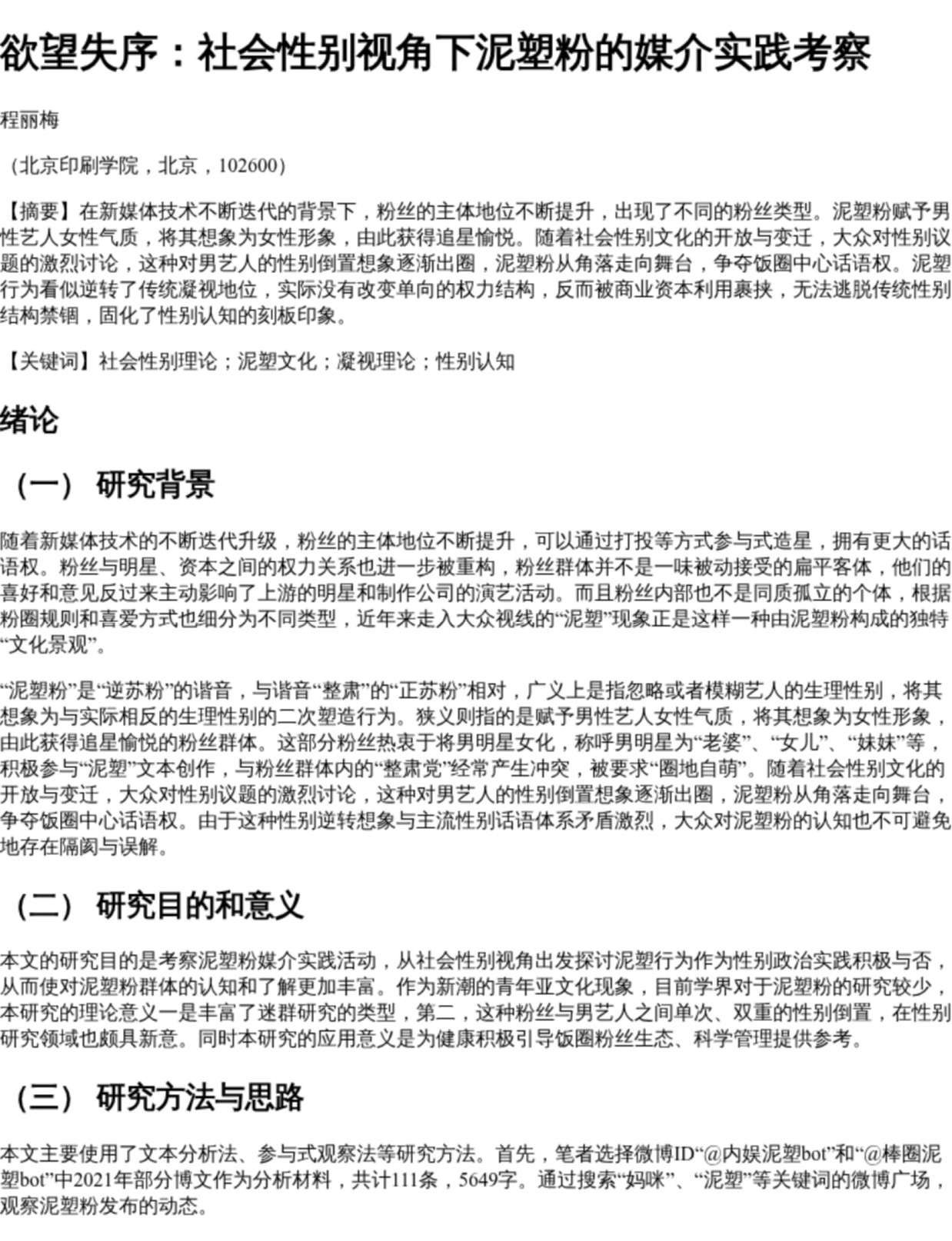}{PP-StructureV3}{gray!80}
\hfill
\ShowcaseCellBoxed{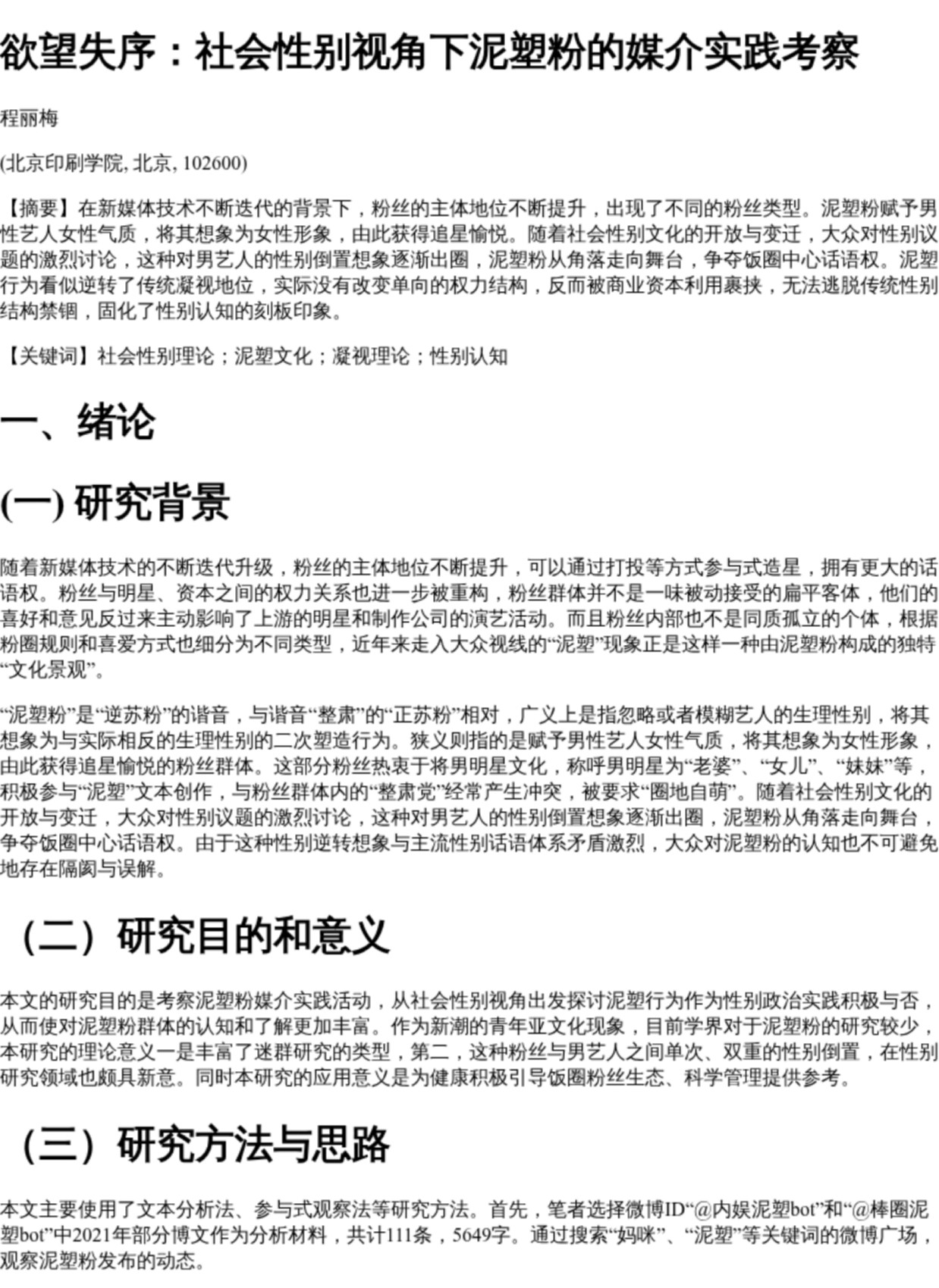}{MinerU2.5}{gray!80}

\end{minipage}
};



\end{tikzpicture}

\end{figure}


\begin{figure}[htpb]
\centering

\begin{tikzpicture}[remember picture]

\node (grid)[inner sep=0pt] {
\begin{minipage}{\textwidth}
\centering

\ShowcaseCellBoxed{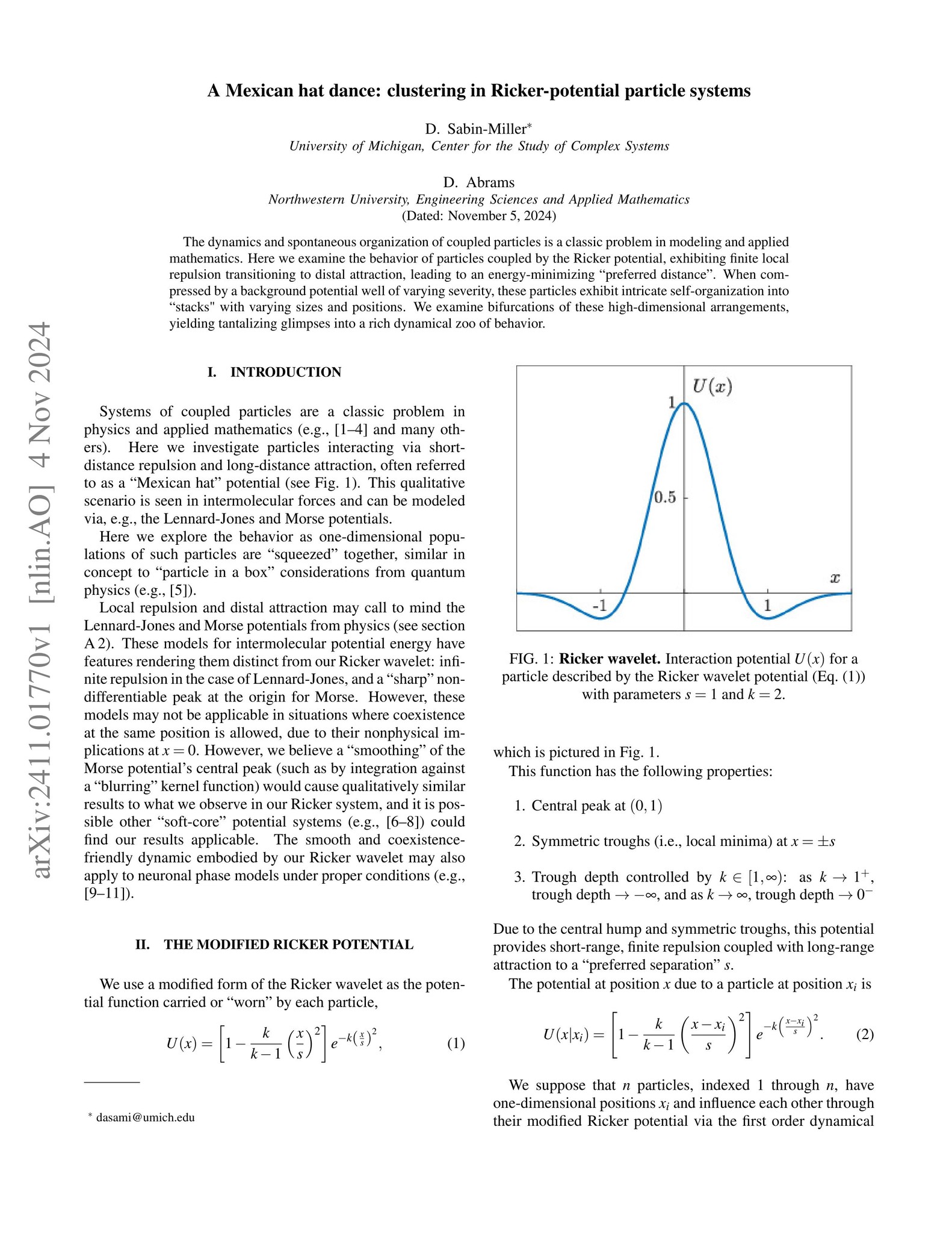}{Original Document}{black!80}
\hfill
\ShowcaseCellBoxed{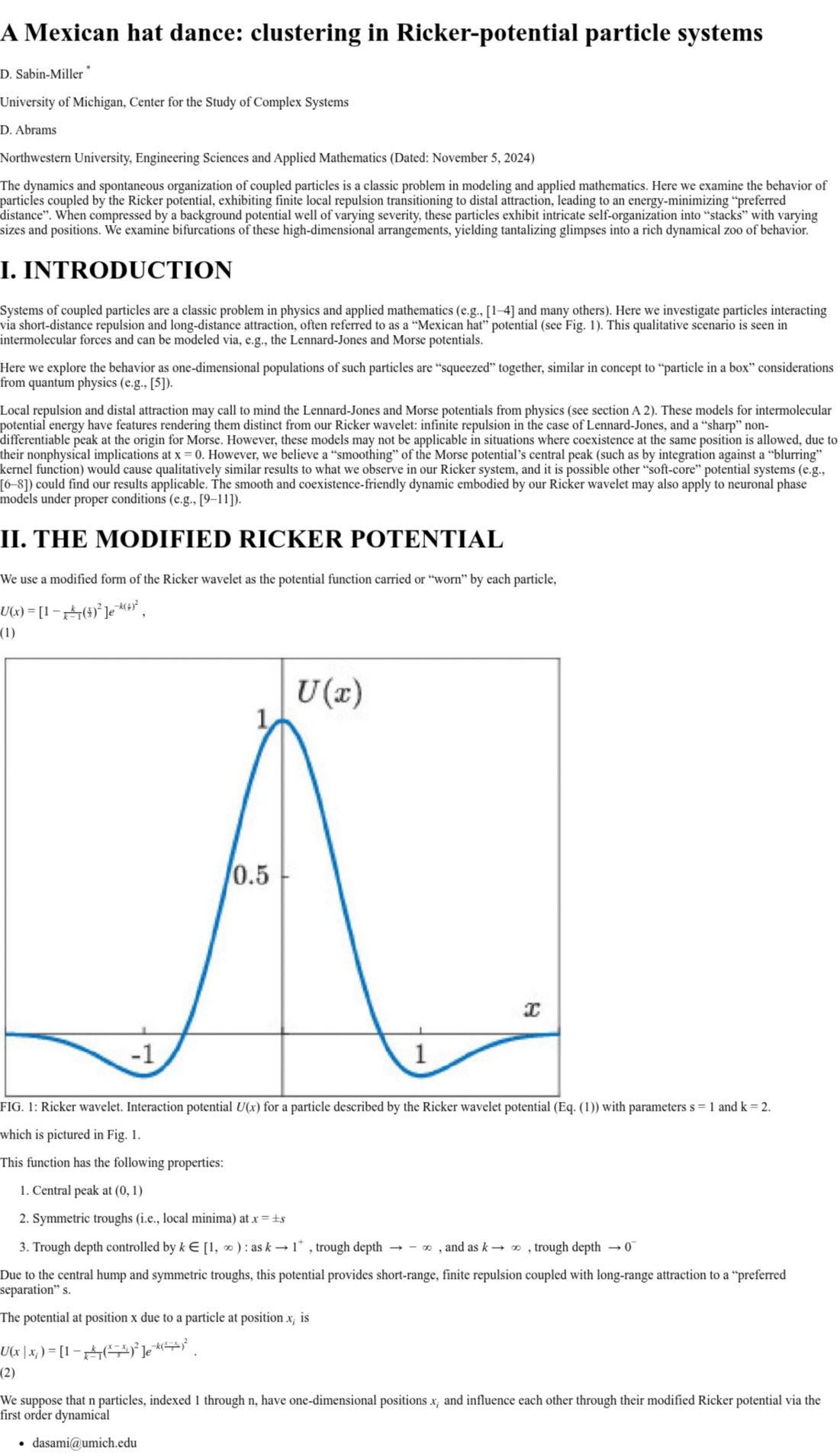}{Uni-Parser}{uniparser-color}

\vspace{12pt} 

\ShowcaseCellBoxed{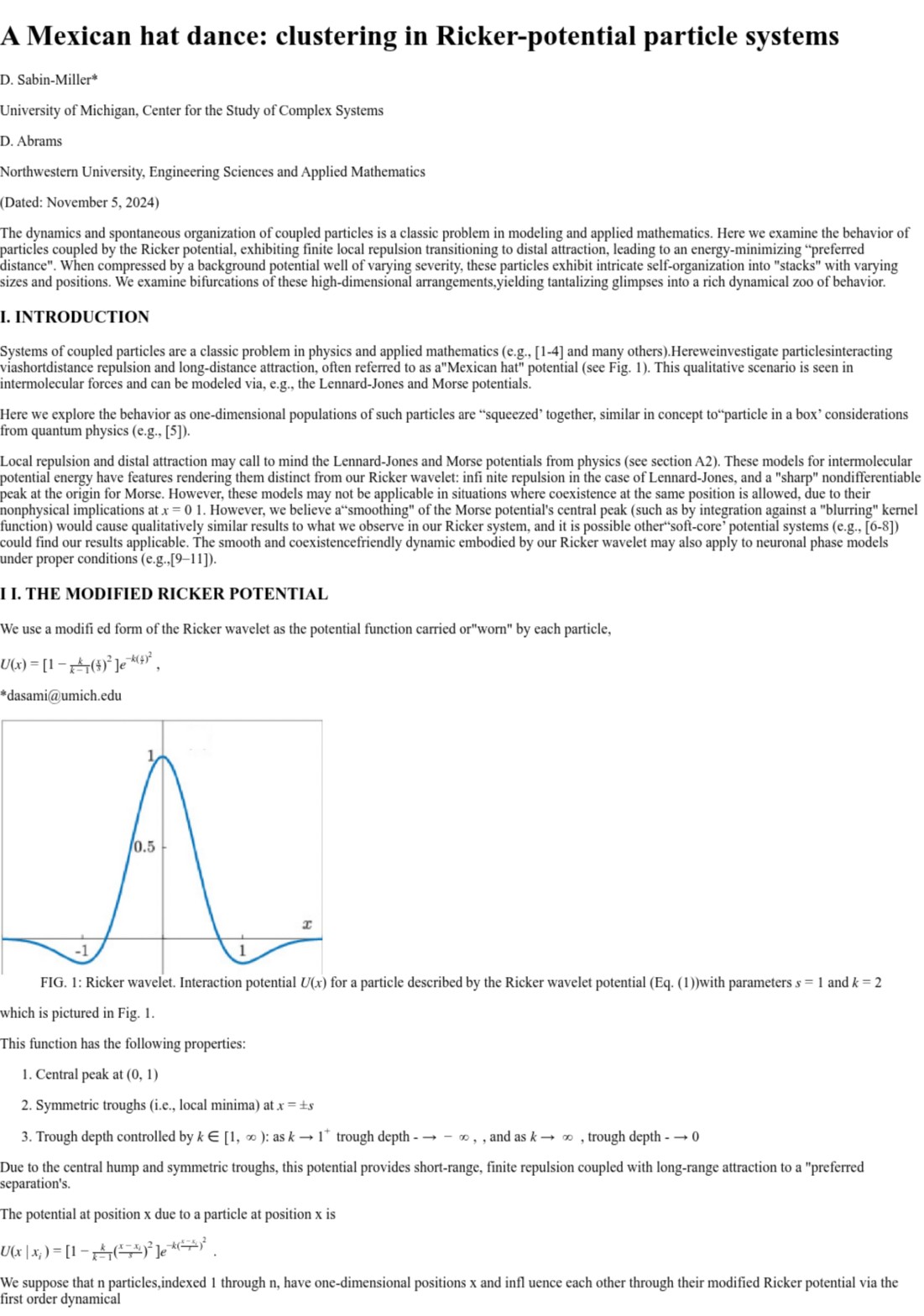}{PP-StructureV3}{gray!80}
\hfill
\ShowcaseCellBoxed{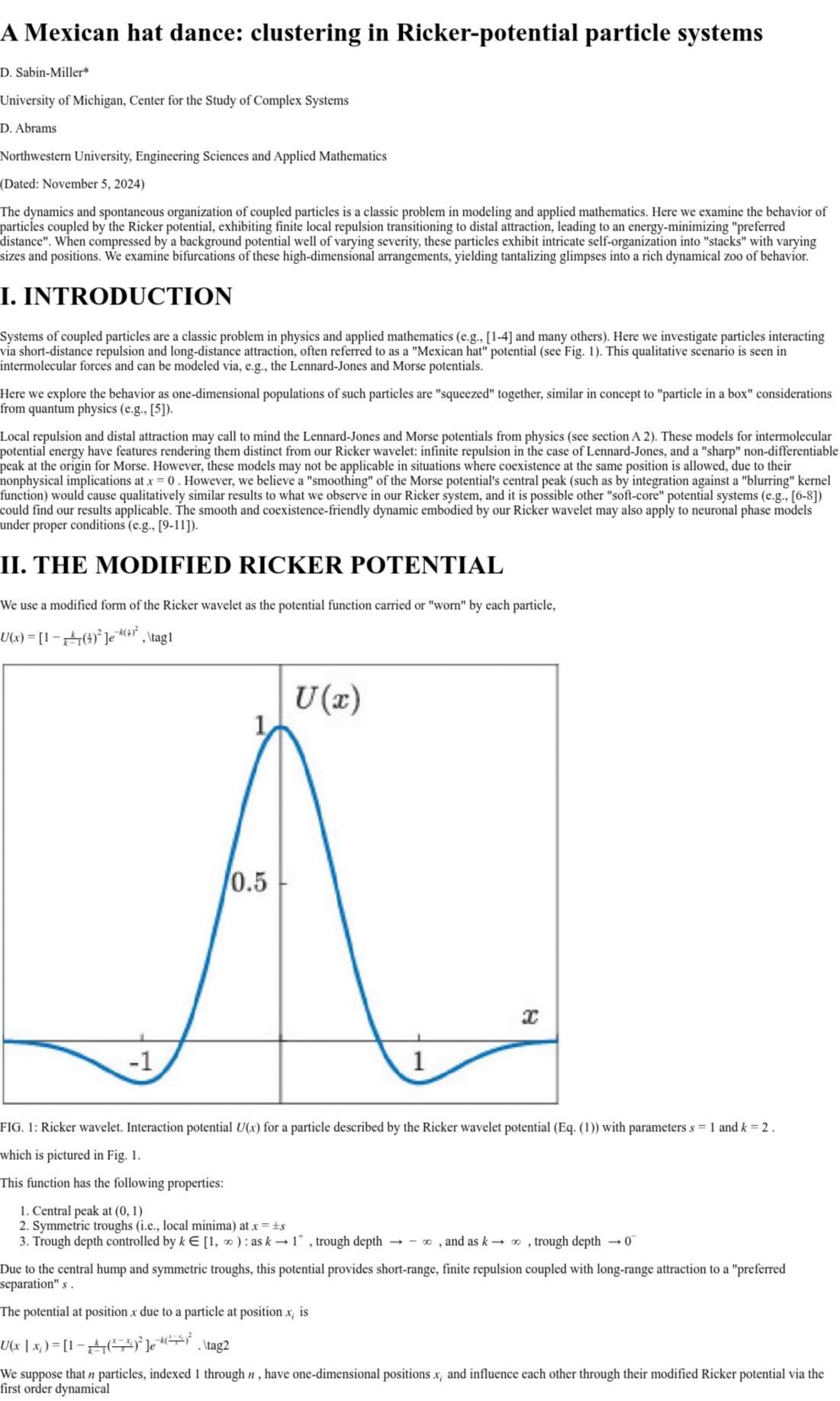}{MinerU2.5}{gray!80}

\end{minipage}
};



\end{tikzpicture}

\end{figure}


\begin{figure}[htpb]
\centering

\begin{tikzpicture}[remember picture]

\node (grid)[inner sep=0pt] {
\begin{minipage}{\textwidth}
\centering

\ShowcaseCellBoxed{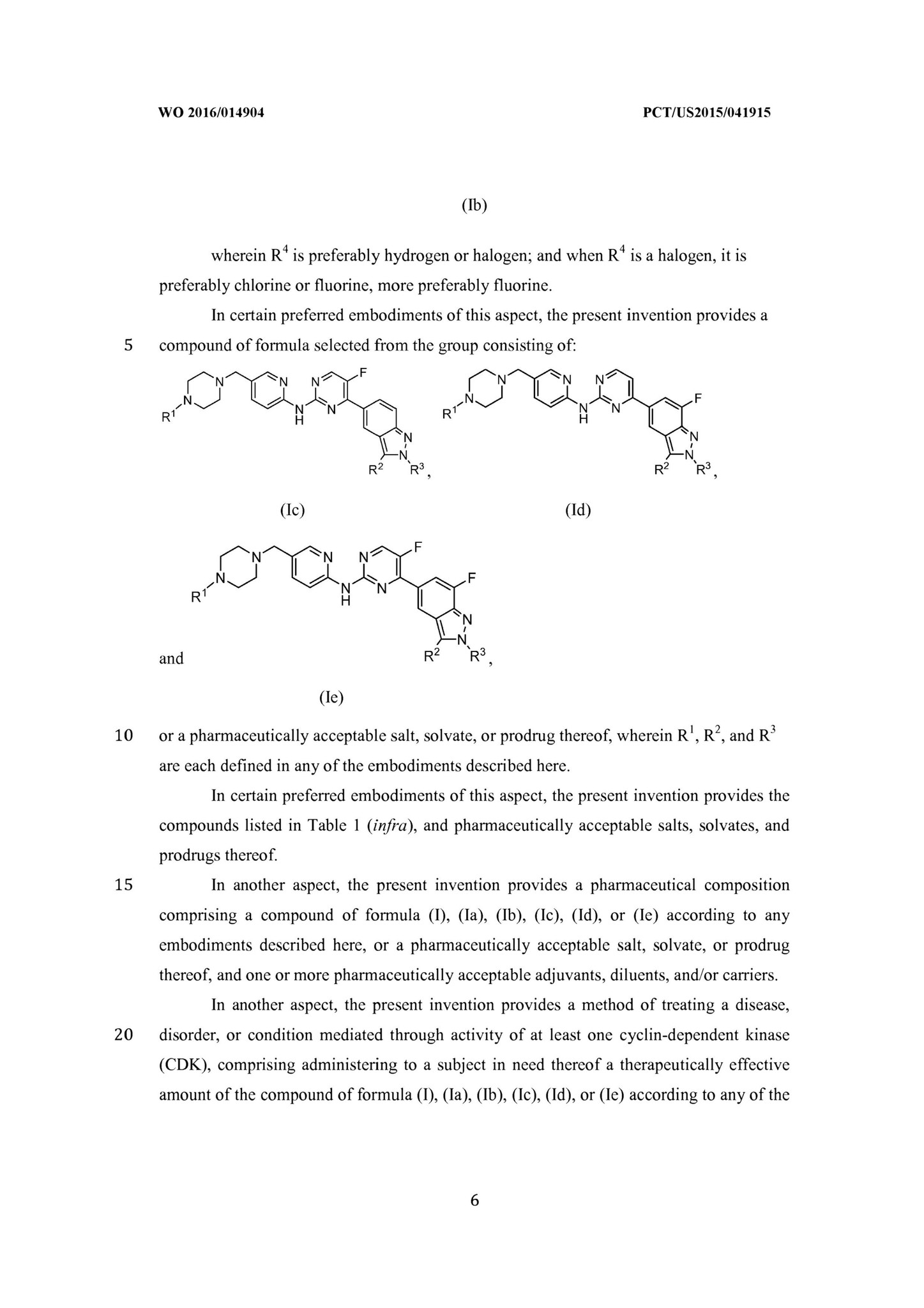}{Original Document}{black!80}
\hfill
\ShowcaseCellBoxed{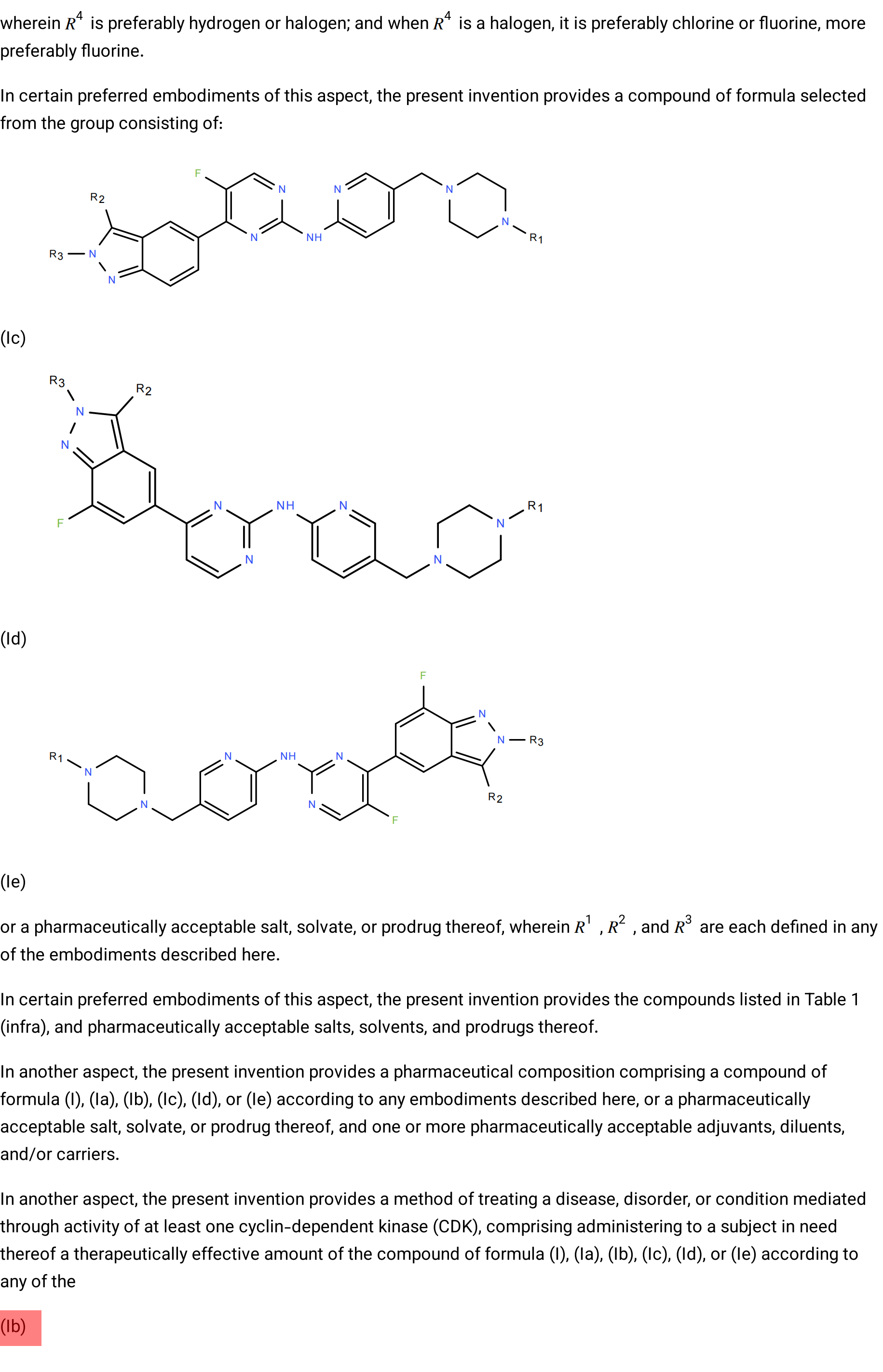}{Uni-Parser}{uniparser-color}

\vspace{12pt} 

\ShowcaseCellBoxed{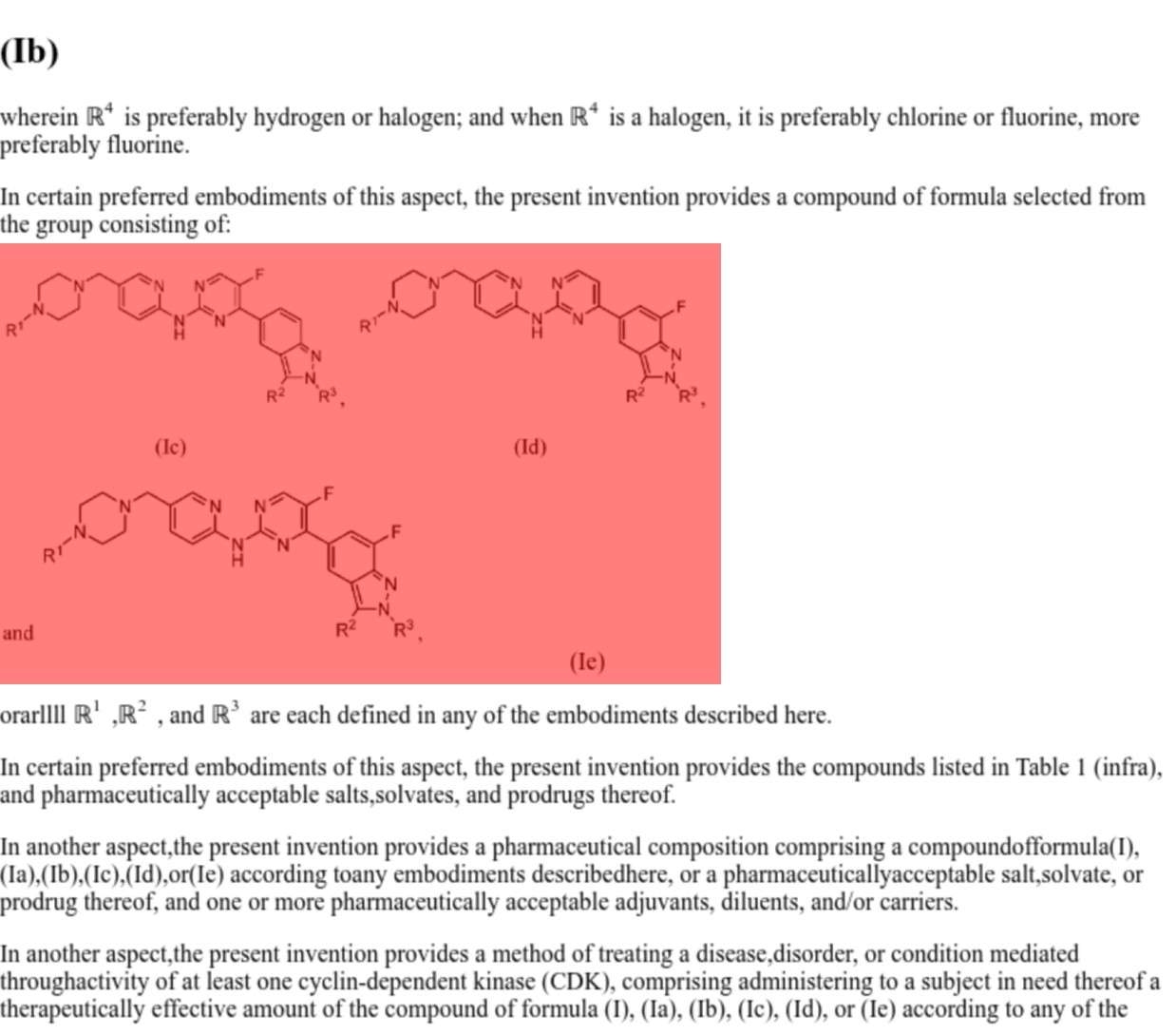}{PP-StructureV3}{gray!80}
\hfill
\ShowcaseCellBoxed{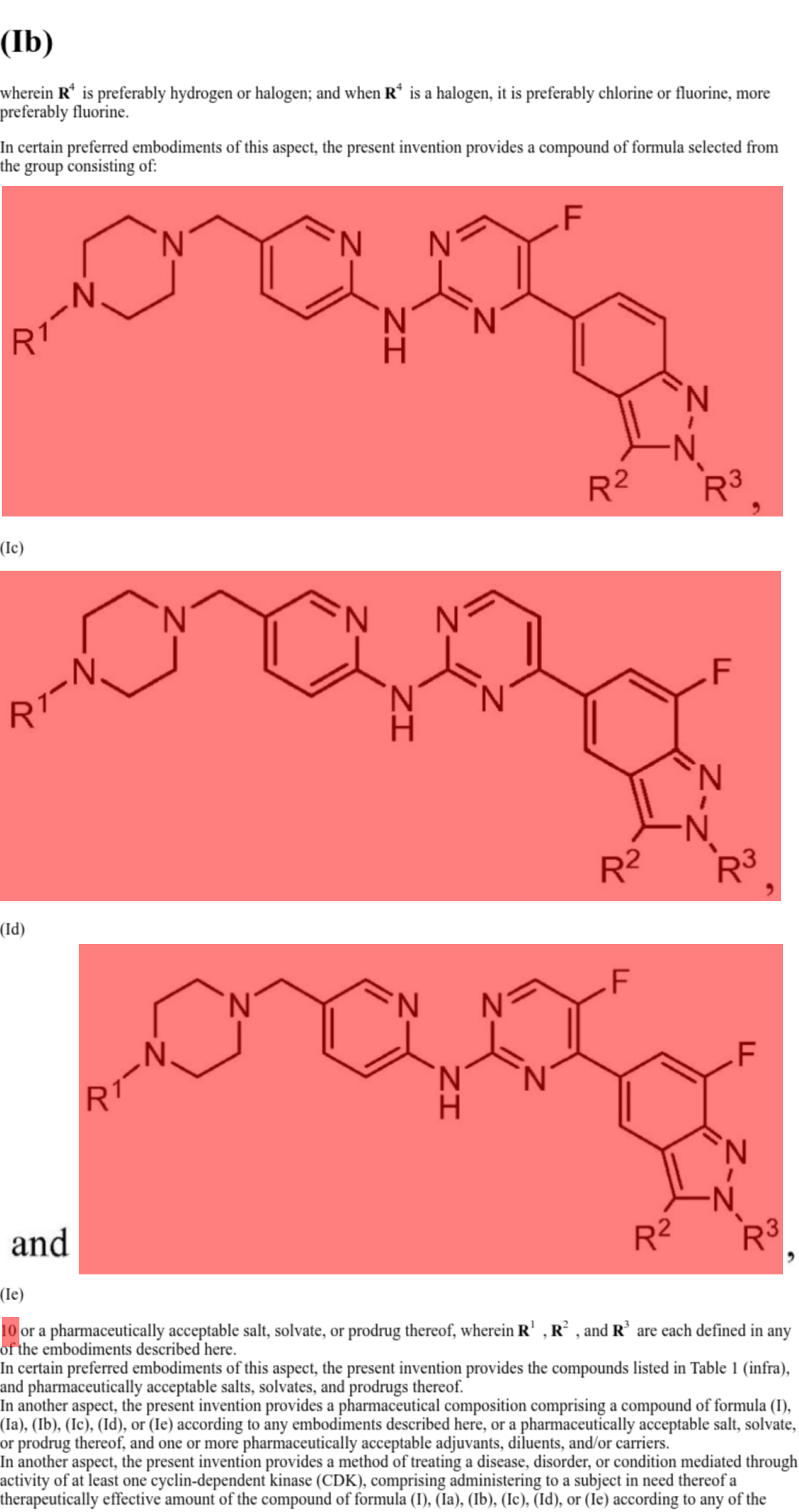}{MinerU2.5}{gray!80}

\end{minipage}
};



\end{tikzpicture}

\end{figure}


\begin{figure}[htpb]
\centering

\begin{tikzpicture}[remember picture]

\node (grid)[inner sep=0pt] {
\begin{minipage}{\textwidth}
\centering

\ShowcaseCellBoxed{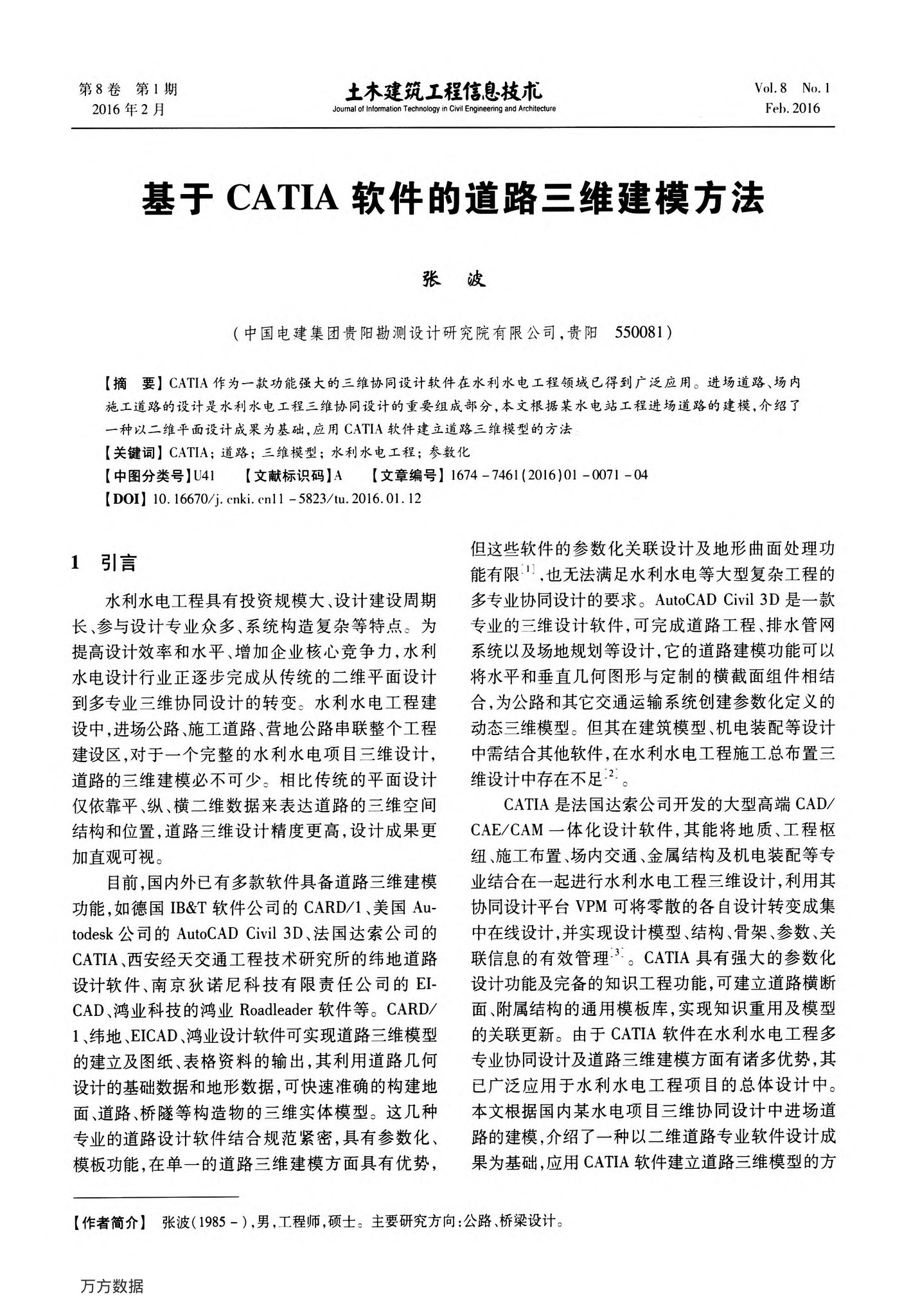}{Original Document}{black!80}
\hfill
\ShowcaseCellBoxed{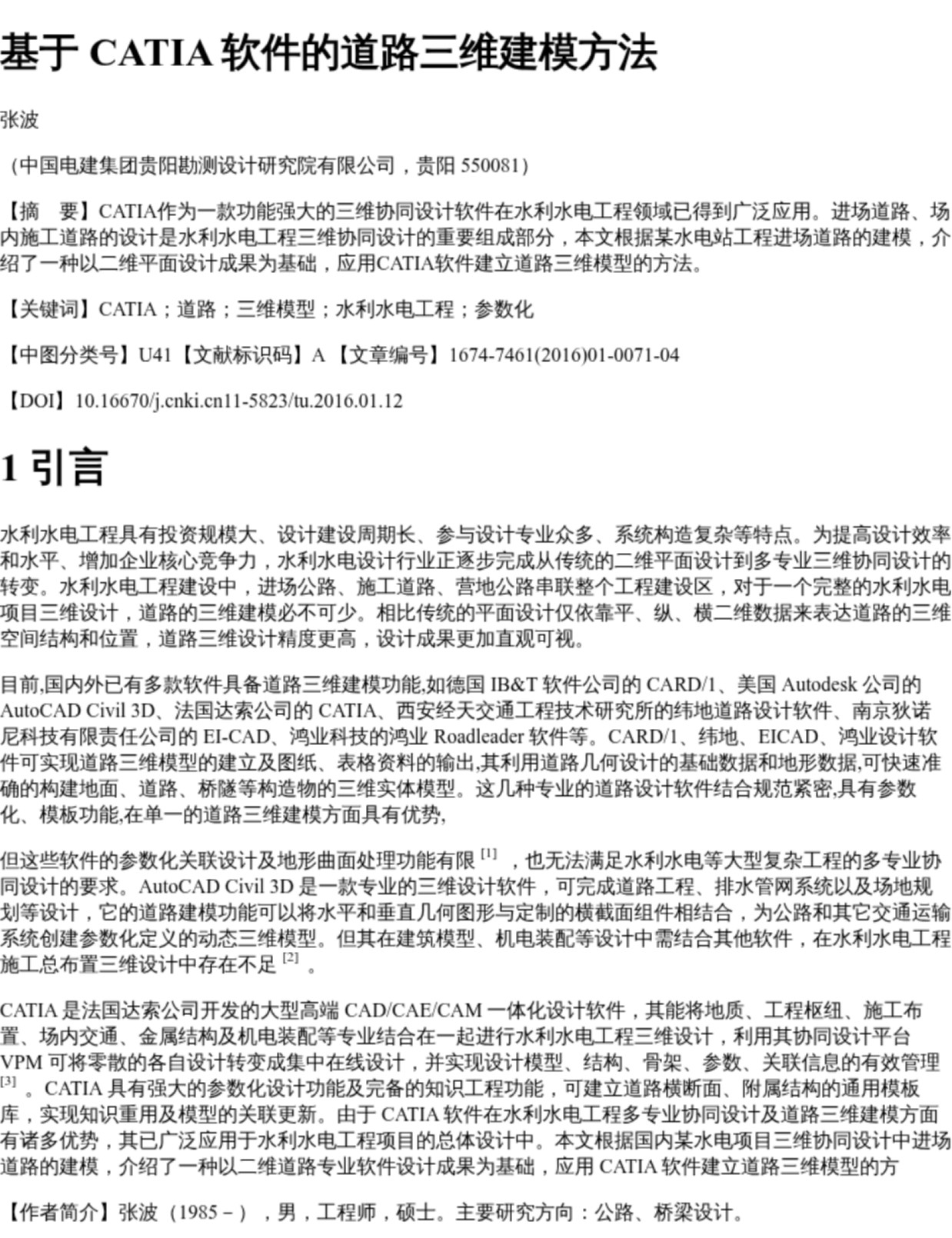}{Uni-Parser}{uniparser-color}

\vspace{12pt} 

\ShowcaseCellBoxed{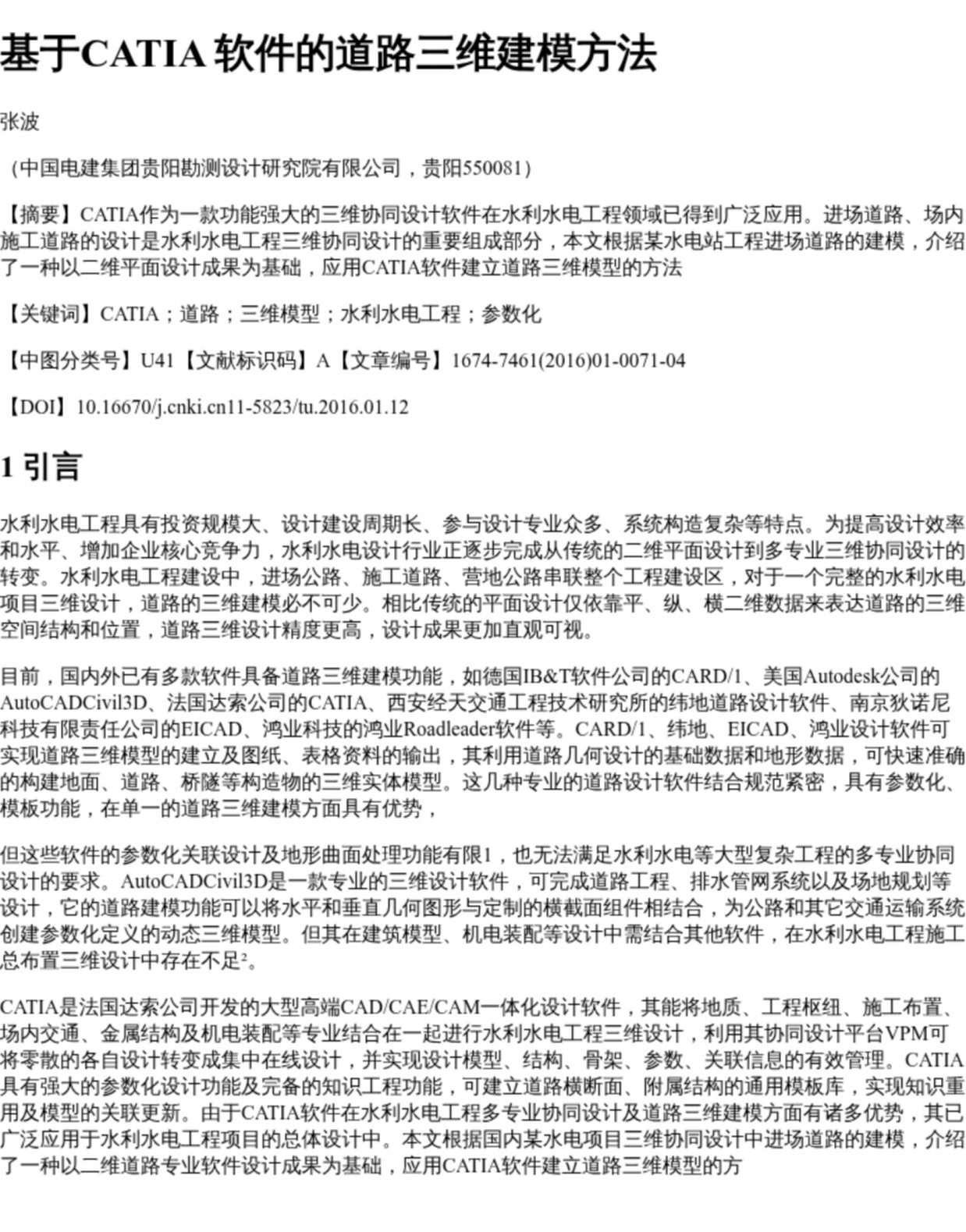}{PP-StructureV3}{gray!80}
\hfill
\ShowcaseCellBoxed{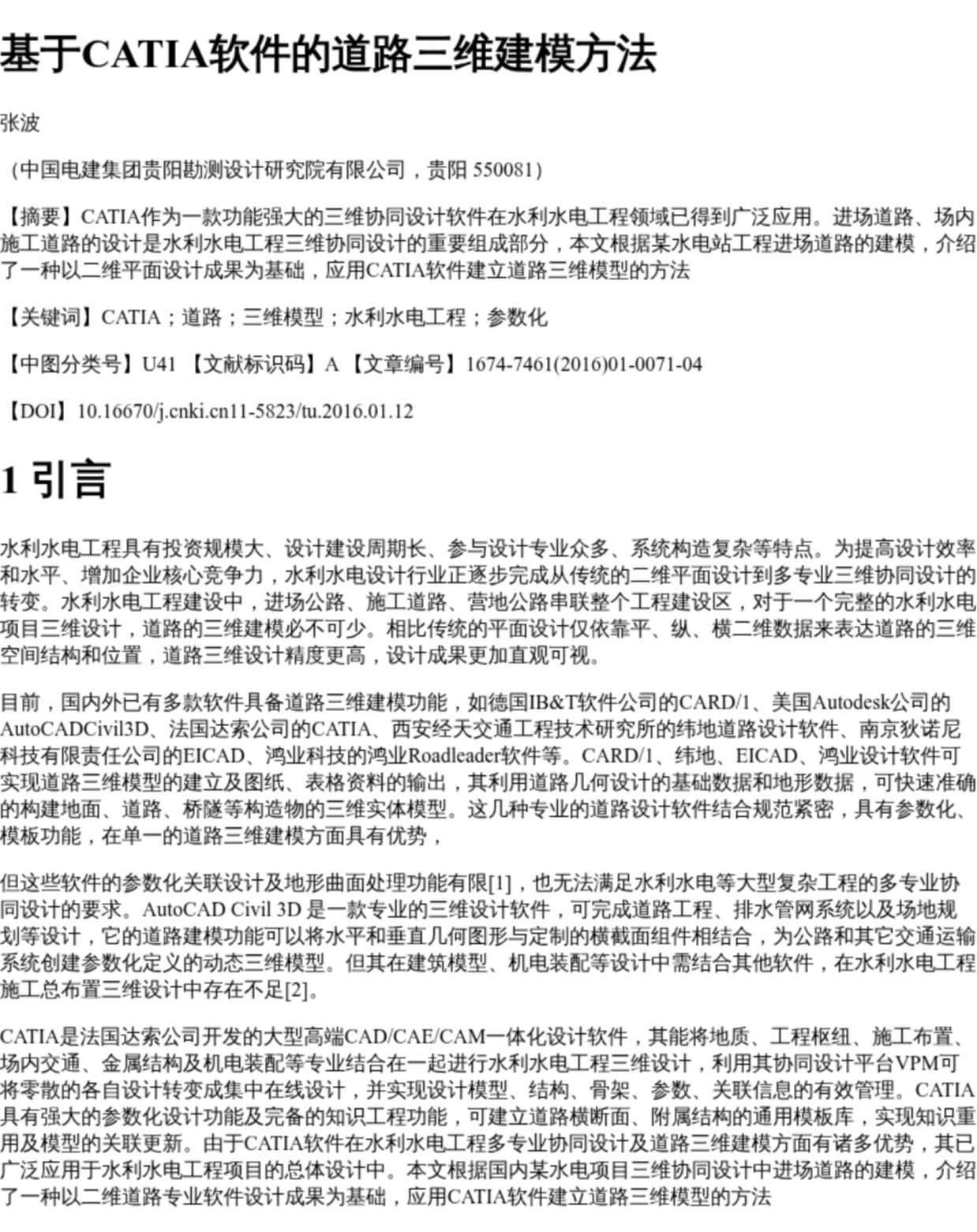}{MinerU2.5}{gray!80}

\end{minipage}
};



\end{tikzpicture}

\end{figure}


\begin{figure}[htpb]
\centering

\begin{tikzpicture}[remember picture]

\node (grid)[inner sep=0pt] {
\begin{minipage}{\textwidth}
\centering

\ShowcaseCellBoxed{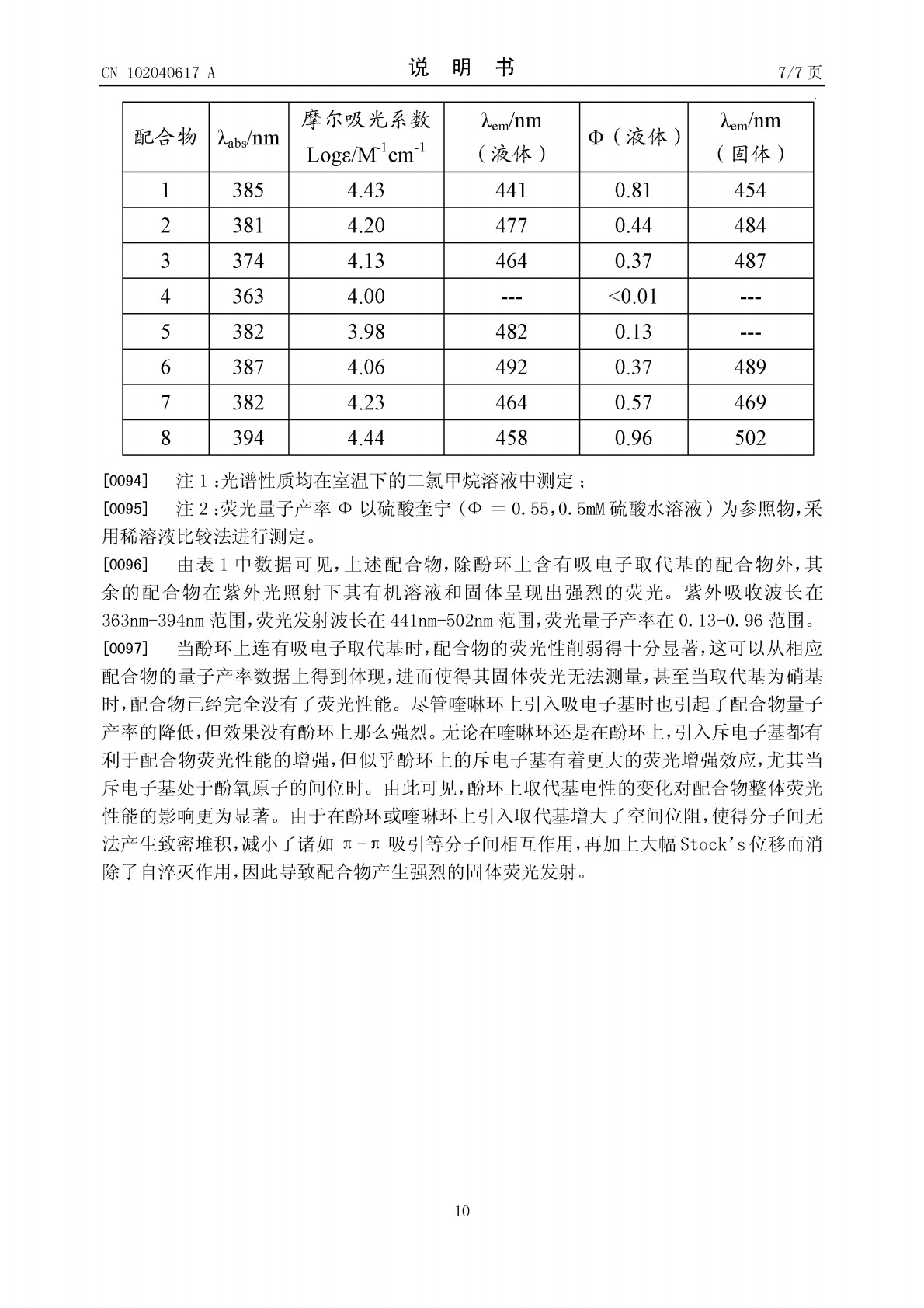}{Original Document}{black!80}
\hfill
\ShowcaseCellBoxed{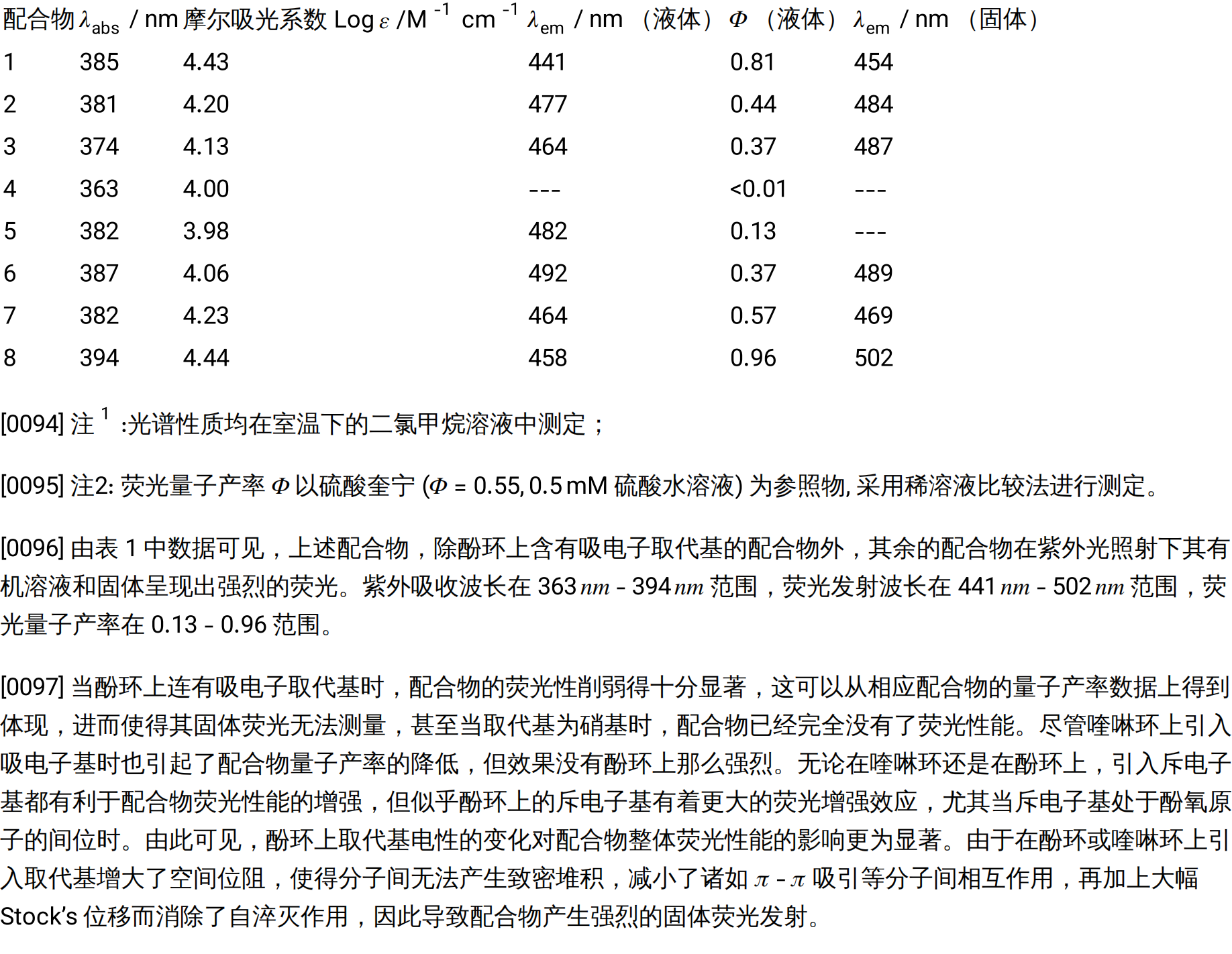}{Uni-Parser}{uniparser-color}

\vspace{12pt} 

\ShowcaseCellBoxed{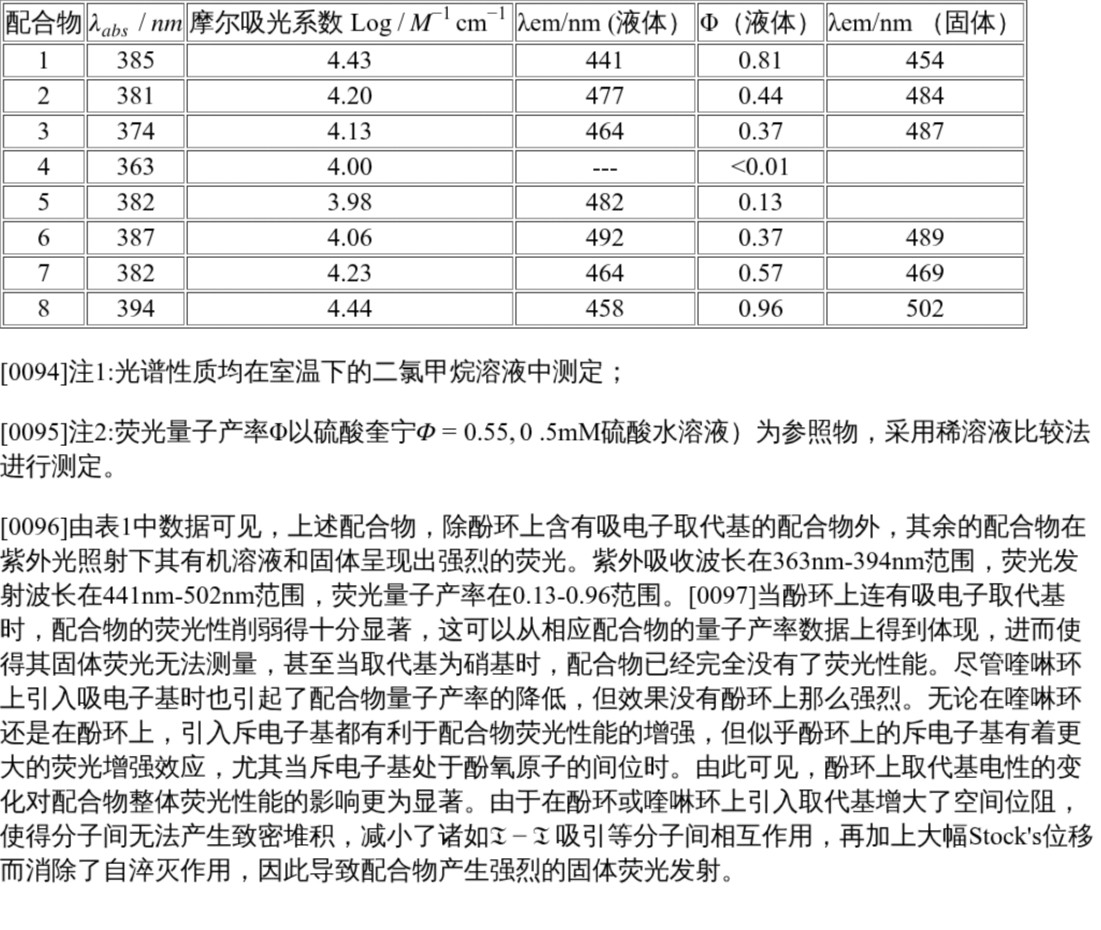}{PP-StructureV3}{gray!80}
\hfill
\ShowcaseCellBoxed{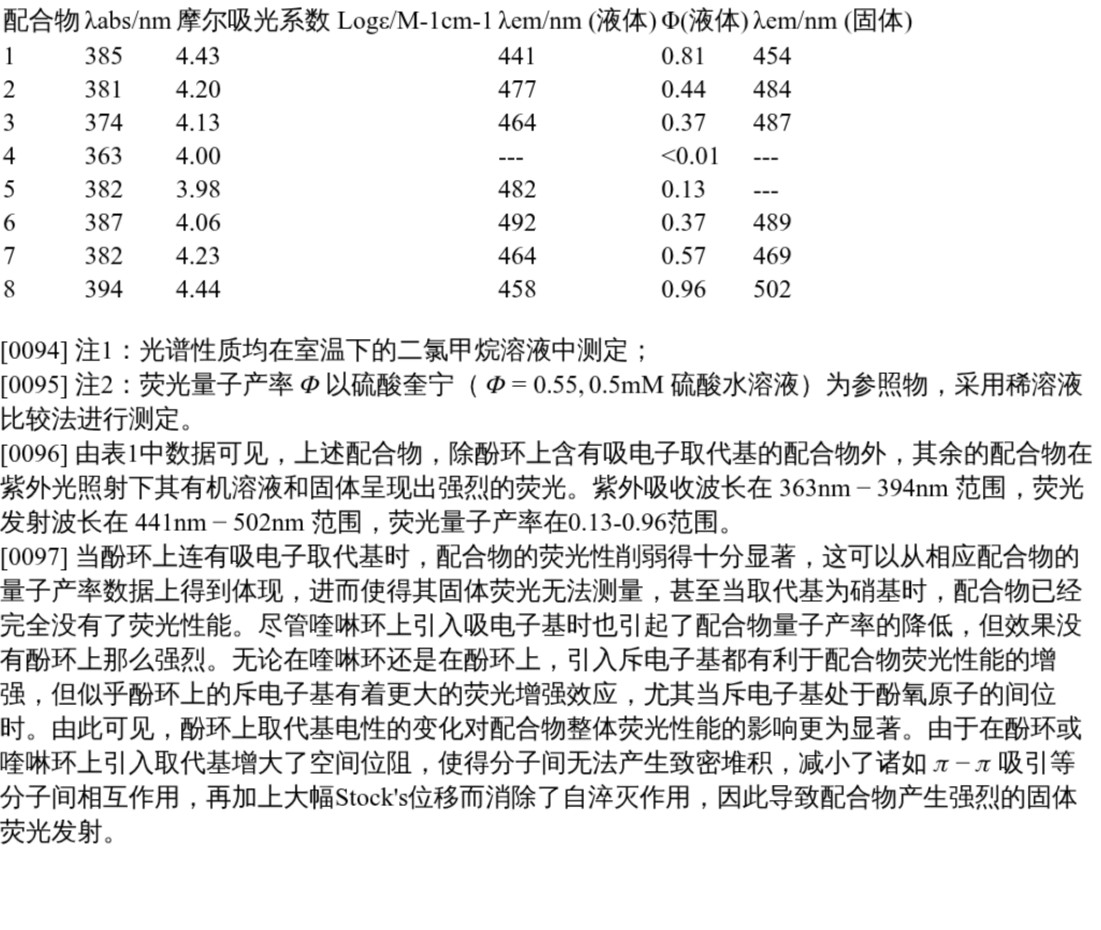}{MinerU2.5}{gray!80}

\end{minipage}
};



\end{tikzpicture}

\end{figure}


%% file: uni-parser.bib
@article{livathinos2025docling,
  title={Docling: An efficient open-source toolkit for ai-driven document conversion},
  author={Livathinos, Nikolaos and Auer, Christoph and Lysak, Maksym and Nassar, Ahmed and Dolfi, Michele and Vagenas, Panos and Ramis, Cesar Berrospi and Omenetti, Matteo and Dinkla, Kasper and Kim, Yusik and others},
  journal={arXiv preprint arXiv:2501.17887},
  year={2025}
}

@article{cui2025paddleocr,
  title={PaddleOCR 3.0 Technical Report},
  author={Cui, Cheng and Sun, Ting and Lin, Manhui and Gao, Tingquan and Zhang, Yubo and Liu, Jiaxuan and Wang, Xueqing and Zhang, Zelun and Zhou, Changda and Liu, Hongen and others},
  journal={arXiv preprint arXiv:2507.05595},
  year={2025}
}

@article{wang2024mineru,
  title={Mineru: An open-source solution for precise document content extraction},
  author={Wang, Bin and Xu, Chao and Zhao, Xiaomeng and Ouyang, Linke and Wu, Fan and Zhao, Zhiyuan and Xu, Rui and Liu, Kaiwen and Qu, Yuan and Shang, Fukai and others},
  journal={arXiv preprint arXiv:2409.18839},
  year={2024}
}

@misc{mathpix,
  title={{MathPix}},
  howpublished={\url{https://mathpix.com/}},
  year={2025},
  author={{MathPix Team}},
}

@article{blecher2023nougat,
  title={Nougat: Neural optical understanding for academic documents},
  author={Blecher, Lukas and Cucurull, Guillem and Scialom, Thomas and Stojnic, Robert},
  journal={arXiv preprint arXiv:2308.13418},
  year={2023}
}

@article{wei2024got,
  title={General ocr theory: Towards ocr-2.0 via a unified end-to-end model},
  author={Wei, Haoran and Liu, Chenglong and Chen, Jinyue and Wang, Jia and Kong, Lingyu and Xu, Yanming and Ge, Zheng and Zhao, Liang and Sun, Jianjian and Peng, Yuang and others},
  journal={arXiv preprint arXiv:2409.01704},
  year={2024}
}

@article{nassar2025smoldocling,
  title={SmolDocling: An ultra-compact vision-language model for end-to-end multi-modal document conversion},
  author={Nassar, Ahmed and Marafioti, Andres and Omenetti, Matteo and Lysak, Maksym and Livathinos, Nikolaos and Auer, Christoph and Morin, Lucas and de Lima, Rafael Teixeira and Kim, Yusik and Gurbuz, A Said and others},
  journal={arXiv preprint arXiv:2503.11576},
  year={2025}
}

@article{li2025monkeyocr,
  title={MonkeyOCR: Document Parsing with a Structure-Recognition-Relation Triplet Paradigm},
  author={Li, Zhang and Liu, Yuliang and Liu, Qiang and Ma, Zhiyin and Zhang, Ziyang and Zhang, Shuo and Guo, Zidun and Zhang, Jiarui and Wang, Xinyu and Bai, Xiang},
  journal={arXiv preprint arXiv:2506.05218},
  year={2025}
}

@misc{ocrflux,
  title={{OCRFlux}: Mastering Complex Layouts and Seamless Page Merging},
  howpublished={\url{https://ocrflux.pdfparser.io/}},
  year={2025},
  author={{OCRFlux Team}},
}

@misc{rednote_hilab_dots_ocr_2025,
  title={{dots.ocr}: Multilingual Document Layout Parsing in a Single Vision-Language Model},
  author={{rednote-hilab}},
  howpublished={\url{https://huggingface.co/rednote-hilab/dots.ocr}},
  year={2025},
}

@misc{mistral_ocr,
  title={{MistralOCR}},
  howpublished={\url{https://mistral.ai/news/}},
  year={2025},
  author={{Mistral Team}},
}

@article{fang2024molparser,
  title={Molparser: End-to-end visual recognition of molecule structures in the wild},
  author={Fang, Xi and Wang, Jiankun and Cai, Xiaochen and Chen, Shangqian and Yang, Shuwen and Tao, Haoyi and Wang, Nan and Yao, Lin and Zhang, Linfeng and Ke, Guolin},
  journal={arXiv preprint arXiv:2411.11098},
  year={2024}
}

@article{yan2025biominer,
  title={BioMiner: A Multi-modal System for Automated Mining of Protein-Ligand Bioactivity Data from Literature},
  author={Yan, Jiaxian and Zhu, Jintao and Yang, Yuhang and Liu, Qi and Zhang, Kai and Zhang, Zaixi and Liu, Xukai and Zhang, Boyan and Gao, Kaiyuan and Xiao, Jinchuan and others},
  journal={bioRxiv},
  year={2025},
  publisher={Cold Spring Harbor Laboratory}
}

@inproceedings{ouyang2025omnidocbench,
  title={Omnidocbench: Benchmarking diverse pdf document parsing with comprehensive annotations},
  author={Ouyang, Linke and Qu, Yuan and Zhou, Hongbin and Zhu, Jiawei and Zhang, Rui and Lin, Qunshu and Wang, Bin and Zhao, Zhiyuan and Jiang, Man and Zhao, Xiaomeng and others},
  booktitle={Proceedings of the Computer Vision and Pattern Recognition Conference},
  pages={24838--24848},
  year={2025},
}

@article{shi2024intelligent,
  title={Intelligent system for automated molecular patent infringement assessment},
  author={Shi, Yaorui and Li, Sihang and Zhang, Taiyan and Fang, Xi and Wang, Jiankun and Liu, Zhiyuan and Zhao, Guojiang and Zhu, Zhengdan and Gao, Zhifeng and Zhong, Renxin and others},
  journal={arXiv preprint arXiv:2412.07819},
  year={2024}
}

@article{zhuang2025doc2sar,
  title={Doc2SAR: A Synergistic Framework for High-Fidelity Extraction of Structure-Activity Relationships from Scientific Documents},
  author={Zhuang, Jiaxi and Li, Kangning and Hou, Jue and Xu, Mingjun and Gao, Zhifeng and Cai, Hengxing},
  journal={arXiv preprint arXiv:2506.21625},
  year={2025}
}

@article{xu2025toward,
  title={Toward a unified benchmark and framework for deep learning-based prediction of nuclear magnetic resonance chemical shifts},
  author={Xu, Fanjie and Guo, Wentao and Wang, Feng and Yao, Lin and Wang, Hongshuai and Tang, Fujie and Gao, Zhifeng and Zhang, Linfeng and E, Weinan and Tian, Zhong-Qun and others},
  journal={Nature Computational Science},
  pages={1--9},
  year={2025},
  publisher={Nature Publishing Group US New York}
}

@article{pang2025paper2poster,
  title={Paper2Poster: Towards Multimodal Poster Automation from Scientific Papers},
  author={Pang, Wei and Lin, Kevin Qinghong and Jian, Xiangru and He, Xi and Torr, Philip},
  journal={arXiv preprint arXiv:2505.21497},
  year={2025}
}

@article{cai2024uni,
  title={Uni-SMART: universal science multimodal analysis and research transformer},
  author={Cai, Hengxing and Cai, Xiaochen and Yang, Shuwen and Wang, Jiankun and Yao, Lin and Gao, Zhifeng and Chang, Junhan and Li, Sihang and Xu, Mingjun and Wang, Changxin and others},
  journal={arXiv preprint arXiv:2403.10301},
  year={2024}
}

@article{li2024scilitllm,
  title={Scilitllm: How to adapt llms for scientific literature understanding},
  author={Li, Sihang and Huang, Jin and Zhuang, Jiaxi and Shi, Yaorui and Cai, Xiaochen and Xu, Mingjun and Wang, Xiang and Zhang, Linfeng and Ke, Guolin and Cai, Hengxing},
  journal={arXiv preprint arXiv:2408.15545},
  year={2024}
}

@article{liao2025innovator,
  title={Innovator: Scientific Continued Pretraining with Fine-grained MoE Upcycling},
  author={Liao, Ning and Wang, Xiaoxing and Lin, Zehao and Guo, Weiyang and Hong, Feng and Song, Shixiang and Yu, Geng and Zhao, Zihua and Xie, Sitao and Wei, Longxuan and others},
  journal={arXiv preprint arXiv:2507.18671},
  year={2025}
}

@misc{openai2025deepresearch,
  author       = {{OpenAI}},
  title        = {Deep Research},
  year         = {2025},
  howpublished = {\url{https://openai.com/deep-research}},
  note         = {[AI research tool]}
}

@article{wang2025uniem3muniversalelectronmicrograph,
  title={UniEM-3M: A Universal Electron Micrograph Dataset for Microstructural Segmentation and Generation},
  author={Xia, Zhiyi and Li, Yiming and Tang, Shi and Fan, Zuxin and Fang, Xi and Tao, Haoyi and Cai, Xiaochen and Ke, Guolin and Zhang, Linfeng and Hong, Yanhui and others},
  journal={arXiv preprint arXiv:2508.16239},
  year={2025}
}

@article{molminer,
    author = {Xu, Youjun and Xiao, Jinchuan and Chou, Chia-Han and Zhang, Jianhang and Zhu, Jintao and Hu, Qiwan and Li, Hemin and Han, Ningsheng and Liu, Bingyu and Zhang, Shuaipeng and Han, Jinyu and Zhang, Zhen and Zhang, Shuhao and Zhang, Weilin and Lai, Luhua and Pei, Jianfeng},
    title = {MolMiner: You Only Look Once for Chemical Structure Recognition},
    journal = {Journal of Chemical Information and Modeling},
    volume = {62},
    number = {22},
    pages = {5321-5328},
    year = {2022},
    doi = {10.1021/acs.jcim.2c00733},
}

@article{qian2023molscribe,
  title={MolScribe: robust molecular structure recognition with image-to-graph generation},
  author={Qian, Yujie and Guo, Jiang and Tu, Zhengkai and Li, Zhening and Coley, Connor W and Barzilay, Regina},
  journal={Journal of Chemical Information and Modeling},
  volume={63},
  number={7},
  pages={1925--1934},
  year={2023},
  publisher={ACS Publications}
}

@article{chen2024molnextr,
  title={MolNexTR: A Generalized Deep Learning Model for Molecular Image Recognition},
  author={Chen, Yufan and Leung, Ching Ting and Huang, Yong and Sun, Jianwei and Chen, Hao and Gao, Hanyu},
  journal={arXiv preprint arXiv:2403.03691},
  year={2024}
}

@article{rajan2023decimer,
  title={DECIMER. ai: an open platform for automated optical chemical structure identification, segmentation and recognition in scientific publications},
  author={Rajan, Kohulan and Brinkhaus, Henning Otto and Agea, M Isabel and Zielesny, Achim and Steinbeck, Christoph},
  journal={Nature communications},
  volume={14},
  number={1},
  pages={5045},
  year={2023},
  publisher={Nature Publishing Group UK London}
}

@inproceedings{rtdetr,
  title={Detrs beat yolos on real-time object detection},
  author={Zhao, Yian and Lv, Wenyu and Xu, Shangliang and Wei, Jinman and Wang, Guanzhong and Dang, Qingqing and Liu, Yi and Chen, Jie},
  booktitle={Proceedings of the IEEE/CVF conference on computer vision and pattern recognition},
  pages={16965--16974},
  year={2024}
}

@article{peng2024dfine,
  title={D-FINE: Redefine regression task in DETRs as fine-grained distribution refinement},
  author={Peng, Yansong and Li, Hebei and Wu, Peixi and Zhang, Yueyi and Sun, Xiaoyan and Wu, Feng},
  journal={arXiv preprint arXiv:2410.13842},
  year={2024}
}

@article{zhao2024doclayout,
  title={Doclayout-yolo: Enhancing document layout analysis through diverse synthetic data and global-to-local adaptive perception},
  author={Zhao, Zhiyuan and Kang, Hengrui and Wang, Bin and He, Conghui},
  journal={arXiv preprint arXiv:2410.12628},
  year={2024}
}

@inproceedings{pfitzmann2022doclaynet,
  title={Doclaynet: A large human-annotated dataset for document-layout segmentation},
  author={Pfitzmann, Birgit and Auer, Christoph and Dolfi, Michele and Nassar, Ahmed S and Staar, Peter},
  booktitle={Proceedings of the 28th ACM SIGKDD conference on knowledge discovery and data mining},
  pages={3743--3751},
  year={2022}
}

@misc{jocher2023ultralytics,
  author = {Glenn Jocher and Jing Qiu and Ayush Chaurasia},
  title = {Ultralytics YOLO},
  repository = {https://github.com/ultralytics/ultralytics},
  url = {https://ultralytics.com},
  license = {AGPL-3.0},
  organization = {Ultralytics},
  year={2025}
}

@article{tian2025yolov12,
  title={Yolov12: Attention-centric real-time object detectors},
  author={Tian, Yunjie and Ye, Qixiang and Doermann, David},
  journal={arXiv preprint arXiv:2502.12524},
  year={2025}
}

@article{steiner2024paligemma,
  title={Paligemma 2: A family of versatile vlms for transfer},
  author={Steiner, Andreas and Pinto, Andr{\'e} Susano and Tschannen, Michael and Keysers, Daniel and Wang, Xiao and Bitton, Yonatan and Gritsenko, Alexey and Minderer, Matthias and Sherbondy, Anthony and Long, Shangbang and others},
  journal={arXiv preprint arXiv:2412.03555},
  year={2024}
}

@article{li2022ppslanet,
  title={Pp-structurev2: A stronger document analysis system},
  author={Li, Chenxia and Guo, Ruoyu and Zhou, Jun and An, Mengtao and Du, Yuning and Zhu, Lingfeng and Liu, Yi and Hu, Xiaoguang and Yu, Dianhai},
  journal={arXiv preprint arXiv:2210.05391},
  year={2022}
}

@article{nassar2022tableformer,
  title={TableFormer: Table Structure Understanding with Transformers},
  author={Nassar, Ahmed and Livathinos, Nikolaos and Lysak, Maksym and Staar, Peter},
  journal={arXiv preprint arXiv:2203.01017},
  year={2022}
}

@article{zhong2019image,
  title={Image-based table recognition: data, model, and evaluation},
  author={Zhong, Xu and ShafieiBavani, Elaheh and Yepes, Antonio Jimeno},
  journal={arXiv preprint arXiv:1911.10683},
  year={2019}
}

@article{bai2025qwen2,
  title={Qwen2. 5-vl technical report},
  author={Bai, Shuai and Chen, Keqin and Liu, Xuejing and Wang, Jialin and Ge, Wenbin and Song, Sibo and Dang, Kai and Wang, Peng and Wang, Shijie and Tang, Jun and others},
  journal={arXiv preprint arXiv:2502.13923},
  year={2025}
}

@misc{openai2025gpt5,
  author       = {OpenAI},
  title        = {Introducing GPT-5},
  year         = {2025},
  url          = {https://openai.com/index/introducing-gpt-5/},
}

@misc{deepmind2025gemini25pro,
  author       = {DeepMind},
  title        = {Gemini 2.5 Pro},
  year         = {2025},
  url          = {https://deepmind.google/models/gemini/pro/},
}

@article{mirza2025chempile,
  title={ChemPile: A 250GB Diverse and Curated Dataset for Chemical Foundation Models},
  author={Mirza, Adrian and Alampara, Nawaf and R{\'\i}os-Garc{\'\i}a, Marti{\~n}o and Abdelalim, Mohamed and Butler, Jack and Connolly, Bethany and Dogan, Tunca and Nezhurina, Marianna and {\c{S}}en, B{\"u}nyamin and Tirunagari, Santosh and others},
  journal={arXiv preprint arXiv:2505.12534},
  year={2025}
}

@misc{cui2025paddleocrvlboostingmultilingualdocument,
      title={PaddleOCR-VL: Boosting Multilingual Document Parsing via a 0.9B Ultra-Compact Vision-Language Model}, 
      author={Cheng Cui and Ting Sun and Suyin Liang and Tingquan Gao and Zelun Zhang and Jiaxuan Liu and Xueqing Wang and Changda Zhou and Hongen Liu and Manhui Lin and Yue Zhang and Yubo Zhang and Handong Zheng and Jing Zhang and Jun Zhang and Yi Liu and Dianhai Yu and Yanjun Ma},
      year={2025},
      eprint={2510.14528},
      archivePrefix={arXiv},
      primaryClass={cs.CV},
      url={https://arxiv.org/abs/2510.14528}, 
}

@misc{dataflow2025,
  author       = {DataFlow Develop Team},
  title        = {DataFlow: A Unified Framework for Data-Centric AI},
  year         = {2025},
  howpublished = {\url{https://github.com/OpenDCAI/DataFlow}},
  note         = {Accessed: 2025-07-08}
}

@misc{dataflow-agent2025,
  author       = {DataFlow-Agent Develop Team},
  title        = {DataFlow-Agent: AI-Powered Data \& Paper Workflow Orchestration Platform},
  year         = {2025},
  howpublished = {\url{https://github.com/OpenDCAI/DataFlow-Agent}},
}

@misc{MinerU.Chem_2025,
  author       = {{MinerU Team}},
  title        = {MinerU.Chem},
  year         = {2025},
  url          = {https://mineru.net/chem/},
}

@misc{MolVision_2025,
  author       = {{XtalPi}},
  title        = {MolVision},
  year         = {2025},
  url          = {https://www.xinsight-ai.com/mol-recognition?mol-recognition=%2Fmol-recognition},
}

@article{xiong2023alphaextractor,
  title={$\alpha$Extractor: a system for automatic extraction of chemical information from biomedical literature},
  author={Xiong, Jiacheng and Liu, Xiaohong and Li, Zhaojun and Xiao, Hongzhong and Wang, Guangchao and Niu, Zhenjiang and Fei, Chaoyuan and Zhong, Feisheng and Wang, Gang and Zhang, Wei and others},
  journal={Sci. China: Life Sci.},
  volume={67},
  pages={618--621},
  year={2023}
}

@misc{kydlicek2025finepdfs,
      title={FinePDFs}, 
      author={Hynek Kydl{\'\i}{\v{c}}ek and Guilherme Penedo and Leandro von Werra},
      year={2025},
      publisher = {Hugging Face},
      journal = {Hugging Face repository},
      howpublished = {\url{https://huggingface.co/datasets/HuggingFaceFW/finepdfs}}
}

@article{wang2025nmrexp,
  title={NMRexp: A database of 3.3 million experimental NMR spectra},
  author={Wang, Jun-Jie and Jin, Yongqi and Zhi, Chen-Yu and Liu, Yu-Jie and Huang, Xu-Hao and Xu, Fanjie and Ji, Xiaohong and Fang, Xi and Tao, Haoyi and E, Weinan and others},
  journal={Scientific Data},
  volume={12},
  number={1},
  pages={1954},
  year={2025},
  publisher={Nature Publishing Group UK London}
}

@misc{marker2023,
  author = {Vikrant Varma},
  title = {Marker: Convert PDF to Markdown and JSON quickly with high accuracy},
  year = {2023},
  publisher = {GitHub},
  journal = {GitHub repository},
  howpublished = {\url{https://github.com/datalab-to/marker}},
}

@article{zhu2025internvl3,
  title={Internvl3: Exploring advanced training and test-time recipes for open-source multimodal models},
  author={Zhu, Jinguo and Wang, Weiyun and Chen, Zhe and Liu, Zhaoyang and Ye, Shenglong and Gu, Lixin and Tian, Hao and Duan, Yuchen and Su, Weijie and Shao, Jie and others},
  journal={arXiv preprint arXiv:2504.10479},
  year={2025}
}

@article{feng2025dolphin,
  title={Dolphin: Document image parsing via heterogeneous anchor prompting},
  author={Feng, Hao and Wei, Shu and Fei, Xiang and Shi, Wei and Han, Yingdong and Liao, Lei and Lu, Jinghui and Wu, Binghong and Liu, Qi and Lin, Chunhui and others},
  journal={arXiv preprint arXiv:2505.14059},
  year={2025}
}

@misc{ocrflux3b2024,
  author = {ChatDOC},
  title = {OCRFlux-3B: A High-Accuracy Multimodal OCR Model for Document Parsing},
  year = {2024},
  publisher = {Hugging Face},
  howpublished = {\url{https://huggingface.co/ChatDOC/OCRFlux-3B}},
}

@article{points-reader,
  title={POINTS-Reader: Distillation-Free Adaptation of Vision-Language Models for Document Conversion},
  author={Liu, Yuan and Zhongyin Zhao and Tian, Le and Haicheng Wang and Xubing Ye and Yangxiu You and Zilin Yu and Chuhan Wu and  Zhou, Xiao and Yu, Yang and Zhou, Jie},
  journal={arXiv preprint arXiv:2509.01215},
  year={2025}
}

@article{poznanski2025olmocr,
  title={olmocr: Unlocking trillions of tokens in pdfs with vision language models},
  author={Poznanski, Jake and Rangapur, Aman and Borchardt, Jon and Dunkelberger, Jason and Huff, Regan and Lin, Daniel and Wilhelm, Christopher and Lo, Kyle and Soldaini, Luca},
  journal={arXiv preprint arXiv:2502.18443},
  year={2025}
}

@misc{Nanonets-OCR-S,
  title={Nanonets-OCR-S: A model for transforming documents into structured markdown with intelligent content recognition and semantic tagging},
  author={Souvik Mandal and Ashish Talewar and Paras Ahuja and Prathamesh Juvatkar},
  year={2025},
}

@article{wei2025deepseek,
  title={Deepseek-ocr: Contexts optical compression},
  author={Wei, Haoran and Sun, Yaofeng and Li, Yukun},
  journal={arXiv preprint arXiv:2510.18234},
  year={2025}
}
